\documentclass{article}

\usepackage{PRIMEarxiv}

\usepackage[utf8]{inputenc} % allow utf-8 input
\usepackage[T1]{fontenc}    % use 8-bit T1 fonts
\usepackage{hyperref}       % hyperlinks
\usepackage{url}            % simple URL typesetting
\usepackage{booktabs}       % professional-quality tables
\usepackage{amsfonts}       % blackboard math symbols
\usepackage{amsthm}
\usepackage{amssymb}
\usepackage{amsmath} 
\usepackage{nicefrac}       % compact symbols for 1/2, etc.
\usepackage{microtype}      % microtypography
\usepackage{lipsum}
\usepackage{fancyhdr}       % header
\usepackage{graphicx}       % graphics
\graphicspath{{media/}}     % organize your images and other figures under media/ folder
\usepackage{amsfonts}
\usepackage{amsmath}
\usepackage{amssymb}
\usepackage{multirow} 
\usepackage{tikz} % Required for TikZ figures
\usetikzlibrary{arrows.meta, positioning, calc, shapes.geometric}
\usepackage{caption}
\usepackage{subcaption}
\usepackage{algpseudocode}
\usepackage{algorithm}
\usepackage[numbers]{natbib}

\DeclareMathOperator{\p2}{p_Z}
\DeclareMathOperator{\Da}{\mathcal{D}_A}
\DeclareMathOperator{\Db}{\mathcal{D}_B}
\DeclareMathOperator{\txy}{\mu}
\DeclareMathOperator{\Gaz}{G_{\Da\to \mathcal{Z}}}
\DeclareMathOperator{\Gab}{G_{\Da\to \Db}}
\DeclareMathOperator{\Gbz}{G_{\Db\to \mathcal{Z}}}
\DeclareMathOperator{\Gba}{G_{\Db\to \Da}}

\newtheorem*{remark}{Remark}

\title{SerpentFlow: Generative Unpaired Domain Alignment via Shared-Structure Decomposition}

\author{
Julie Keisler \\
ARCHES Team\\
INRIA Paris \\
\texttt{julie.keisler@inria.fr}
\And
Anastase Charantonis \\
ARCHES Team\\
INRIA Paris
\And
Yannig Goude \\
EDF Lab Paris-Saclay \\
Université Paris-Saclay
\And
Boutheina Oueslati \\
EDF Lab Paris-Saclay \\
\And
Claire Monteleoni \\
ARCHES Team\\
INRIA Paris \\
University of Colorado Boulder
}

\begin{document}

\maketitle

\begin{abstract}
Domain alignment refers broadly to learning correspondences between data distributions from distinct domains. In this work, we focus on a setting where domains share underlying structural patterns despite differences in their specific realizations. The task is particularly challenging in the absence of paired observations, which removes direct supervision across domains. We introduce a generative framework, called SerpentFlow (SharEd-structuRe decomPosition for gEnerative domaiN adapTation), for unpaired domain alignment. SerpentFlow decomposes data within a latent space into a shared component common to both domains and a domain-specific one. By isolating the shared structure and replacing the domain-specific component with stochastic noise, we construct synthetic training pairs between shared representations and target-domain samples, thereby enabling the use of conditional generative models that are traditionally restricted to paired settings. We apply this approach to super-resolution tasks, where the shared component naturally corresponds to low-frequency content while high-frequency details capture domain-specific variability. The cutoff frequency separating low- and high-frequency components is determined automatically using a classifier-based criterion, ensuring a data-driven and domain-adaptive decomposition. By generating pseudo-pairs that preserve low-frequency structures while injecting stochastic high-frequency realizations, we learn the conditional distribution of the target domain given the shared representation. We implement SerpentFlow using Flow Matching as the generative pipeline, although the framework is compatible with other conditional generative approaches. Experiments on synthetic images, physical process simulations, and a climate downscaling task demonstrate that the method effectively reconstructs high-frequency structures consistent with underlying low-frequency patterns, supporting shared-structure decomposition as an effective strategy for unpaired domain alignment.\footnote{Our code is available here: \url{https://github.com/JulieKeisler/serpentflow}}

\end{abstract}

% keywords can be removed
%\end{keywords}
\section{Introduction}

Generative modeling aims to synthesize realistic samples from complex data distributions that are observed only through examples. Typically, this is achieved by learning a mapping from a simple source distribution, such as a Gaussian, to the target distribution represented by the available data points. Frameworks such as normalizing flows~\citep{normalizingflows}, variational autoencoders (VAEs)~\citep{autoencoder}, Generative Adversarial Networks (GANs)~\citep{gan}, and diffusion-based approaches~\citep{ddpm, score, flowmatching} provide flexible tools for learning such mappings, enabling high-fidelity generation of images~\citep{gan, ddpm}, audio signals~\citep{kong2020diffwave}, or protein structures~\citep{ingraham2019generative, jing2022learning}. Framing data generation as a transport between distributions has proven both theoretically elegant and practically powerful.

Extending this perspective, one can consider transport between two empirical domains, $\mathcal{A}$ and $\mathcal{B}$, each associated with a probability distribution, $p_A$ and $p_B$, again observed only through examples. The goal is to learn a mapping that transforms samples from $\mathcal{A}$ to $\mathcal{B}$ while respecting the underlying distributions. When paired samples $(x_\mathcal{A}, x_\mathcal{B})$ are available, conditional generative methods can directly learn $p(x_\mathcal{B} \mid x_\mathcal{A})$, as in conditional GANs~\citep{conditionnalgans} or diffusion-based conditioned models~\citep{conditionaldiffusion}.

In many real-world scenarios, however, paired data are not available. Domains may correspond to measurements collected under different sensors, resolutions, or physical conditions, without a one-to-one correspondence. This unpaired domain alignment problem has motivated approaches such as CycleGAN~\citep{cyclegan}, MUNIT~\citep{munit}, AlignFlow~\citep{alignflow}, and more recent stochastic-bridge or continuous-time transport methods~\citep{dualimplicitbridge, stochasticinterpolants, bridgematching, schrodingerflow}. These approaches enforce dynamic or latent-space consistency to connect marginal distributions but often lack explicit mechanisms to identify the structures truly shared across domains, which can limit interpretability and consistency between the two domains.

We introduce SerpentFlow (SharEd-structuRe decomPosition for gEnerative domaiN adapTation), a framework for unpaired domain adaptation based on the explicit separation of shared and domain-specific components. We assume that samples from two domains $\mathcal{A}$ and $\mathcal{B}$ can be mapped to a common latent representation that captures the structures shared across domains, while the remaining variability is domain-specific. This shared representation defines a latent distribution that is reachable from both domains and serves as an interface for alignment. SerpentFlow leverages this latent shared space to construct synthetic training pairs. Specifically, samples from the target domain are mapped to the shared representation, while the domain-specific component is replaced by stochastic noise. A conditional generative model is then trained to reconstruct the full target-domain sample from this partially specified input. This training procedure learns the conditional distribution of target-domain–specific content given the shared representation, without requiring paired observations between domains. At inference time, samples from the source domain are projected onto the same shared latent representation. The trained conditional generator is then used to sample target-domain–specific components conditioned on this shared representation, producing aligned outputs in the target domain. Importantly, SerpentFlow does not learn an explicit mapping between domains; instead, alignment is achieved by completing shared structures with domain-consistent details learned on the target domain

In this work, we instantiate SerpentFlow for unsupervised super-resolution tasks, where the shared representation corresponds to coarse-scale information and the domain-specific component captures fine-scale details. To identify the shared component in a data-driven manner, we progressively remove information from the input until samples from both domains become indistinguishable to a domain classifier. This procedure yields a decomposition into shared and domain-specific parts that is consistent across domains. Given this decomposition, we train the generative model on the target domain by reconstructing high-resolution samples from inputs in which only the shared component is preserved and the remaining content is replaced by noise. At inference time, low-resolution samples from the source domain are mapped to the same shared representation and completed using the trained generator, resulting in high-resolution outputs aligned with the target domain. In the super-resolution setting considered here, this shared representation is naturally realized through a frequency-based decomposition: low-frequency components capture large-scale structures common to both domains, while high-frequency components encode fine details that differ across domains. The frequency cutoff is selected automatically using the domain classifier, ensuring that only domain-invariant structure is retained.

We validate SerpentFlow on three datasets. First, a controlled image dataset where high-frequency components are artificially removed to emulate domain shifts. Second, a dataset of simulated physical processes, for which simulation equations were sampled at two different spatial frequencies to generate high- and low-resolution data. Third, a climate downscaling application, enhancing coarse-resolution simulations to recover fine spatial patterns. Across all experiments, SerpentFlow consistently reconstructs coherent high-frequency structures while preserving underlying low-frequency content, ensuring both physical and statistical consistency.

In summary, our contributions are:
\begin{itemize}
    \item We introduce SerpentFlow, a pipeline for unpaired domain alignment that decomposes data into shared and domain-specific components and constructs synthetic paired samples by injecting stochasticity in the domain-specific part.
    \item We propose a frequency-based instantiation of this pipeline for unsupervised super-resolution tasks, using a classifier to determine the low-/high-frequency cutoff automatically, and a generative model to reconstruct high-frequency content.
    \item Our approach demonstrates improved coherence, robustness, and generalization over existing unpaired generative methods across controlled images, physical simulations, and climate downscaling datasets.
\end{itemize}

\section{Background}\label{part:related_work}

\paragraph{Generative Frameworks.}
Generative modeling aims to learn transformations from simple latent distributions, typically Gaussian, to complex data distributions that are observed only through finite examples. Let $z \sim \mathcal{N}(0, I)$ denote a latent variable and $x \sim p(x)$ a data sample. The objective is to learn a mapping $T_\theta : z \mapsto x$ that captures the structure of $p(x)$. Different frameworks implement this idea using distinct mathematical approaches. Generative Adversarial Networks (GANs)~\citep{gan} learn an implicit generator trained adversarially against a discriminator, while Normalizing Flows~\citep{normalizingflows} define invertible mappings with exact likelihood evaluation. Neural Ordinary Differential Equations (Neural ODEs)~\citep{neuralode} extend flows to continuous-time dynamics, and Diffusion Models~\citep{ddpm, score} learn to reverse a stochastic noising process, achieving state-of-the-art fidelity and stability. More recent formulations, such as Flow Matching~\citep{flowmatching} and Stochastic Interpolants~\citep{stochasticinterpolants}, directly learn time-dependent velocity fields that transport probability mass between source and target distributions. Together, these frameworks provide flexible and mathematically grounded tools for defining transformations between distributions, which makes them particularly suitable for domain alignment tasks.

\paragraph{Unpaired Domain Alignment.}
Unpaired domain alignment addresses the problem of learning correspondences between samples from two domains, $\mathcal{D}_A$ and $\mathcal{D}_B$, drawn from distributions $p_A$ and $p_B$, without access to paired observations. Classical applications include image-to-image translation, such as converting horses to zebras or summer to winter scenes~\citep{cyclegan}, artistic style transfer, for instance, transforming photographs into Van Gogh-like paintings~\citep{styletransfer}, or more generally generating images with controlled style mixing using StyleGAN architectures~\citep{stylegan}. Another example is unsupervised super-resolution, where the goal is to infer fine-scale details from coarse-resolution inputs~\citep{unsupervisedsr}. Related ideas also appear in scientific domains, including compressive sensing~\citep{freqreconstruction, compressedsensing} and climate downscaling, where high-resolution physical fields are reconstructed from coarse simulations~\citep{climalign, bischoff2024unpaired}.

Early methods, such as CycleGAN~\citep{cyclegan}, rely on cycle-consistent adversarial training with bidirectional generators to preserve content across domains. UNIT~\citep{munit}, AlignFlow~\citep{alignflow}, and Dual Diffusion Implicit Bridges (DDIB)~\citep{dualimplicitbridge} leverage a shared latent representation (usually a Gaussian) as an interface between both domains. More recent stochastic transport frameworks, including Schrödinger Bridge Flows~\citep{schrodingerflow} and Rectified Flows~\citep{rectifiedflow}, model continuous trajectories between $p_A$ and $p_B$, often avoiding adversarial training while retaining explicit transport dynamics. In unsupervised super-resolution, diffusion-bridge formulations~\citep{bischoff2024unpaired} exploit the idea that by partially adding noise to low-resolution data, we arrive on the path of a diffusion model that generates high-resolution data.

Although these methods achieve visually convincing results in classical image translation tasks, it remains challenging to quantify which structures should be preserved or modified. In scientific and physical applications, however, well-defined metrics allow evaluation of whether outputs respect the underlying domain structure. This motivates approaches that explicitly separate shared and domain-specific components for robust and interpretable unpaired domain alignment.

\section{Method}\label{method}

Although our method builds upon existing frameworks, it differs in that it aims to establish a more explicit mapping between the two domains, making it clear which aspects of the source should be retained and which should be transformed. The underlying principle is that, for domain alignment to be meaningful, there must exist a shared structure between the domains alongside genuine differences in other parts of the data. The goal of domain alignment is then to preserve this shared component while appropriately modifying the domain-specific parts of the source, ensuring consistency with the portions that are not intended to change.

\subsection{Domain alignment with shared structure}

Let us consider two domains $\mathcal{D}_A \subset \mathcal{X}$ and $\mathcal{D}_B \subset \mathcal{X}$ embedded in the same space $\mathcal{X} \subset \mathbb{R}^D$. We denote by $p_A$ and $p_B$ the distributions of samples in $\mathcal{D}_A$ and $\mathcal{D}_B$, respectively. The goal of domain alignment is to learn a mapping that transforms samples from $p_A$ to $p_B$. For such an alignment to be meaningful, the two domains must share at least a minimal amount of common structure.

Formally, let there exist a bijective mapping towards a given latent space $\mathcal{Z}$:
\begin{equation}
\txy : \mathcal{X} \to \mathcal{Z}.
\end{equation}

such that
\begin{equation}
\txy(\mathcal{D}_A) = \mathcal{B}_A \subset \mathcal{Z}, \qquad 
\txy(\mathcal{D}_B) = \mathcal{B}_B \subset \mathcal{Z}.
\end{equation}

We assume that both $\mathcal{B}_A$ and $\mathcal{B}_B$ can be decomposed into a shared component and a domain-specific (distinct) one:
\begin{equation}
\mathcal{B}_A = \mathcal{B}_A^S \oplus \mathcal{B}_A^D, \qquad 
\mathcal{B}_B = \mathcal{B}_B^S \oplus \mathcal{B}_B^D,
\end{equation}
with the assumption that $\mathcal{B}_A^S = \mathcal{B}_B^S = \mathcal{B}^S$. In this setting, $\mathcal{B}^S$ represents the latent structure shared between the two domains, while $\mathcal{B}_A^D$ and $\mathcal{B}_B^D$ capture the domain-specific variability.

For a sample $x_A^i \sim p_A$, we write
\begin{equation}
\txy(x_A^i) = z_A^i = z_A^{i,S} + z_A^{i,D},
\end{equation}
where $z_A^{i,S} \in \mathcal{B}^S$ and $z_A^{i,D} \in \mathcal{B}_A^D$. Similarly, for $x_B^i \sim p_B$,
\begin{equation}
\txy(x_B^i) = z_B^i = z_B^{i,S} + z_B^{i,D}.
\end{equation}

A white noise sample $\epsilon \sim \mathcal{N}(0, \mathbf{I})$ in the input space is mapped to
\begin{equation}
\txy(\epsilon) = z_\epsilon = z_\epsilon^S + z_\epsilon^D \in \mathcal{B}_\epsilon^S \oplus \mathcal{B}_\epsilon^D \subset \mathcal{Z}.
\end{equation}

We can then construct pseudo-pairs by combining the shared component from the target domain with a stochastic realization of the domain-specific part:
\begin{equation}
(\tilde{x}_B^i, x_B^i) \in \big( \txy^{-1}(\mathcal{B}^S \oplus \mathcal{B}_\epsilon^D) \times \mathcal{D}_B\big), \qquad 
\tilde{x}_B^i =  \txy^{-1}(z_B^{i,S} + z_\epsilon^D).
\end{equation}

A generative model $f_\theta$ can then be trained to map these pseudo-pairs:
\begin{equation}
f_\theta :  \txy^{-1}(\mathcal{B}^S \oplus \mathcal{B}_\epsilon^D) \to \mathcal{D}_B.
\end{equation}

This formulation can be interpreted as learning the conditional distribution of domain-specific components of $p_B$ given the shared component $z_B^S$:
\begin{equation}
f_\theta : \tilde{x}_B \mapsto x_B \sim p_B(x_B \mid z_B^S).
\end{equation}

Once trained, a new sample $x_A^i \in \mathcal{D}_A$ can be transferred to the target domain $\mathcal{D}_B$ via
\begin{equation}
f_\theta(\tilde{x}_A^i) = f_\theta\!\left(\txy^{-1}(z_A^{i,S} + z_\epsilon^D)\right).
\end{equation}

The main challenges of applying this general framework in practice lie in (i) identifying a latent space $\mathcal{Z}$ where the shared and domain-specific components are clearly separable, and (ii) finding a suitable bijective mapping $\txy$. Moreover, $\mathcal{B}^S$ should be as large as possible to retain sufficient information from the source sample $x_A^i$. In the limiting case where $\mathcal{D}_A$ and $\mathcal{D}_B$ share no common structure, we would have $\mathcal{B}^S = \varnothing$, which corresponds to unconditional generative modeling.

\begin{remark}
In this formulation, conditioning is expressed as a summation in the latent space $\mathcal{Z}$. Other conditioning mechanisms could be considered without loss of generality; these alternatives will be discussed in the experimental section (see Section~\ref{part:exp}).
\end{remark}

\begin{center}
\begin{tikzpicture}[node distance=2.5cm, every node/.style={align=center}]
    % Nodes
    \node (DA) [draw, rectangle, minimum width=2.5cm, minimum height=1cm] {$\mathcal{D}_A$};
    \node (DB) [draw, rectangle, right=3cm of DA, minimum width=2.5cm, minimum height=1cm] {$\mathcal{D}_B$};
    \node (Y) [draw, rectangle, above=of $(DA)!0.5!(DB)$, minimum width=3cm, minimum height=1cm] {$\mathcal{Z} = \mathcal{B}^S \oplus \mathcal{B}^D$};

    % Arrows to latent space
    \draw[->, thick] ([yshift=1mm]DA.north) -- node[left, xshift=-2mm] {$\txy$} ([xshift=-2mm, yshift=-1mm]Y.south);
    \draw[->, thick] ([yshift=1mm]DB.north) -- node[right, xshift=2mm] {$\txy$} ([xshift=2mm, yshift=-1mm]Y.south);

    % Main transformation
    \draw[->, thick, dashed, bend left=30] ([xshift=1mm]DA.east) to node[above, yshift=1mm] {$f_\theta: \; \txy^{-1}(\mathcal{B}^S \oplus \mathcal{B}^D_\epsilon) \to \mathcal{D}_B$} ([xshift=-1mm]DB.west);

    % Flow matching learning
    \draw[->, thick] ([xshift=1mm, yshift=-1mm]Y.east) to[out=0,in=90] node[right, align=left, xshift=5mm] {learn $f_\theta$\\(e.g.\ via flow matching)} ([xshift=2mm, yshift=1mm]DB.north);

    % Labels
    \node[below=0.2cm of DA] {Source domain};
    \node[below=0.2cm of DB] {Target domain};
    \node[above=0.2cm of Y] {Shared latent space};
\end{tikzpicture}
\end{center}

\subsection{Frequency-based instantiation of the shared-domain hypothesis}

The proposed framework assumes a latent space $\mathcal{Z}$ in which shared and domain-specific structures can be effectively disentangled. 
In the context of unsupervised hyperresolution, we instantiate this latent space in the Fourier domain, where each field is decomposed into its spatial frequency components. 
Low-frequency modes encode large-scale, coherent structures that are typically shared across domains, whereas high-frequency modes capture fine-scale, domain-specific variations that are often unresolved in coarse observations. 
This frequency-based decomposition therefore provides a natural and interpretable foundation for separating and reconstructing the components relevant to unsupervised hyperresolution.

The Fourier representation is particularly well suited to this setting, as it organizes information by scale and locality. 
Many complex systems exhibit scale-dependent behavior that becomes more transparent in the frequency domain \citep{boyd2001fourier, canuto2006spectral, falck2025fourier}. 
In physics-based simulations, resolving fine-scale features is often computationally prohibitive, and the governing operators frequently act selectively across distinct frequency bands \citep{doumeche2024physics}. 
Moreover, practical data acquisition and sampling processes effectively behave as low-pass filters, attenuating or truncating high-frequency content \citep{shannon1949communication}. 
Prior work on signal reconstruction has demonstrated that missing high-frequency information can, under sparsity or convex regularization assumptions, be recovered from limited low-frequency observations \citep{freqreconstruction}, although such guarantees typically hold only in lower-dimensional or idealized settings. 
Taken together, these considerations motivate the Fourier domain as a principled and versatile latent space for disentangling shared low-frequency structure from domain-specific high-frequency detail in hyperresolution tasks.

\vspace{1em}
\paragraph{Fourier-domain formulation.}
Let $\mathcal{X} \subset \mathbb{R}^D$ denote the space of observations (e.g., gridded physical fields or images). We define the Fourier transform and its inverse as:
\begin{equation}
\mathcal{F}\{x\}(\boldsymbol{\xi}) = \int_{\mathbb{R}^D} x(\mathbf{r}) e^{-2\pi i \boldsymbol{\xi}\cdot\mathbf{r}} \, d\mathbf{r}, 
\qquad
\mathcal{F}^{-1}\{y\}(\mathbf{r}) = \int_{\mathbb{R}^D} y(\boldsymbol{\xi}) e^{2\pi i \boldsymbol{\xi}\cdot\mathbf{r}} \, d\boldsymbol{\xi}.
\label{eq:fourier}
\end{equation}

In this representation, large-scale structures correspond to low-frequency components (small $\|\boldsymbol{\xi}\|$), while small-scale, domain-specific variations correspond to high-frequency components. Introducing a cutoff frequency $\omega_c$, we define the decomposition:
\begin{equation}
\mathcal{B}^S = \{ z : \|\boldsymbol{\xi}\| < \omega_c \}, \qquad
\mathcal{B}^D = \{ z : \|\boldsymbol{\xi}\| \geq \omega_c \}.
\label{eq:cutoff}
\end{equation}

The shared structure $\mathcal{B}^S$ thus corresponds to the low-frequency backbone common to both domains, while $\mathcal{B}^D$ captures domain-specific variability.

\vspace{1em}
\paragraph{Automatic cutoff selection.}\label{method_cutoff}
The cutoff $\omega_c$ defines the boundary between shared and domain-specific content. To estimate $\omega_c$ in a data-driven manner, we introduce a simple discriminator-based criterion.

For a candidate $\omega_c$, we apply a low-pass filter to each sample:
\begin{equation}
x^S = \mathcal{F}^{-1}\!\left[\chi_{<\omega_c} \cdot \mathcal{F}(x)\right],
\label{eq:lowpass}
\end{equation}
where $\chi_{<\omega_c}$ is the indicator function of frequencies $\|\boldsymbol{\xi}\| < \omega_c$. A convolutional neural network classifier $D_\psi$ is trained to discriminate whether $x^S$ originates from $\mathcal{D}_A$ or $\mathcal{D}_B$, using the standard binary cross-entropy loss:
\begin{equation}
\mathcal{L}_D(\psi; \omega_c) = 
- \mathbb{E}_{x_A \sim \mathcal{D}_A}\!\left[\log D_\psi(x_A^D)\right]
- \mathbb{E}_{x_B \sim \mathcal{D}_B}\!\left[\log \left(1 - D_\psi(x_B^D)\right)\right].
\label{eq:discriminator_loss}
\end{equation}

Starting from a high cutoff $\omega_c$, we progressively decrease its value. As long as the classifier can reliably distinguish between domains (i.e., accuracy significantly higher than 50\%), the low-frequency representation still contains domain-specific information. The optimal cutoff $\omega_c^*$ is defined as the smallest value for which the discriminator accuracy approaches random guessing:
\begin{equation}
\text{Acc}(D_\psi; \omega_c^*) \approx 0.5.
\label{eq:omega_opt}
\end{equation}

This stopping criterion ensures that $\mathcal{B}^S$ contains the maximal domain-invariant component while excluding domain-specific features. As a sanity check, we verify that when no filtering is applied ($\omega_c \to \infty$), the discriminator achieves perfect separation, confirming that the network has sufficient discriminative capacity.

\vspace{1em}

\paragraph{Practical workflow.}
Once $\omega_c^*$ is selected, pseudo-pairs are generated by replacing the high-frequency content of one domain with noise, and the generative model $f_\theta$ is trained on these pairs to learn the conditional mapping to the target domain. SerpentFlow pseudo-code is given Algorithm~\ref{alg:serpentflow}.
\begin{algorithm}[htbp]
\caption{SerpentFlow: Frequency-based pseudo-pair generation, alignment, and inference}
\label{alg:serpentflow}
\begin{algorithmic}[1]
\Require Source domain $\mathcal{D}_A$, target domain $\mathcal{D}_B$, discriminator $D_\psi$, generator $f_\theta$
\State Initialize cutoff $\omega_c \leftarrow \omega_{\text{max}}$
\vspace{0.3em}
\Repeat
    \State Apply low-pass filter: 
    \[
        x^S = \mathcal{F}^{-1}\big[\chi_{<\omega_c} \cdot \mathcal{F}(x)\big]
    \]
    
    \State Train and compute discriminator accuracy $\text{Acc}(D_\psi; \omega_c)$
    \State Decrease cutoff frequency $\omega_c$
\Until{$\text{Acc}(D_\psi; \omega_c) \approx 0.5$}
\State Set $\omega_c^* \leftarrow \omega_c$
\vspace{0.5em}
\Statex
\Comment{\textbf{Phase 1: Pseudo-pair generation and training}}
\For{each sample $x_B \in \mathcal{D}_B$}
    \State Sample noise $\epsilon \sim \mathcal{N}(0, \mathbf{I})$
    \State Construct pseudo-input:
    \begin{equation}
        \tilde{x}_B = \mathcal{F}^{-1}\big[\chi_{<\omega_c^*}\mathcal{F}(x_B) 
        + \chi_{\geq \omega_c^*}\mathcal{F}(\epsilon)\big]
    \end{equation}
    \State Train generative pipeline $f_\theta$ to reconstruct $x_B$:
    \begin{equation}
        \mathcal{L}_G(\theta) = \ell(f_\theta, \tilde{x}_B, x_B)
    \end{equation}
\EndFor
\vspace{0.5em}
\Statex
\Comment{\textbf{Phase 2: Inference on source domain $\mathcal{D}_A$}}
\For{each sample $x_A \in \mathcal{D}_A$}
    \State Sample $\epsilon \sim \mathcal{N}(0, \mathbf{I})$
    \State Construct pseudo-sample:
    \begin{equation}
        \tilde{x}_A = \mathcal{F}^{-1}\big[\chi_{<\omega_c^*}\mathcal{F}(x_A)
        + \chi_{\geq \omega_c^*}\mathcal{F}(\epsilon)\big]
    \end{equation}
    \State Generate aligned target-domain output:
    \begin{equation}
        \hat{x}_A = \mathcal{G}(f_\theta, \tilde{x}_A)
    \end{equation}
\EndFor
\end{algorithmic}
\end{algorithm}

\subsection{Choice of generative model}

SerpentFlow is, in theory, agnostic to the choice of the generative approach used for $f_\theta$. 
In principle, $f_\theta$ could be instantiated using GANs, diffusion models, normalizing flows, or other generative paradigms, as long as it learns to map the pseudo-paired distribution $p_0$ onto the target domain $p_B$. In this work, we adopt a Flow Matching approach~\citep{flowmatching}, which provides a deterministic, continuous-time generative transport pipeline. 
Flow Matching defines a neural velocity field $f_\theta(x,t)$ that governs the evolution of samples according to an ordinary differential equation (ODE):
\begin{equation}
\frac{dx_t}{dt} = f_\theta(x_t, t), \qquad x_0 \sim p_0,
\label{eq:ode_velocity}
\end{equation}
where $f_\theta$ is the learnable component and $p_0$ the distribution of pseudo-samples:
\[
\tilde{x}_B = \txy^{-1}(z_B^S + z_\epsilon^D),
\]
with $z_B^S$ the shared low-frequency component and $z_\epsilon^D$ the stochastic high-frequency noise. The flow transports source samples continuously from $p_0$ to the target distribution $p_B$. We define a linear interpolation path (alternative paths could be used in principle):
\[
x_t = (1-t)\tilde{x}_B + t x_B, \qquad \dot{x}_t^* = x_B - \tilde{x}_B,
\]
with skewed sampling of $t$ to emphasize regions of high variability. 
The Flow Matching objective is
\begin{equation}
\mathcal{L}_{\mathrm{FM}}(\theta) = 
\mathbb{E}_{t} \, \mathbb{E}_{(\tilde{x}_B, x_B) \sim (p_0,p_B)}
\Big[\| f_\theta(x_t,t) - \dot{x}_t^* \|_2^2 \Big] = \ell(f_\theta, \tilde{x}_B,x_B),
\label{eq:flow_matching}
\end{equation}
which trains $f_\theta$ to approximate the instantaneous displacement along the chosen path.

At inference, a source sample $x_A \in \mathcal{D}_A$ is mapped to the Fourier latent space $\mathcal{Z}$, and its shared component $z_A^S$ is extracted. A pseudo-sample is then constructed
\[
\tilde{x}_A = \txy^{-1}(z_A^S + z_\epsilon^D).
\]
The generated sample is obtained by integrating the velocity field along the ODE
\begin{equation}
\frac{dx_t}{dt} = f_\theta(x_t, t), \quad x_{t_0} = \tilde{x}_A,
\end{equation}
from $t_0=1$ to $t_1=0$ using a numerical solver, denoted $\mathrm{ODS}$:
\begin{equation}\label{eq:ods}
    \hat{x}_A = \mathrm{ODS}(\tilde{x}_A, f_\theta, t_0=1, t_1=0) = \tilde{x}_A + \int_{t_1=1}^{t_0=0} f_\theta(x_t,t) \, dt = \mathcal{G}(f_\theta, \tilde{x}_A),
\end{equation}
producing a sample consistent with $p_B$ while preserving the low-frequency structure from the source.

\begin{remark}
In score-based generative approaches such as Flow Matching or diffusion models, it is natural to combine the low-frequency backbone $\mathcal{B}^S$ with stochastic high-frequency components $\mathcal{B}^D$ as a sum. 
The noise schedule used to construct the stochastic path from a Gaussian prior to the data typically removes high-frequency content before affecting low frequencies \citep{bischoff2024unpaired, falck2025fourier}, which aligns with our decomposition: the low-frequency structures in $\mathcal{B}^S$ are preserved, while the high-frequency details in $\mathcal{B}^D$ are generated stochastically. 
Consequently, the model naturally respects the underlying data structure while allowing realistic high-frequency variability.
\end{remark}

\section{Experiments}
\label{part:exp}

We evaluate SerpentFlow across synthetic and physical settings to assess its ability to align domains that share common large-scale structures but differ in fine-scale components. In the experiments presented here, we focus on unsupervised spatial hyper-resolution, aiming to reconstruct high-resolution fields from coarse observations. Without loss of generality, the framework can also be applied to temporal signals or spatio-temporal data, for instance, to interpolate or refine time series or dynamical fields (a toy example of time series super-resolution is available in Appendix~\ref{app:ts}). Three datasets are used: (i) a controlled synthetic image dataset \citep{larochelle2007empirical} called MNIST Rotated with Background Images (MRBI), where high-frequency content has been removed, (ii) a fluid simulation dataset \citep{schrodingerflow,bischoff2024unpaired}, and (iii) a realistic climate downscaling task from a Coupled Model Intercomparison Project Phase 6 \citep{cmip6}, CMIP6 in short, global circulation model to ECMWF Reanalysis v5 \citep{hersbach2020era5}, also called ERA5, wind fields over France.

\subsection{Experimental setup}

For all experiments, the generative model $f_\theta$ is instantiated as a U-Net-based architecture trained within the frequency-conditioned framework described in Section~\ref{method}. The shared-domain component corresponds to the low-frequency band, while domain-specific variability is introduced as stochastic high-frequency realizations. Unless specified otherwise, models are trained using an Adam optimizer with a learning rate of $10^{-4}$, batch size of 32, and skewed time sampling $t \sim \text{Skewed}(0,1)$, emphasizing regions where the distribution exhibits more variability. In practice, the skewed sampling is implemented as
\begin{equation}
t = \frac{1}{1 + \sigma}, \quad \sigma = \exp(P_\text{std} \cdot \epsilon + P_\text{mean}), \quad \epsilon \sim \mathcal{N}(0,1),
\end{equation}
with $P_\text{mean}=-1.2$ and $P_\text{std}=1.2$, and $t$ clipped to $[10^{-4},1]$. This concentrates training on intermediate regions along the Flow Matching path where the velocity field exhibits larger variations.

During inference, pseudo-samples are transported through the learned velocity field $f_\theta$ by integrating the ordinary differential equation (Eq.~\ref{eq:ode_velocity}). In our implementation, we use the Dormand-Prince 5(4) method~\citep{dormand1980family} for adaptive integration.

\paragraph{Baselines.}  
We compare our method against the following state‑of‑the‑art approaches for unpaired domain alignment:  
\begin{itemize}
    \item \textbf{Diffusion Bridge} — the approach from “Unpaired Downscaling of Fluid Flows with Diffusion Bridges”~\citep{bischoff2024unpaired}, which uses a diffusion‐based latent bridge for unpaired data translation, see Figure~\ref{fig:diffbridge}.  
    \item \textbf{Dual FM} — Dual Diffusion Implicit Bridge~\citep{dualimplicitbridge}, which leverages two diffusion processes and an intermediate Gaussian latent space for bidirectional transport between domains, see Figure~\ref{fig:ddib}. In their original paper, Dual FM demonstrated superior performance compared to earlier image‐translation methods such as CycleGAN \citep{cyclegan} or AlignFlow \citep{alignflow}. 
\end{itemize}

We also note that other paradigms—such as the stochastic‑interpolants framework~\citep{stochasticinterpolants} or bridge‐matching algorithms~\citep{bridgematching}—while promising from a theoretical standpoint, have been found in practice to be difficult to stabilize and to converge reliably in high‐dimensional generative transport tasks \citep{hardStoInt1, hardStoInt2}, and therefore did not include them within the baselines.  

All baselines were trained in a Flow Matching setting (hence the name Dual FM for the Dual Diffusion Implicit Bridge baseline), using the same UNet architecture, with identical hyperparameters, including number of epochs, batch size, diffusion path length, and learning rate scheduler. For the Diffusion Bridge baseline, we optimized the interpolation parameter $t^\ast$ using a classifier to determine the point at which the low-frequency components of the source and target domains become indistinguishable. To evaluate robustness, we additionally tested $t^\ast - 0.1$ and $t^\ast + 0.1$.

\subsection{Synthetic frequency reconstruction on an image dataset}
\label{subsec:mrbi}

The first experiment relies on the Modified Rotated Background Images (MRBI) dataset~\citep{larochelle2007empirical}, a variant of the Modified National Institute of Standards and Technology (MNIST) digits \citep{mnist} in which handwritten digits are superimposed on random natural-image backgrounds. 
Each sample is a grayscale image of size $(1, 28, 28)$.

\paragraph{Setup.}
We construct two distinct domains from the MRBI dataset:
\begin{itemize}
    \item $\mathcal{D}_A$: low-pass filtered MRBI images, where high-frequency components have been removed;
    \item $\mathcal{D}_B$: original MRBI images containing the full frequency spectrum.
\end{itemize}

The filtering is performed in the Fourier domain using a cutoff frequency $\omega_c$, which isolates the shared low-frequency structure $\mathcal{B}^S$ while removing the domain-specific high-frequency details $\mathcal{B}^D$. 
As a result, samples from $\mathcal{D}_A$ retain only the coarse, large-scale content of the digits, whereas samples from $\mathcal{D}_B$ preserve the complete frequency composition, including fine-grained details.

A random train--test split is applied to both domains. 
The generative model $f_\theta$ is trained to map filtered samples from $\mathcal{D}_A$ to their corresponding full-resolution counterparts in $\mathcal{D}_B$, thereby reconstructing the missing high-frequency content conditioned on the available low-frequency information. 
This experimental setting constitutes an idealized instantiation of our framework, characterized by a clean separation between low and high frequencies, a bijective Fourier representation, and direct access to the shared latent subspace.

\paragraph{Evaluation.}
To quantitatively assess reconstruction quality, we use two complementary classifiers. Let $y\in\{0,\dots,9\}$ denote the digit label. The first classifier $h_{\phi}$ is a ResNet-18 \citep{resnet} pretrained on ImageNet-1K and fine-tuned on MRBI to maximise classification accuracy; it is trained with the cross-entropy loss:
\begin{equation}
    \mathcal{L}_{\mathrm{CE}}(\phi)=\mathbb{E}_{(x,y)\sim\mathcal{D}_B}\big[-\log h_{\phi}(y\mid x)\big].
\end{equation}
This classifier evaluates structural preservation: because MRBI backgrounds are primarily high-frequency, the digit encodes the large-scale structure that must be preserved. We report $h_{\phi}$ accuracy on generated samples $f_{\theta}(\tilde{x}_A)$ and compare it to its accuracy on real MRBI images. 
The second classifier $d_{\psi}$ is a convolutional encoder-based network designed to assess whether generated samples belong to $\mathcal{D}_B$. 
We fix its architecture and hyperparameters by first training it to distinguish samples from $\Da$ and $\Db$; this initial tuning yields near-perfect accuracy and ensures that the classifier has sufficient capacity for high-frequency discrimination. 
Once architecture and hyperparameters are fixed, we train $d_{\psi}$ anew to discriminate generated samples $\mathcal{G}(f_\theta, \tilde{x}_A)$ from real $\Db$ samples using the logistic objective:
\begin{equation}
    \mathcal{L}_{\mathrm{disc}}(\psi)=-\mathbb{E}_{x\sim\mathcal{D}_B}[\log d_{\psi}(x)]-\mathbb{E}_{\hat{x}\sim \mathcal{G}(f_\theta,\txy^{-1}(\mathcal{B}_A^S \oplus \mathcal{B}_\epsilon^D))}[\log(1-d_{\psi}(\hat{x}))].
\end{equation}
During evaluation, an accuracy close to $50\%$ (chance level) indicates that generated samples are indistinguishable from examples from $\Db$. 
To avoid misleading conclusions due to overfitting, we keep the model/hyperparameter choices fixed after the initial tuning. We report accuracies from both classifiers in Table~\ref{tab:method_comparison}.
Together, these two metrics probe complementary aspects of reconstruction: preservation of large-scale semantic structure (digit identity) and realism of domain-specific fine-scale details.

We performed a small ablation study on the MRBI dataset, where low-resolution images were filtered using a cutoff frequency $w_c = 4$. We considered several scenarios:
\begin{enumerate}
    \item \textbf{Optimal case:} training and inference both performed on data filtered with $w_c = 4$.
    \item \textbf{Stronger inference filtering:} training on $w_c = 4$, inference on $w_c = 3$.
    \item \textbf{Stronger training filtering:} training on $w_c = 3$, inference on $w_c = 4$.
    \item \textbf{Stronger filtering both:} training and inference on $w_c = 3$.
\end{enumerate}

These settings allow us to assess the effect of mismatched or stronger low-frequency filtering during training and inference on both digit reconstruction and domain alignment.

\paragraph{Results.}
\begin{table}[htbp]
\centering
\caption{Baseline comparison for MRBI on Digits and Domain Classification. For the Digits classifier, performance on the true MRBI dataset is: average accuracy $96.94\%$ and average confidence $98.23\%$.}
\label{tab:method_comparison}
\resizebox{\textwidth}{!}{%
\begin{tabular}{lccc}
\toprule
\multirow{2}{*}{\textbf{Method}} &
\multicolumn{2}{c}{\textbf{Digits Classification} ($\uparrow$, $h_\phi$)} &
\textbf{Domain Classification} ($\downarrow$, $d_\psi$)\\
\cmidrule(lr){2-3}
 & Avg. Accuracy (\%) & Avg. Conf. (\%) & Accuracy (\%) \\
\midrule
Dual FM                                        & 35.85             & 85.44             & 0.90 \\
Diffusion Bridge $t^\star=0.6$              & 41.35             & 88.62             & 0.98 \\
Diffusion Bridge $t^\star=0.5$              & 36.78             & 88.71             & 0.97 \\
Diffusion Bridge $t^\star=0.4$              & 30.04             & 88.62             & 0.96 \\
SerpentFlow $w_c = 4$, inf. $w_c = 4$       & \textbf{88.63}    & \textbf{95.42}    & \textbf{0.50} \\
SerpentFlow $w_c = 4$, inf. $w_c = 3$       & 20.83             & 81.18             & 0.81 \\
SerpentFlow $w_c = 3$, inf. $w_c = 4$       & \underline{75.67} & \underline{91.97} & 0.60 \\
SerpentFlow $w_c = 3$, inf. $w_c = 3$       & 63.22             & 89.19             & \underline{0.53} \\
SerpentFlow conditional, $w_c = 4$                & 28.90             & 79.38             & 1.00 \\
\bottomrule
\end{tabular}%
}
\end{table}

Table~\ref{tab:method_comparison} clearly demonstrates that when the model is applied with the correct cutoff frequency $w_c$, the reconstructed digits are significantly more recognizable than with the other methods. The classifier trained on real MRBI digits achieves an accuracy of 88\% on samples generated by our optimal configuration, compared to 40\% for the best competing baseline, demonstrating a substantial improvement.
When the model is trained with overly aggressive frequency filtering but evaluated on full-resolution data, performance decreases but remains well above all baselines. This robustness is encouraging, as super-resolution in many real-world scenarios is an ill-posed problem, ie, the exact cutoff frequency separating domains is unknown. In contrast, when excessive filtering is applied at inference time, the results deteriorate drastically, indicating that preserving the appropriate range of frequencies at generation time is crucial.

\begin{figure}[htbp]
    \centering
    \includegraphics[width=0.99\linewidth]{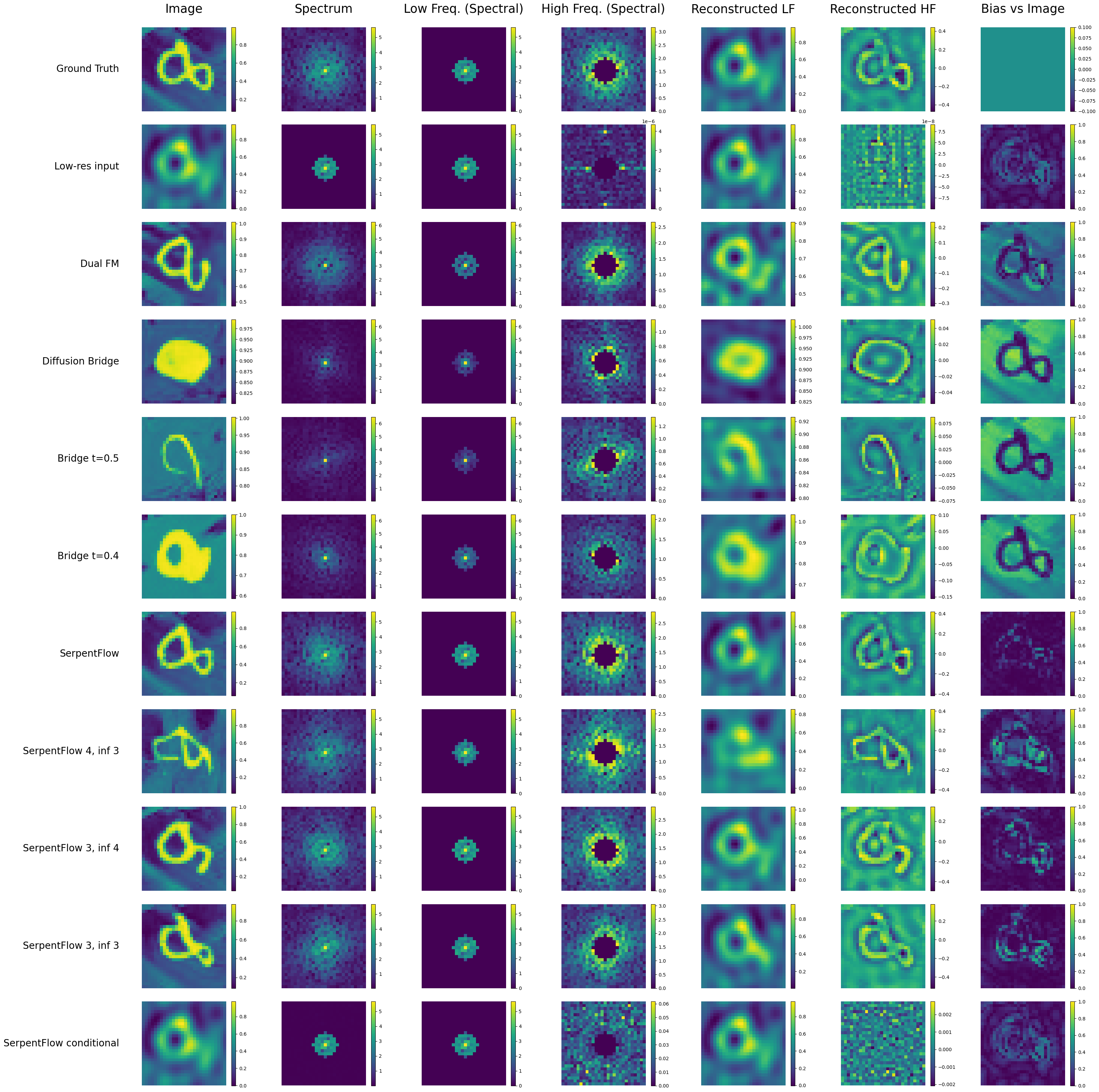}
    \caption{MRBI Reconstruction Comparison, for an image labeled 8, with $w_c=4$.}
    \label{fig:mrbi_results}
\end{figure}

Figure~\ref{fig:mrbi_results} further refines these observations. Qualitatively, our method preserves the low-frequency structure of the original images, unlike Dual FM or Diffusion Bridge, which tend to alter the global structure and sometimes even misgenerate the digit itself. Moreover, our approach successfully reconstructs the high-frequency background textures typical of the MRBI dataset, whereas Dual FM and Diffusion Bridge fail to do so. This property is particularly relevant in many applications where information is generally distributed across the entire spatial domain rather than concentrated in a localized region (as is often the case in natural image datasets). Being able to model both the foreground and background content accurately is therefore essential for downstream analysis. 

Finally, we experimented with a conditional variant of our approach, in which the input to the generator consisted of two channels: one containing white noise and the other the filtered data. This configuration performed significantly worse than the additive combination used in the other versions, suggesting that explicit conditioning is less effective than our implicit fusion strategy for reconstructing missing frequencies.

\paragraph{Effect of noise addition.}
To investigate the impact of stochasticity in high-frequency reconstruction, we generate multiple predictions per low-pass input by sampling independent Gaussian noise components $z_\epsilon^D$. For each input, five predictions are generated with non-zero noise, and a sixth prediction is produced with the noise set to zero. We visualize both the generated images and their deviations (bias) from the ground truth. 

\begin{figure}[htbp]
    \centering
    \begin{subfigure}[c]{0.49\linewidth}
        \centering
        \includegraphics[width=\linewidth]{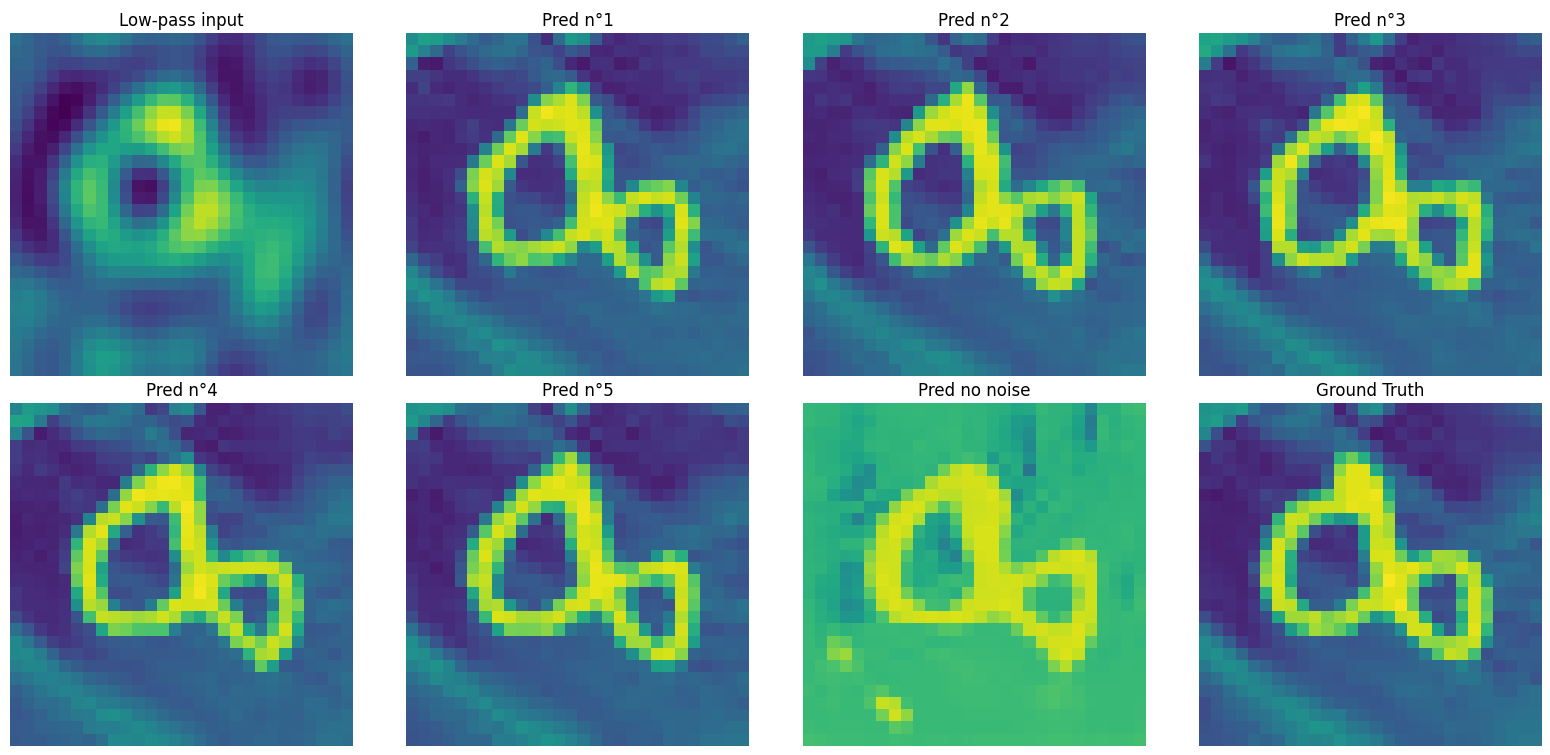}
        \caption{Model $w_c=4$, inference $w_c=4$}
        \label{fig:mrbi_4_4}
    \end{subfigure}
    \hfill
    \begin{subfigure}[c]{0.49\linewidth}
        \centering
        \includegraphics[width=\linewidth]{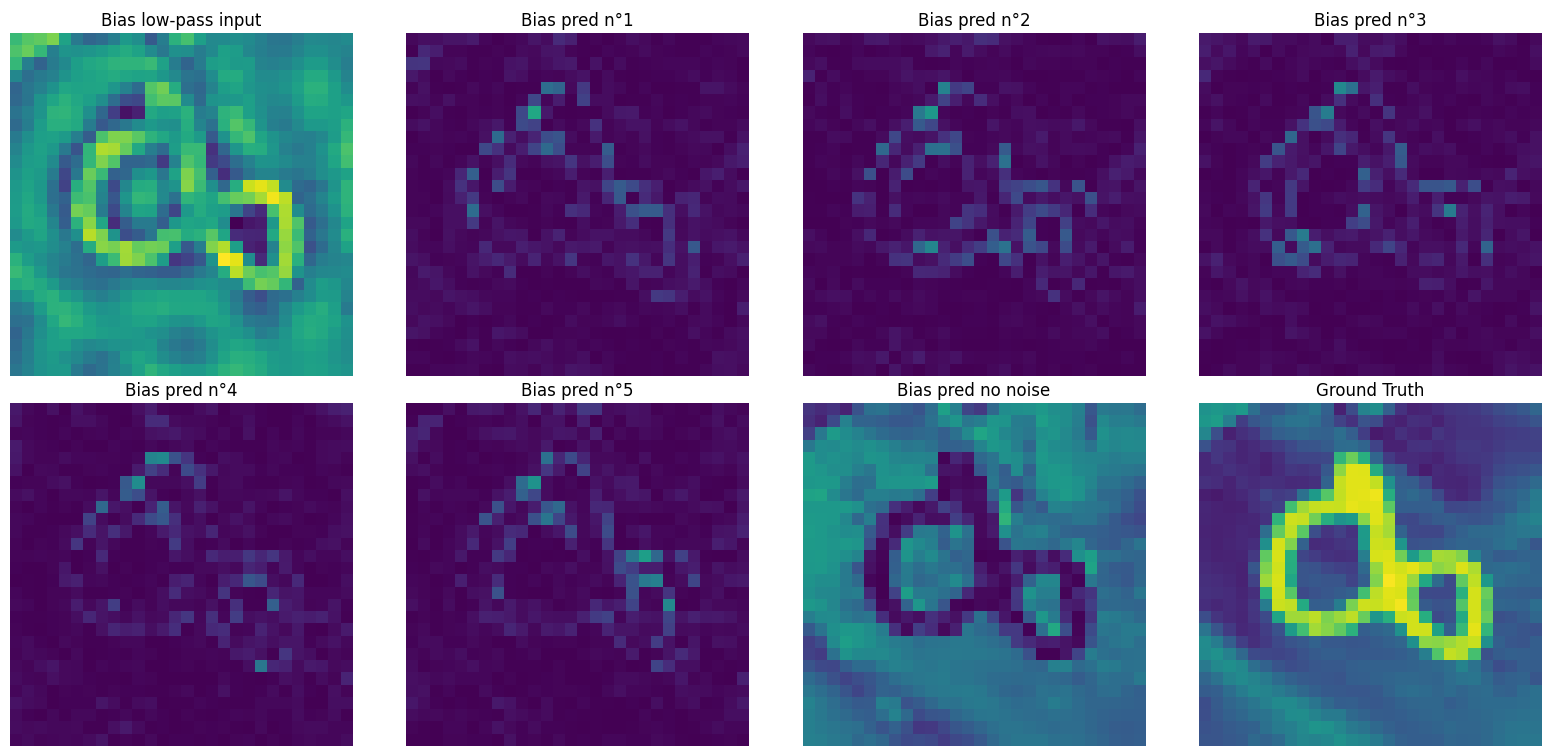}
        \caption{Bias maps, model $w_c=4$, inference $w_c=4$}
        \label{fig:mrbi_bias_4_4}
    \end{subfigure}

    \begin{subfigure}[c]{0.49\linewidth}
        \centering
        \includegraphics[width=\linewidth]{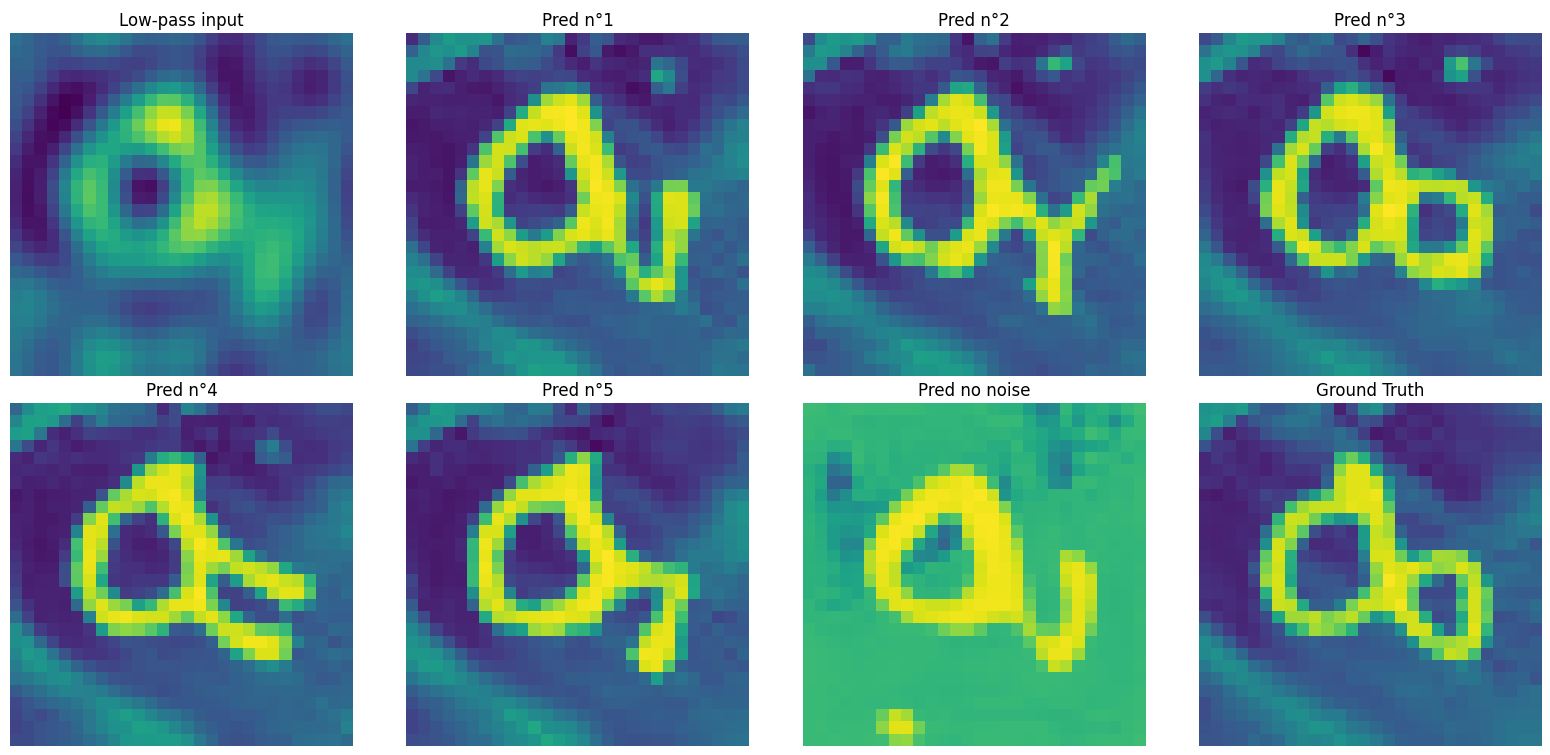}
        \caption{Model $w_c=3$, inference $w_c=4$}
        \label{fig:mrbi_3_4}
    \end{subfigure}
    \hfill
    \begin{subfigure}[c]{0.49\linewidth}
        \centering
        \includegraphics[width=\linewidth]{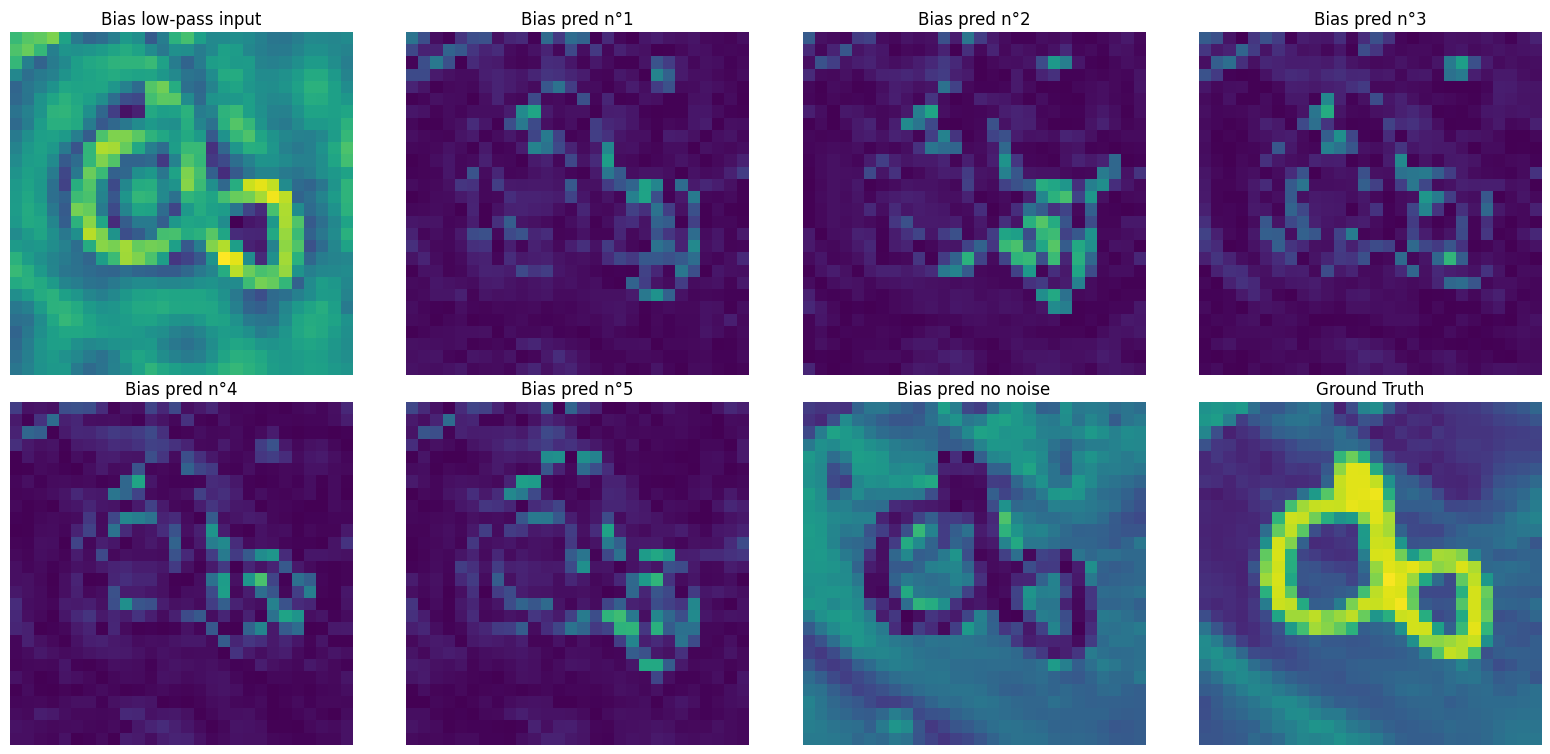}
        \caption{Bias maps, model $w_c=3$, inference $w_c=4$}
        \label{fig:mrbi_bias_3_4}
    \end{subfigure}
    \caption{Effect of stochastic high-frequency components in MRBI reconstruction. Top row: optimal configuration (model trained with $w_c=4$, inference done with $w_c=4$), bottom row: training with stronger low-frequency filtering: $w_c=3$, but inference done with $w_c=4$. Left: generated images, including multiple noisy predictions and one without noise. Right: corresponding deviations from the ground truth.}
    \label{fig:mrbi_noise}
\end{figure}

The addition of noise allows the model to explore a range of plausible high-frequency details, reflecting the fact that in real-world applications, the high-frequency content may not have a unique ground truth but rather a statistical distribution. In the optimal configuration (model trained with $w_c=4$ and inference with $w_c=4$, see Figures~\ref{fig:mrbi_4_4}~and~\ref{fig:mrbi_bias_4_4}), the generated images show little variation across noise realizations, and the bias maps indicate only minor deviations from the ground truth. For the case where the model is trained with $w_c=3$ and the inference is done with $w_c=4$ (see Figures~\ref{fig:mrbi_3_4}~and~\ref{fig:mrbi_bias_3_4}), the diversity of generated images is much higher: one prediction may closely resemble the ground truth, while others deviate more substantially. The noise-free predictions exhibit completely smoothed backgrounds, highlighting the importance of stochastic sampling for reconstructing realistic high-frequency details.

\subsection{Fluid simulation dataset}
\label{subsec:moisture}

We consider the fluid simulation dataset introduced by \cite{bischoff2024unpaired}, which is based on an idealized two-dimensional advection–condensation model \citep{ogorman2006}. The dataset consists of unpaired low-resolution $(64 \times 64)$ and high-resolution $(512 \times 512)$ fields representing two physical variables: vorticity and supersaturation. The high-resolution fields depend on the wavenumber $k_x=k_y \in \{2.0, 4.0, 8.0, 16.0\}$. Both the low-resolution dataset and each version of the high-resolution datasets contain around $2,000$ iterations.

\paragraph{Setup.} For each wavenumber $k$, we define two domains from these simulations:
\begin{itemize}
    \item $\mathcal{D}_A = \text{Low-resolution simulations}$ ;
    \item $\mathcal{D}_{B_k} = \text{High-resolution simulations with wavenumber $k$}.$
\end{itemize}

Importantly, $\mathcal{D}_A$ is not obtained by simply downsampling the high-resolution fields. Instead, it comes from coarse-resolution simulations that are upsampled using nearest-neighbor interpolation combined with a low-pass filter. Consequently, the spectral content of $\mathcal{D}_A$ approximately spans the low-frequency subspace of $\mathcal{D}_{B_k}$. This pre-existing setup naturally aligns with our method: the shared low-frequency structure can be extracted from $\mathcal{D}_A$ and enriched with domain-specific high-frequency components to generate realistic high-resolution samples in $\mathcal{D}_{B_k}$. Due to computational limitations, we worked at a reduced scale: the original $512 \times 512$ fields were averaged to $64 \times 64$ for both low- and high-resolution datasets. The wavenumber $k$ introduced into the vorticity spectrum creates a peak in frequency intensity proportional to $k$ that is not present in low-resolution data which is not parameterized by $k$. Thus, to choose the cutoff frequency $w_c$, we took the maximum radius that would cut this peak in frequency intensity in order to preserve as many low frequencies as possible. The dataset is also available for $k \in [0,1]$, but the peak is so close to the lowest frequency in the dataset that we would have to destroy almost everything to preserve it. In Appendix~\ref{app:fluids}, we show the results of our method when cutting less: the structure of the low-frequency data is lost, and when cutting more: the peak is not reconstructed.

\begin{figure}[htbp]
    \centering
    \includegraphics[width=0.98\linewidth]{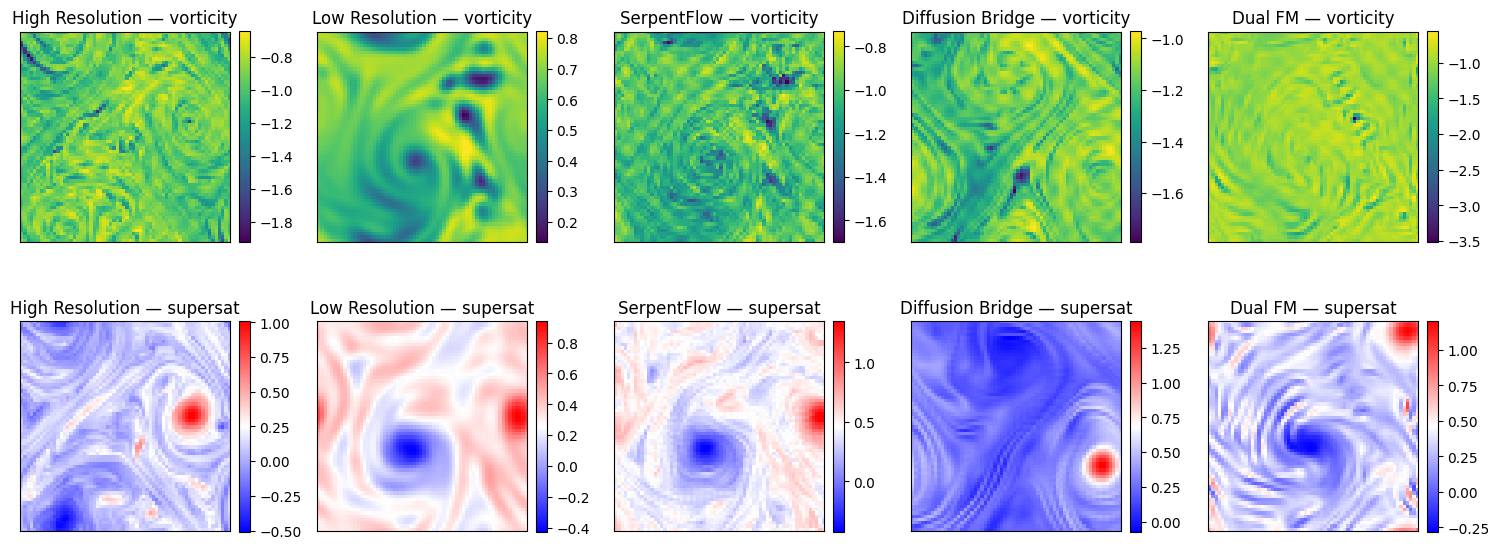}
    \caption{Vorticity and supersaturation fields at a given iteration for $\mathcal{D}_{B_8}$, $\Da$, and reconstructions from the baselines.}
    \label{fig:k8}
\end{figure}

\paragraph{Evaluation.} Figure~\ref{fig:k8} provides an initial overview of baseline performance for $k=8.0$. While all three baselines produce reconstructions that plausibly resemble high-resolution data ($\mathcal{D}_{B_8}$), SerpentFlow is, qualitatively on this example, the only method that consistently preserves the overall structure of the input signal for both vorticity and supersaturation.

To quantitatively analyze performance more precisely, we examined the temporal evolution of the reconstructions, as well as the distribution and spectral properties of the fields.

\begin{figure}[htbp]
    \centering
    \begin{subfigure}[c]{0.9\linewidth}
        \centering
        \includegraphics[width=\linewidth]{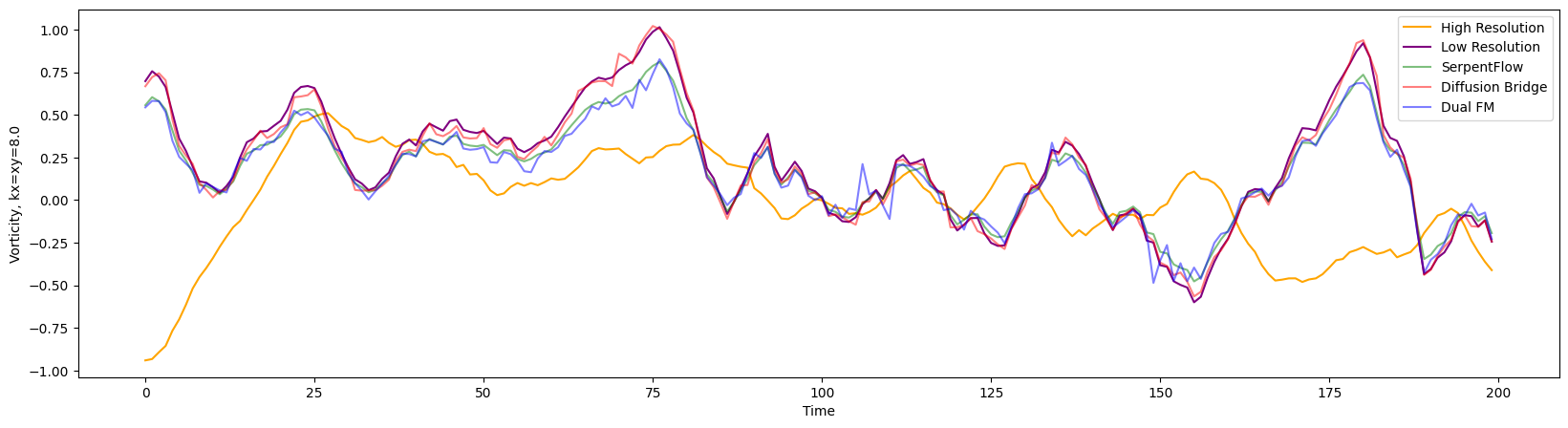}
        \caption{Vorticity temporal evolution for the first 200 samples ($k_x=k_y=8$).}
        \label{fig:temp_vorticity}
    \end{subfigure}
    \begin{subfigure}[c]{0.9\linewidth}
        \centering
        \includegraphics[width=\linewidth]{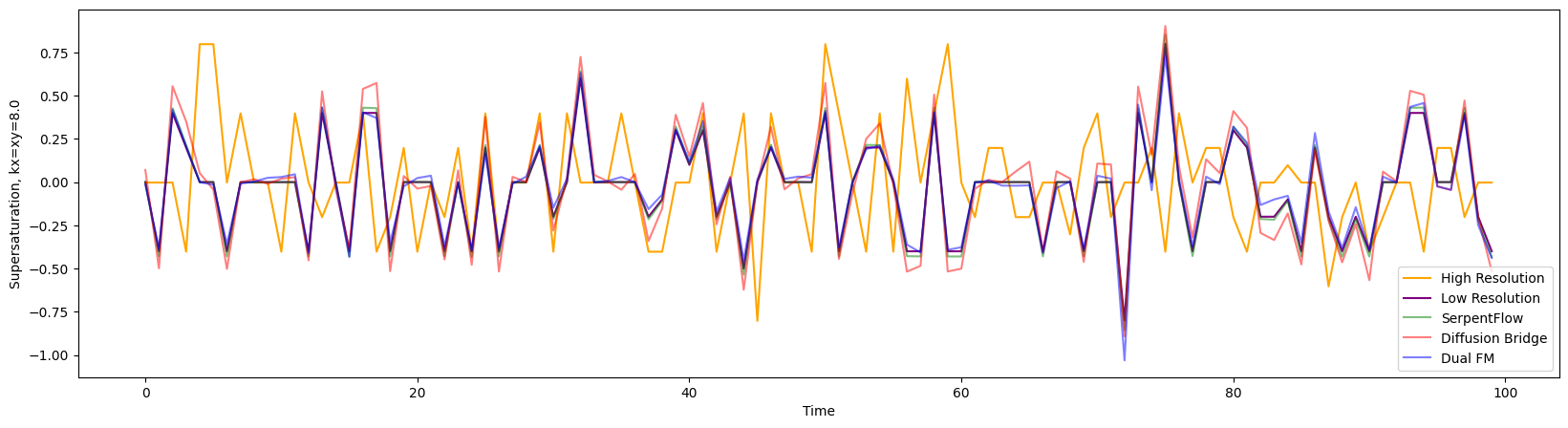}   
        \caption{Supersaturation temporal evolution for the first 100 samples ($k_x=k_y=8$).}
        \label{fig:temp_supersat}
    \end{subfigure}
    \caption{Spatially averaged temporal evolution of low- and high-resolution fields for $k_x=k_y=8$.}
    \label{fig:temp_evolution}
\end{figure}

As shown in Figure~\ref{fig:temp_evolution}, while all methods generally follow the temporal dynamics of the low-resolution data, our reconstructions remain closest to the original signal, confirming that our approach preserves temporal dynamics.

\begin{figure}[htbp]
    \centering
    \begin{subfigure}[c]{0.49\linewidth}
        \centering
        \includegraphics[width=\linewidth]{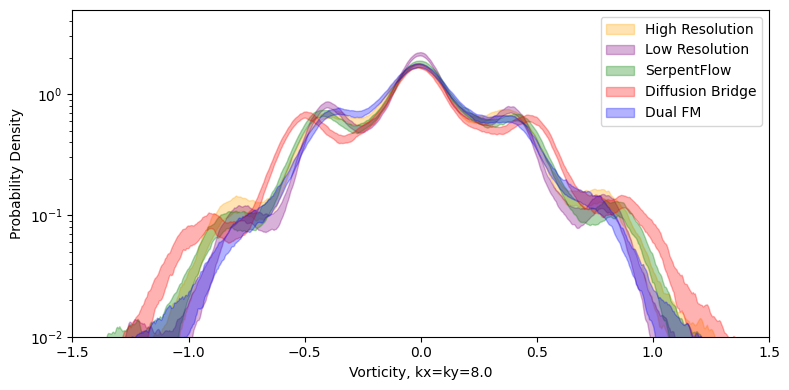}
        \caption{Vorticity field density estimation.}
        \label{fig:df_vorticity_k8}
    \end{subfigure}
    \begin{subfigure}[c]{0.49\linewidth}
        \centering
        \includegraphics[width=\linewidth]{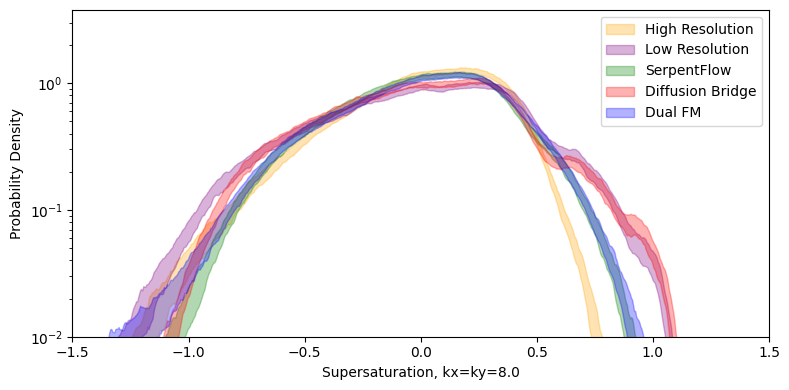}   
        \caption{Supersaturation field density estimation.}
        \label{fig:df_supersat_k8}
    \end{subfigure}
    \begin{subfigure}[c]{0.49\linewidth}
        \centering
        \includegraphics[width=\linewidth]{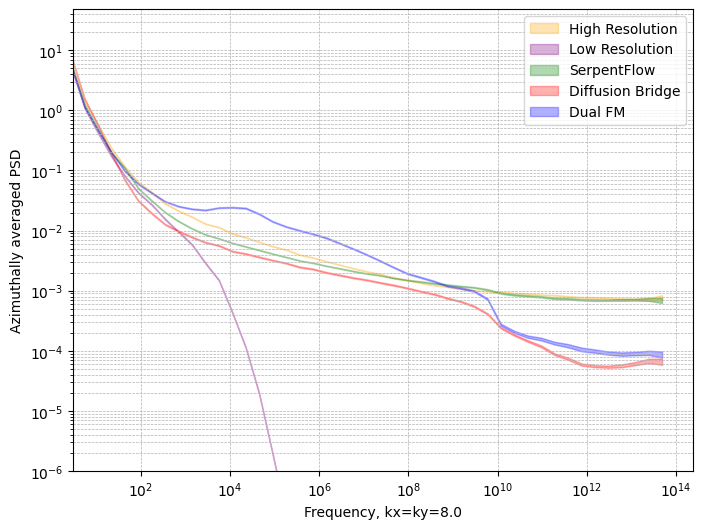}
        \caption{Vorticity field spectral density.}
        \label{fig:spec_vorticity_k8}
    \end{subfigure}
    \begin{subfigure}[c]{0.49\linewidth}
        \centering
        \includegraphics[width=\linewidth]{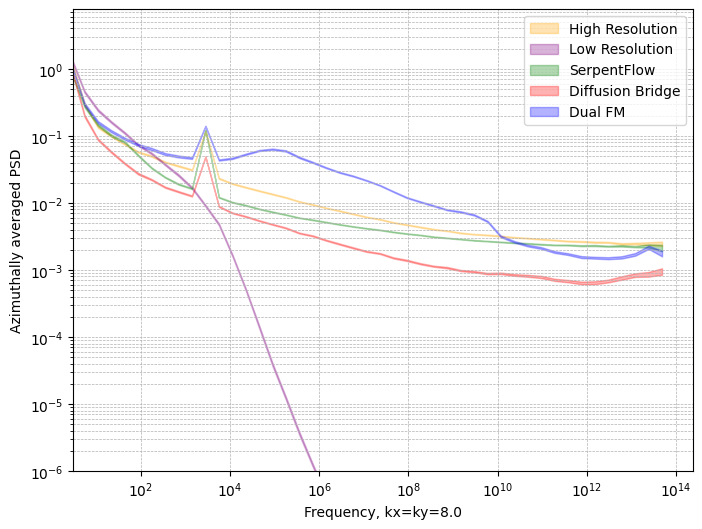}   
        \caption{Supersaturation field spectral density.}
        \label{fig:spec_supersat_k8}
    \end{subfigure}
    \caption{Density and spectral density comparisons for low- and high-resolution fields and baselines ($k_x=k_y=8$).}
    \label{fig:density_estimation_k8}
\end{figure}

Regarding the transfer to $\mathcal{D}_{B_8}$, Figure~\ref{fig:density_estimation_k8} shows how closely each baseline reproduces high-resolution statistics. Figures~\ref{fig:df_vorticity_k8} and~\ref{fig:df_supersat_k8} indicate that the distributions of the Dual FM and our reconstructions closely match the high-resolution fields, whereas the Bridge Matching baseline produces tails that are too heavy. In terms of spectral density (Figures~\ref{fig:spec_vorticity_k8} and~\ref{fig:spec_supersat_k8}), all methods recover high-frequency content comparable to the high-resolution fields. However, for vorticity, Diffusion Bridge and Dual FM generate some artifacts, whereas our approach faithfully follows the high-resolution spectrum. For supersaturation, our method still improves upon low-resolution inputs, though some discrepancies remain in the high-frequency range.

\subsection{Climate downscaling: CMIP6 to ERA5 wind fields}
\label{subsec:downscaling}

\paragraph{Data presentation.}
A General Circulation Model (GCM) simulates climate variables while reproducing their key statistical properties, such as dominant modes, variability, and the frequency or return periods of extreme events. GCMs are particularly useful for generating spatially and temporally coherent large-scale climate statistics; however, their computational cost prevents them from being run at fine spatial resolutions. To obtain useful information at regional or local scales, downscaling techniques are therefore required. As GCMs do incorporate key statistical properties, the goal of downscaling is not to predict exact local values, but rather to downscale their statistical properties. Approaches designed for this purpose are referred to as probabilistic downscaling methods (PDMs). A PDM is evaluated against observations, which should be interpreted as one possible realization of the GCM dynamics rather than an absolute ``true'' state.

Here, we consider one GCM, the ACCESS Earth System Model \citep{access}, from the Coupled Model Intercomparison Project Phase 6 (CMIP6 \citep{cmip6}), which provides climate simulations at daily temporal resolution and coarse spatial resolution. As observational reference, we use ERA5 reanalysis data \citep{hersbach2020era5}, which are available at higher spatial resolution and hourly frequency. To ensure temporal consistency with the GCM, we compute daily averages of the ERA5 fields. In this study, we focus exclusively on the daily wind intensity over France. Our goal is to downscale the GCM outputs spatially from their native resolution ($1.875 \times 1.25$ degrees) to the ERA5 grid ($0.25 \times 0.25$ degrees), while maintaining one value per day.

\paragraph{Experimental setup.}
Although the GCM and ERA5 data represent the same physical quantity, they are not aligned at the grid scale. By construction, the GCM exhibits stronger spatial correlations and smoother fields than the observational ERA5 data, even when both are represented at the same resolution. This means that while the large-scale, low-frequency structures are largely shared between the two domains, the small-scale, high-frequency variations differ and are specific to each domain. Our low-frequency / high-frequency decomposition is particularly well-suited to this setting: the low-frequency backbone can be extracted from the GCM and preserved. In contrast, the high-frequency component can be stochastically reconstructed to match the variability observed in ERA5. This allows us to perform downscaling without requiring explicit paired data, and ensures that generated high-resolution wind fields are both physically coherent and statistically consistent with ERA5 observations. Data from 1981 to 2001 are used to train the model, while the validation data covers the years 2002 to 2022, ensuring a validation period of twenty years to compute climate statistics. We trained the classifier mentioned in Section~\ref{method_cutoff} to select $w_c^\star$. The classifier returned a value of $1$, allowing effects of spatial scales greater than approximately 1200 km to be retained as low frequencies. A further discussion on the classifier is given in Appendix~\ref{app:cutoff_classifier}.

\begin{figure}[htbp]
    \centering
    \begin{subfigure}[c]{0.98\linewidth}
        \centering
        \includegraphics[width=\linewidth]{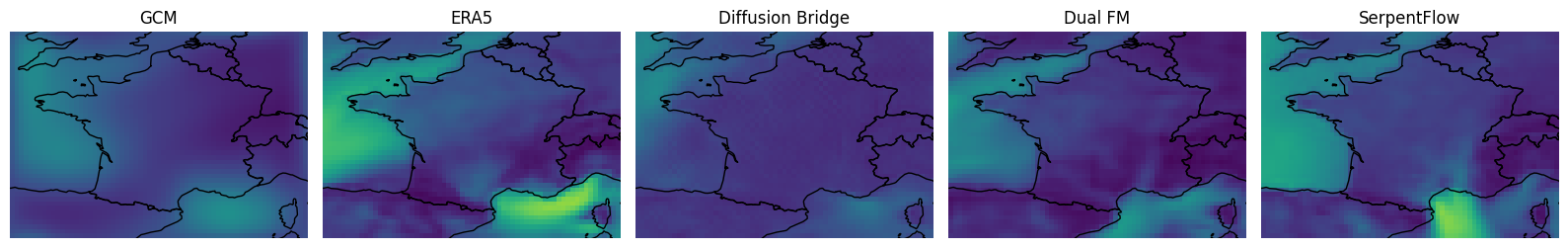}
        \caption{Date: 2004-04-11.}
        \label{fig:climate_i_100}
    \end{subfigure}
    \begin{subfigure}[c]{0.98\linewidth}
        \centering
        \includegraphics[width=\linewidth]{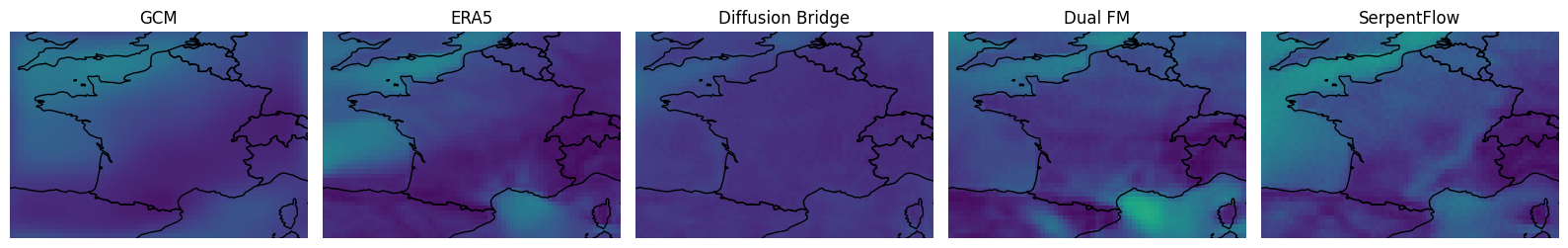}
        \caption{Date: 2011-08-04.}
        \label{fig:climate_i_3500}
    \end{subfigure}
    \begin{subfigure}[c]{0.98\linewidth}
        \centering
        \includegraphics[width=\linewidth]{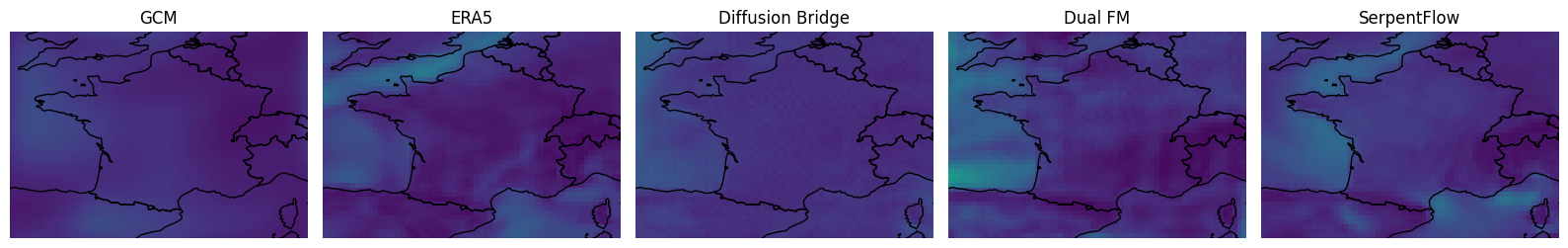}
        \caption{Date: 2018-06-10.}
        \label{fig:climate_i_6000}
    \end{subfigure}
    \caption{Downscaling outputs for various days within the test set, for our baselines.}
    \label{fig:climate_i}
\end{figure}
\paragraph{Evaluation.} Figure~\ref{fig:climate_i} gives a first visual overview of the baselines' downscaling performance. While Dual FM and SerpentFlow seem to have refined the wind field of the GCM, Diffusion Bridge seems to have further smoothed out the already highly correlated data from the climate model. 

\begin{figure}[htbp]
    \centering
    \begin{subfigure}[c]{0.98\linewidth}
        \centering
        \includegraphics[width=\linewidth]{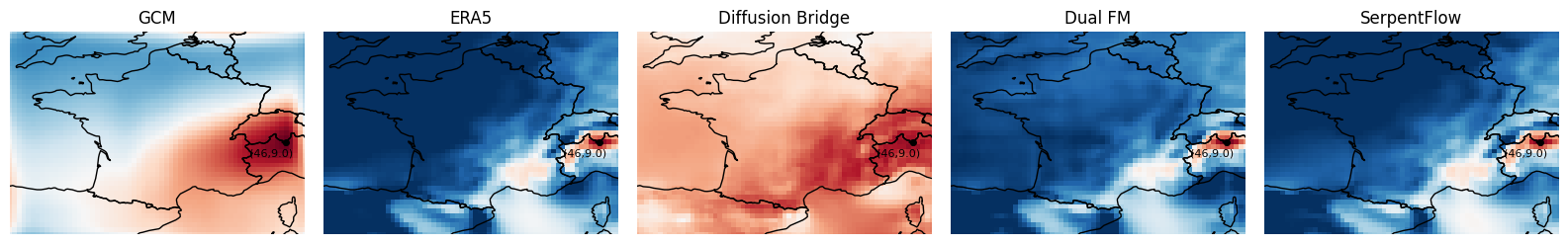}
        \caption{Correlation map, coordinate (46,9).}
        \label{fig:corr_alps}
    \end{subfigure}
    \begin{subfigure}[c]{0.98\linewidth}
        \centering
        \includegraphics[width=\linewidth]{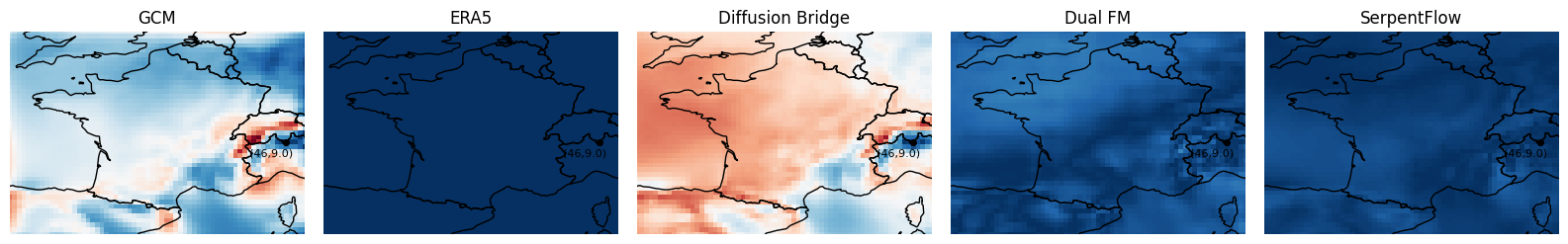}
        \caption{Bias correlation map w.r.t. ERA5, coordinate (46,9).}
        \label{fig:corr_bias_alps}
    \end{subfigure}
    \begin{subfigure}[c]{0.98\linewidth}
        \centering
        \includegraphics[width=\linewidth]{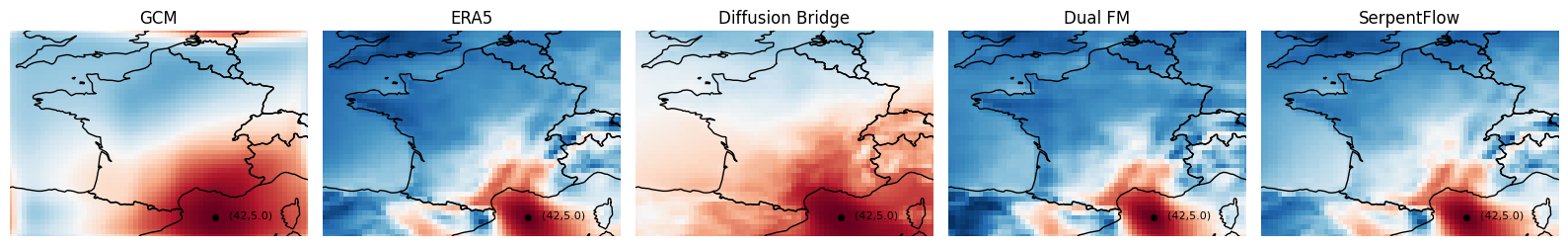}
        \caption{Correlation map, coordinate (42,5).}
        \label{fig:corr_med}
    \end{subfigure}
    \begin{subfigure}[c]{0.98\linewidth}
        \centering
        \includegraphics[width=\linewidth]{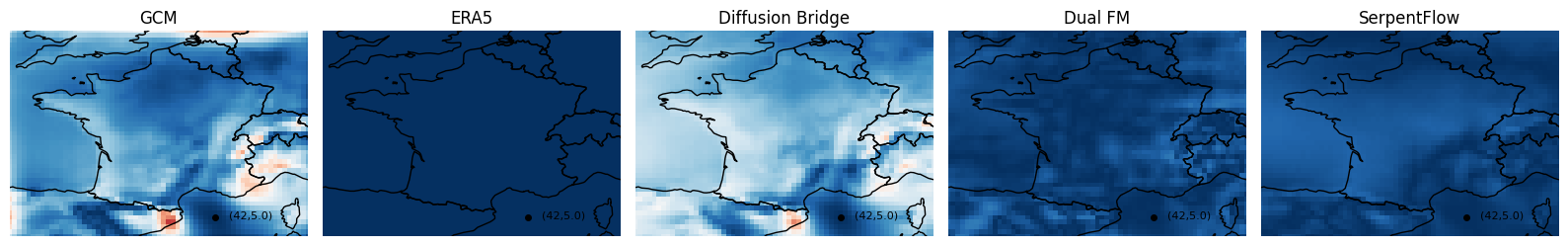}
        \caption{Bias correlation map w.r.t. ERA5, coordinate (42,5).}
        \label{fig:corr_bias_med}
    \end{subfigure}
    \caption{Correlation maps of wind intensity fields. We select one point of interest, and compute correlations between the corresponding time series over the period compared to the time series of the other grid points. For each method, we compare the correlation map with that from ERA5. In Figures~\ref{fig:corr_alps} and~\ref{fig:corr_bias_alps}, the reference grid point is in the Alps. For Figures~\ref{fig:corr_med} and~\ref{fig:corr_bias_med}, the reference grid point is in the Mediterannean sea.}
    \label{fig:climate_corr}
\end{figure}
To assess the degree of spatial refinement achieved by our downscaling, we compared correlation maps computed over the validation period with those of ERA5. For a reference grid point $(\text{lat},\text{lon})$, the correlation map is generated by correlating its 20-year temporal series with those of all other points in the dataset. Figures~\ref{fig:corr_alps} and~\ref{fig:corr_med} show example correlation maps for a point in the Alps and one in the Mediterranean, respectively, while Figures~\ref{fig:corr_bias_alps} and~\ref{fig:corr_bias_med} display the corresponding biases relative to ERA5.  
The difference between ERA5 and the GCM is striking: ERA5 maps clearly reflect topography, whereas the GCM fields are overly correlated and smooth, making these maps a strong discriminator for downscaling methods. Both Dual FM and our SerpentFlow produce correlation maps that closely resemble ERA5, capturing the terrain-induced spatial structure, while the Diffusion Bridge outputs fail to reproduce it. In the bias maps, particularly for the Mediterranean point, Dual FM appears slightly closer to ERA5 than SerpentFlow.  
By averaging the correlation biases across all grid points, we obtain an overall correlation score reported in Table~\ref{tab:climate_res}. Dual FM achieves a score 15\% better than SerpentFlow, confirming its superior ability to reconstruct the spatial variability of the target domain $\mathcal{D}_B$.

\begin{figure}[htbp]
    \centering
    \begin{subfigure}[c]{0.55\linewidth}
        \centering
        \includegraphics[width=\linewidth]{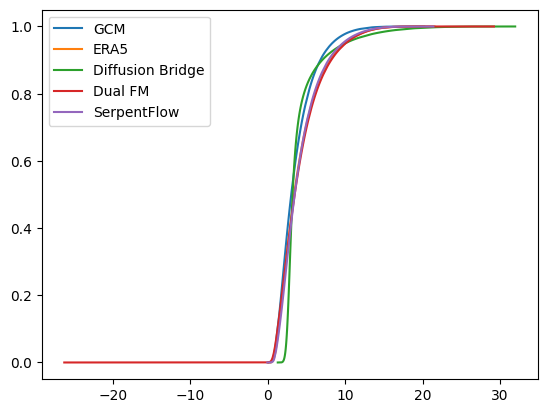}
        \caption{Cumulative Distribution Function (CDF).}
        \label{fig:climate_cdf_full}
    \end{subfigure}
    \begin{subfigure}[c]{0.49\linewidth}
        \centering
        \includegraphics[width=\linewidth]{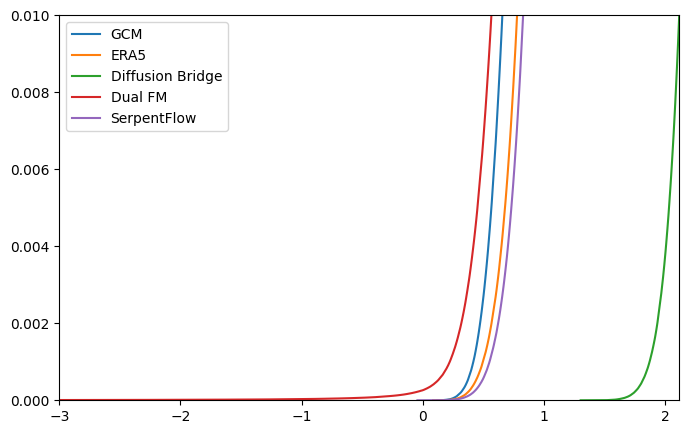}
        \caption{Zoom quantiles $<0.01$.}
        \label{fig:climate_cdf_01}
    \end{subfigure}
    \begin{subfigure}[c]{0.49\linewidth}
        \centering
        \includegraphics[width=\linewidth]{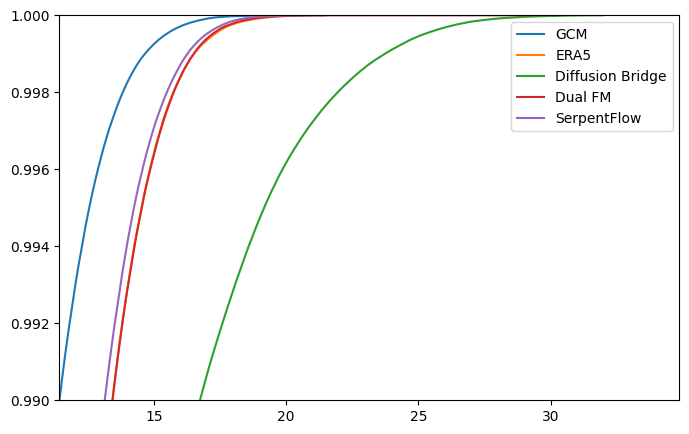}
        \caption{Zoom quantile $>0.99$.}
        \label{fig:climate_cdf_09}
    \end{subfigure}
    \caption{Cumulative distribution function of wind intensity of the baselines compared to the GCM and ERA5.}
    \label{fig:climate_cdf}
\end{figure}
We also observe the strong reconstruction performance of the Dual FM and SerpentFlow methods in Figure~\ref{fig:climate_cdf}, which shows the cumulative distribution functions (CDFs) of the baseline methods compared to ERA5 and the GCM. Recall that the CDF $F(x)$ of a random variable $X$ is defined as the probability that $X$ takes a value less than or equal to $x$, i.e., $F(x) = \mathbb{P}(X \le x)$. Figure~\ref{fig:climate_cdf_full} illustrates that both Dual FM and SerpentFlow successfully adjust the GCM distribution to approach that of ERA5, with their CDFs nearly overlapping. In contrast, Diffusion Bridge deviates further from the target distribution.
Examining the lower tail (Figure~\ref{fig:climate_cdf_01}), Dual FM produces a small number of negative artefacts, whereas SerpentFlow aligns more closely with ERA5. In the extreme upper tail (Figure~\ref{fig:climate_cdf_09}), SerpentFlow slightly underperforms, while Dual FM remains closely aligned with ERA5. This difference in performance is captured by the Kolmogorov–Smirnov (KS) score reported in Table~\ref{tab:climate_res}, which measures the maximum distance between two CDFs. Dual FM achieves a slightly lower KS loss than SerpentFlow, reflecting its better overall reconstruction of the target distribution.

\begin{figure}[htbp]
    \centering
    \begin{subfigure}[c]{0.7\linewidth}
        \centering
        \includegraphics[width=\linewidth]{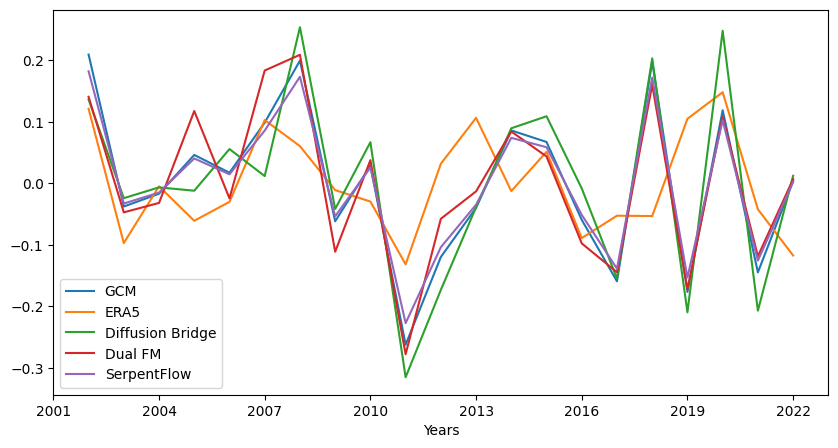}
        \caption{Normalized Yearly mean.}
        \label{fig:climate_yearly_mean}
    \end{subfigure}
    \begin{subfigure}[c]{0.98\linewidth}
        \centering
        \includegraphics[width=\linewidth]{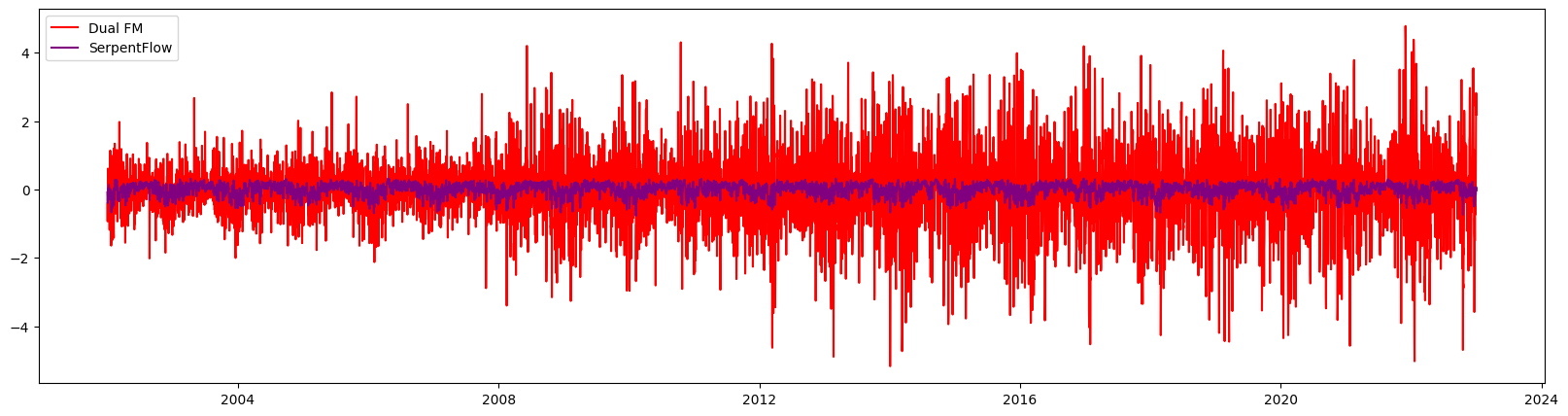}
        \caption{Temporal bias w.r.t. normalized spatially averaged GCM}
        \label{fig:climate_ts_bias}
    \end{subfigure}
    \caption{Illustration of the temporal consistency of baselines compared to climate model dynamics.}
    \label{fig:climate_temporal}
\end{figure}
So far, we have only evaluated the baselines on their ability to reconstruct data belonging to $\mathcal{D}_B$, without assessing their capacity to preserve the structure of $\mathcal{D}_A$. While this preservation may not be immediately visible by inspection, Figure~\ref{fig:climate_i} allows us to compare the spatially averaged and normalized temporal dynamics of the GCM with those of the baselines. In Figure~\ref{fig:climate_yearly_mean}, we show the normalized spatial and annual mean of the baselines, ERA5, and the GCM, capturing inter-annual variability. The goal is to follow the GCM signal as closely as possible. We observe that Diffusion Bridge and Dual FM struggle to track the signal, particularly around the year 2005, whereas SerpentFlow follows it almost perfectly.

This difference is even more striking when examining the daily signal. Figure~\ref{fig:climate_ts_bias} shows the bias between the original GCM signal and the values reconstructed by Dual FM and SerpentFlow. It is clear that, for all timesteps, Dual FM produces data substantially more distant from the original signal than SerpentFlow. We quantify this discrepancy by computing the RMSE between these signals, reported in Table~\ref{tab:climate_res}; on this metric, SerpentFlow achieves an order-of-magnitude lower error than Dual FM.
In addition to RMSE, we also report the Nash–Sutcliffe efficiency index (NSE, \citep{nash1970river}), a standard metric in geophysical modeling used to assess how well a reconstruction reproduces the temporal variability of a reference series relative to its climatological mean. An NSE of $1$ indicates a perfect reconstruction, whereas values close to $0$ or negative imply performance no better — or worse — than replacing the series with its temporal mean. On this metric as well, SerpentFlow performs markedly better, improving from $0.29$ (Dual FM) to $0.98$.
Therefore, although Dual FM is slightly better than SerpentFlow at reconstructing data in $\mathcal{D}_B$, it does so at the cost of disrupting the original climatic structure. Preserving the GCM dynamics is crucial in univariate downscaling applications, as these dynamics represent the primary scientific value of such simulations.

\begin{table}[htbp]
\centering
\caption{
Evaluation of downscaling performance on the climate dataset.
KS: Kolmogorov–Smirnov statistic between generated and target distributions;
Correlation Score: mean correlation bias relative to ERA5;
Temporal RMSE: root mean square error of the spatially-averaged daily wind intensity over the validation period;
NSE: Nash–Sutcliffe efficiency index computed on the spatially-averaged daily series (1 = perfect, 0 = mean predictor).
}
\label{tab:climate_res}
\begin{tabular}{lcccc}
\toprule
\textbf{Method} & \textbf{KS} ($\downarrow$) & \textbf{Corr. Score} ($\downarrow$) & \textbf{Temporal RMSE} ($\downarrow$) & \textbf{NSE} ($\uparrow$)\\
\midrule
Diffusion Bridge & 0.25119             & 0.293             & 1.779             & -0.029 \\
Dual FM          & \textbf{0.02495}    & \textbf{0.045}    & \underline{1.226} & \underline{0.291} \\
SerpentFlow      & \underline{0.02525}    & \underline{0.052} & \textbf{0.031}    & \textbf{0.982} \\
\bottomrule
\end{tabular}
\end{table}

\section{Conclusion}
In this work, we introduced SerpentFlow (SharEd-structuRe decomPosition for gEnerative domaiN adapTation), a generative framework for unpaired domain alignment that leverages a principled decomposition of data into shared and domain-specific components. By isolating the shared structure and replacing domain-specific content with stochastic variations, SerpentFlow generates pseudo-pairs that enable the use of conditional generative models even in the absence of real paired data. The decomposition is fully adaptive and data-driven, making it applicable across diverse types of domains without requiring prior knowledge of how structures should be separated.

We instantiated SerpentFlow on unsupervised super-resolution tasks using three diverse datasets: synthetic images (MRBI), simulated physical processes (fluid simulations), and climate downscaling (CMIP6 to ERA5 wind fields). In these tasks, the shared component corresponds to low-frequency content, while high-frequency details capture domain-specific variability. The cutoff frequency separating low- and high-frequency components is determined automatically from the dataset using a classifier-based criterion, ensuring a fully automated and robust decomposition. By preserving shared structures and modeling domain-specific variations stochastically, SerpentFlow successfully reconstructs domain-specific features while maintaining consistency with the shared patterns. Across all experiments, it outperforms existing unpaired generative methods in fidelity, statistical consistency, and physical interpretability.

Beyond these experiments, SerpentFlow provides a broadly applicable framework. The shared–vs.–specific decomposition could be extended to temporal or spatio-temporal domains, enabling high-resolution reconstructions for time series, videos, or other structured signals. Importantly, when no clear prior knowledge exists about how the decomposition should be performed, the use of SerpentFlow will not be effective (see Appendix~\ref{app:mnist_failure}) and state-of-the-art approaches such as Dual FM remain strong baselines.

Future work could explore other $\txy$ functions, such as blurring or noising strategies, as well as alternative decomposition strategies, such as wavelet-based multi-scale analysis or data-driven approaches including PCA, independent component analysis, or learned embeddings, potentially improving flexibility and interpretability. Overall, SerpentFlow offers a robust, fully automated, and general-purpose pipeline for unpaired domain alignment, combining adaptive shared–specific decomposition with conditional generative modeling to bridge diverse domains while respecting both global structure and local variability.

%Bibliography
\bibliographystyle{unsrt}  
\bibliography{references}  

\begin{thebibliography}{10}

\bibitem{normalizingflows}
George Papamakarios, Eric Nalisnick, Danilo~Jimenez Rezende, Shakir Mohamed,
  and Balaji Lakshminarayanan.
\newblock Normalizing flows for probabilistic modeling and inference.
\newblock {\em Journal of Machine Learning Research}, 22(57):1--64, 2021.

\bibitem{autoencoder}
Diederik~P Kingma and Max Welling.
\newblock Auto-encoding variational bayes.
\newblock {\em arXiv preprint arXiv:1312.6114}, 2013.

\bibitem{gan}
Ian~J Goodfellow, Jean Pouget-Abadie, Mehdi Mirza, Bing Xu, David Warde-Farley,
  Sherjil Ozair, Aaron Courville, and Yoshua Bengio.
\newblock Generative adversarial nets.
\newblock {\em Advances in neural information processing systems}, 27, 2014.

\bibitem{ddpm}
Jonathan Ho, Ajay Jain, and Pieter Abbeel.
\newblock Denoising diffusion probabilistic models.
\newblock {\em Advances in neural information processing systems},
  33:6840--6851, 2020.

\bibitem{score}
Yang Song, Jascha Sohl-Dickstein, Diederik~P Kingma, Abhishek Kumar, Stefano
  Ermon, and Ben Poole.
\newblock Score-based generative modeling through stochastic differential
  equations.
\newblock In {\em International Conference on Learning Representations}, 2021.

\bibitem{flowmatching}
Yaron Lipman, Ricky~TQ Chen, Heli Ben-Hamu, Maximilian Nickel, and Matthew Le.
\newblock Flow matching for generative modeling.
\newblock In {\em The Eleventh International Conference on Learning
  Representations}, 2023.

\bibitem{kong2020diffwave}
Zhiyao Kong, Wei Ping, Jing Huang, Kexin Zhao, and Bryan Catanzaro.
\newblock Diffwave: A versatile diffusion model for audio synthesis.
\newblock In {\em International Conference on Learning Representations (ICLR)},
  2021.

\bibitem{ingraham2019generative}
John Ingraham, Vijay Garg, Regina Barzilay, and Tommi Jaakkola.
\newblock Generative models for graph-based protein design.
\newblock In {\em Advances in Neural Information Processing Systems},
  volume~32, 2019.

\bibitem{jing2022learning}
Bowen Jing, Stefan Eismann, Prafulla Suriana, Richard Townshend, and Ron Dror.
\newblock Learning functional protein sequences with diffusion models.
\newblock {\em arXiv preprint arXiv:2210.09238}, 2022.

\bibitem{conditionnalgans}
Mehdi Mirza and Simon Osindero.
\newblock Conditional generative adversarial nets.
\newblock {\em arXiv preprint arXiv:1411.1784}, 2014.

\bibitem{conditionaldiffusion}
Lvmin Zhang, Anyi Rao, and Maneesh Agrawala.
\newblock Adding conditional control to text-to-image diffusion models.
\newblock In {\em Proceedings of the IEEE/CVF international conference on
  computer vision}, pages 3836--3847, 2023.

\bibitem{cyclegan}
Jun-Yan Zhu, Taesung Park, Phillip Isola, and Alexei~A Efros.
\newblock Unpaired image-to-image translation using cycle-consistent
  adversarial networks.
\newblock In {\em Proceedings of the IEEE international conference on computer
  vision}, pages 2223--2232, 2017.

\bibitem{munit}
Ming-Yu Liu, Thomas Breuel, and Jan Kautz.
\newblock Unsupervised image-to-image translation networks.
\newblock {\em Advances in neural information processing systems}, 30, 2017.

\bibitem{alignflow}
Aditya Grover, Christopher Chute, Rui Shu, Zhangjie Cao, and Stefano Ermon.
\newblock Alignflow: Cycle consistent learning from multiple domains via
  normalizing flows.
\newblock In {\em Proceedings of the AAAI Conference on Artificial
  Intelligence}, volume~34, pages 4028--4035, 2020.

\bibitem{dualimplicitbridge}
Xuan Su, Jiaming Song, Chenlin Meng, and Stefano Ermon.
\newblock Dual diffusion implicit bridges for image-to-image translation.
\newblock In {\em The Eleventh International Conference on Learning
  Representations}, 2023.

\bibitem{stochasticinterpolants}
Michael~S Albergo, Nicholas~M Boffi, and Eric Vanden-Eijnden.
\newblock Stochastic interpolants: A unifying framework for flows and
  diffusions.
\newblock {\em arXiv preprint arXiv:2303.08797}, 2023.

\bibitem{bridgematching}
Yuyang Shi, Valentin De~Bortoli, Andrew Campbell, and Arnaud Doucet.
\newblock Diffusion schr{\"o}dinger bridge matching.
\newblock {\em Advances in Neural Information Processing Systems},
  36:62183--62223, 2023.

\bibitem{schrodingerflow}
Valentin De~Bortoli, Iryna Korshunova, Andriy Mnih, and Arnaud Doucet.
\newblock Schrodinger bridge flow for unpaired data translation.
\newblock {\em Advances in Neural Information Processing Systems},
  37:103384--103441, 2024.

\bibitem{neuralode}
Ricky~TQ Chen, Yulia Rubanova, Jesse Bettencourt, and David~K Duvenaud.
\newblock Neural ordinary differential equations.
\newblock {\em Advances in neural information processing systems}, 31, 2018.

\bibitem{styletransfer}
Leon~A Gatys, Alexander~S Ecker, and Matthias Bethge.
\newblock Image style transfer using convolutional neural networks.
\newblock In {\em Proceedings of the IEEE conference on computer vision and
  pattern recognition}, pages 2414--2423, 2016.

\bibitem{stylegan}
Tero Karras, Samuli Laine, and Timo Aila.
\newblock A style-based generator architecture for generative adversarial
  networks.
\newblock In {\em Proceedings of the IEEE/CVF conference on computer vision and
  pattern recognition}, pages 4401--4410, 2019.

\bibitem{unsupervisedsr}
Yuan Yuan, Siyuan Liu, Jiawei Zhang, Yongbing Zhang, Chao Dong, and Liang Lin.
\newblock Unsupervised image super-resolution using cycle-in-cycle generative
  adversarial networks.
\newblock In {\em Proceedings of the IEEE conference on computer vision and
  pattern recognition workshops}, pages 701--710, 2018.

\bibitem{freqreconstruction}
Emmanuel~J Cand{\`e}s, Justin Romberg, and Terence Tao.
\newblock Robust uncertainty principles: Exact signal reconstruction from
  highly incomplete frequency information.
\newblock {\em IEEE Transactions on information theory}, 52(2):489--509, 2006.

\bibitem{compressedsensing}
Ashish Bora, Ajil Jalal, Eric Price, and Alexandros~G Dimakis.
\newblock Compressed sensing using generative models.
\newblock In {\em International conference on machine learning}, pages
  537--546. PMLR, 2017.

\bibitem{climalign}
Brian Groenke, Luke Madaus, and Claire Monteleoni.
\newblock Climalign: Unsupervised statistical downscaling of climate variables
  via normalizing flows.
\newblock In {\em Proceedings of the 10th International Conference on Climate
  Informatics}, pages 60--66, 2020.

\bibitem{bischoff2024unpaired}
Tobias Bischoff and Katherine Deck.
\newblock Unpaired downscaling of fluid flows with diffusion bridges.
\newblock {\em Artificial Intelligence for the Earth Systems}, 3(2):e230039,
  2024.

\bibitem{rectifiedflow}
Xingchao Liu, Chengyue Gong, and Qiang Liu.
\newblock Flow straight and fast: Learning to generate and transfer data with
  rectified flow.
\newblock In {\em The Eleventh International Conference on Learning
  Representations (ICLR)}, 2023.

\bibitem{boyd2001fourier}
John~P Boyd.
\newblock {\em Chebyshev and Fourier Spectral Methods}.
\newblock Dover, 2001.

\bibitem{canuto2006spectral}
Claudio Canuto, M~Yousuff Hussaini, Alfio Quarteroni, and Thomas~A Zang.
\newblock {\em Spectral Methods: Fundamentals in Single Domains}.
\newblock Springer, 2006.

\bibitem{falck2025fourier}
Fabian Falck, Teodora Pandeva, Kiarash Zahirnia, Rachel Lawrence, Richard
  Turner, Edward Meeds, Javier Zazo, and Sushrut Karmalkar.
\newblock A fourier space perspective on diffusion models.
\newblock {\em arXiv preprint arXiv:2505.11278}, 2025.

\bibitem{doumeche2024physics}
Nathan Doumèche, Francis Bach, Gérard Biau, and Claire Boyer.
\newblock Physics-informed kernel learning.
\newblock {\em Journal of Machine Learning Research}, 26(1):1--39, 2025.
\newblock arXiv:2409.13786.

\bibitem{shannon1949communication}
Claude~E Shannon.
\newblock Communication in the presence of noise.
\newblock {\em Proceedings of the IRE}, 37(1):10--21, 1949.

\bibitem{larochelle2007empirical}
Hugo Larochelle, Dumitru Erhan, Aaron Courville, James Bergstra, and Yoshua
  Bengio.
\newblock An empirical evaluation of deep architectures on problems with many
  factors of variation.
\newblock In {\em Proceedings of the 24th international conference on Machine
  learning}, pages 473--480, 2007.

\bibitem{cmip6}
Brian~C O'Neill, Claudia Tebaldi, Detlef~P Van~Vuuren, Veronika Eyring, Pierre
  Friedlingstein, George Hurtt, Reto Knutti, Elmar Kriegler, Jean-Francois
  Lamarque, Jason Lowe, et~al.
\newblock The scenario model intercomparison project (scenariomip) for cmip6.
\newblock {\em Geoscientific Model Development}, 9(9):3461--3482, 2016.

\bibitem{hersbach2020era5}
Hans Hersbach, Bill Bell, Paul Berrisford, Shoji Hirahara, Andr{\'a}s
  Hor{\'a}nyi, Joaqu{\'\i}n Mu{\~n}oz-Sabater, Julien Nicolas, Carole Peubey,
  Raluca Radu, Dinand Schepers, et~al.
\newblock The era5 global reanalysis.
\newblock {\em Quarterly journal of the royal meteorological society},
  146(730):1999--2049, 2020.

\bibitem{dormand1980family}
John~R Dormand and Peter~J Prince.
\newblock A family of embedded runge-kutta formulae.
\newblock {\em Journal of computational and applied mathematics}, 6(1):19--26,
  1980.

\bibitem{hardStoInt1}
Kaiqi Chen, Eugene Lim, Kelvin Lin, Yiyang Chen, and Harold Soh.
\newblock Don't start from scratch: Behavioral refinement via interpolant-based
  policy diffusion.
\newblock {\em arXiv preprint arXiv:2402.16075}, 2024.

\bibitem{hardStoInt2}
Siyi Chen, Yixuan Jia, Qing Qu, He~Sun, and Jeffrey~A Fessler.
\newblock Flowdas: A stochastic interpolant-based framework for data
  assimilation.
\newblock {\em arXiv preprint arXiv:2501.16642}, 2025.

\bibitem{mnist}
Yann LeCun.
\newblock The mnist database of handwritten digits.
\newblock {\em http://yann. lecun. com/exdb/mnist/}, 1998.

\bibitem{resnet}
Kaiming He, Xiangyu Zhang, Shaoqing Ren, and Jian Sun.
\newblock Deep residual learning for image recognition.
\newblock In {\em Proceedings of the IEEE conference on computer vision and
  pattern recognition}, pages 770--778, 2016.

\bibitem{ogorman2006}
P.~A. O’Gorman and T.~Schneider.
\newblock Stochastic models for the kinematics of moisture transport and
  condensation in homogeneous turbulent flows.
\newblock {\em J. Atmos. Sci.}, 63:2992--3005, 2006.

\bibitem{access}
Tilo Ziehn, Matthew~A Chamberlain, Rachel~M Law, Andrew Lenton, Roger~W Bodman,
  Martin Dix, Lauren Stevens, Ying-Ping Wang, and Jhan Srbinovsky.
\newblock The australian earth system model: Access-esm1. 5.
\newblock {\em Journal of Southern Hemisphere Earth Systems Science},
  70(1):193--214, 2020.

\bibitem{nash1970river}
J~Eamonn Nash and Jonh~V Sutcliffe.
\newblock River flow forecasting through conceptual models part i—a
  discussion of principles.
\newblock {\em Journal of hydrology}, 10(3):282--290, 1970.

\bibitem{oppenheim1999signals}
Alan~V Oppenheim, Alan~S Willsky, and Hamid Nawab.
\newblock {\em Signals and Systems}.
\newblock Prentice Hall, 1999.

\end{thebibliography}

\appendix

\section{Extanded state-of-the-art}
The following section reviews prior work on domain alignment from unpaired distributions. 
Unless otherwise specified, we denote the two domains to be aligned as $\mathcal{D}_A$ and $\mathcal{D}_B$, with individual samples $x_A^i \in \mathcal{D}_A$ and $x_B^i \in \mathcal{D}_B$. 
The samples in $\mathcal{D}_A$ and $\mathcal{D}_B$ are assumed to follow the distributions $p_A$ and $p_B$, respectively. 
Latent representations are typically denoted by $\mathcal{Z}$, with $z_A^i \in \mathcal{Z}$ corresponding to $x_A^i$, and $z_B^i \in \mathcal{Z}$ corresponding to $x_B^i$. 
For a bijective function $f$, its inverse is written as $f^{-1}$, and a neural network with parameters $\theta$ is denoted $f_\theta$.

\paragraph{CycleGAN}

\begin{figure}[htbp]
    \centering

    % CycleGAN
    \begin{subfigure}[c]{0.45\linewidth}
        \centering
        \begin{tikzpicture}[node distance=1.2cm, every node/.style={align=center, font=\small}]
            \node (px) [draw, rectangle] {$p_A$};
            \node (py) [draw, rectangle, right=3cm of px] {$p_B$};
            \node (X) [draw, rectangle, below=of px] {$\Da$};
            \node (Y) [draw, rectangle, below=of py] {$\Db$};
            \draw[->, thick, bend left=30] ([yshift=1mm]px.east) to node[above] {$\Gab$} ([yshift=1mm]py.west);
            \draw[->, thick, bend left=30] ([yshift=-1mm]py.west) to node[below] {$\Gba$} ([yshift=-1mm]px.east);
            \draw[->, thick] (px.south) to node[left] {$C_{\Da}$} (X.north);
            \draw[->, thick] (py.south) to node[left] {$C_{\Db}$} (Y.north);
        \end{tikzpicture}
        \caption{CycleGAN \citep{cyclegan}.}
        \label{fig:cyclegan}
    \end{subfigure}
    \hfill
    % UNIT
    \begin{subfigure}[c]{0.45\linewidth}
        \centering
        \begin{tikzpicture}[node distance=1cm, every node/.style={align=center, font=\small}]
            \node (px) [draw, rectangle] {$p_A$};
            \node (py) [draw, rectangle, right=3cm of px] {$p_B$};
            \node (z) [draw, rectangle, above=of $(px)!0.5!(py)$] {$Z \sim \p2$};
            \draw[<-, thick, bend left=30] (px.north) to node[left, xshift=-1mm, yshift=3mm] {$G_{\mathcal{Z}\to \Da}$} (z.west);
            \draw[<-, thick, bend right=30] (py.north) to node[right, xshift=1mm, yshift=3mm] {$G_{Z\to \Db}$} (z.east);
            \draw[->, thick, bend right=30] (px.east) node[right, yshift=-2mm] {$E_{\Da \to \mathcal{Z}}$} to ([xshift=-2mm]z.south);
            \draw[->, thick, bend left=30] (py.west) node[left, yshift=-2mm] {$E_{\Db \to \mathcal{Z}}$}  to  ([xshift=2mm]z.south);
            \node (X) [draw, rectangle, below=1cm of px] {$\Da$};
            \node (Y) [draw, rectangle, below=1cm of py] {$\Db$};
            \draw[->, thick] (px.south) to node[left] {$C_{\Da}$} (X.north);
            \draw[->, thick] (py.south) to node[right] {$C_{\Db}$} (Y.north);
        \end{tikzpicture}
        \caption{UNIT \citep{munit}.}
        \label{fig:unit}
    \end{subfigure}

    \vspace{0.5cm}

    % AlignFlow
    \begin{subfigure}[c]{0.4\linewidth}
        \centering
        \begin{tikzpicture}[node distance=2.5cm, every node/.style={align=center, font=\small}]
            \node (px) [draw, rectangle] {$p_A$};
            \node (py) [draw, rectangle, right=3.8cm of px] {$p_B$};
            \node (z) [draw, rectangle, above=of $(px)!0.5!(py)$] {$Z \sim \p2$};
            \draw[<->, thick, align=left] (px.north) -- node[left, xshift=-1mm, yshift=3mm] {$\Gaz=$\\$G^{-1}_{\mathcal{Z}\to \Da}$} ([xshift=-1mm]z.south);
            \draw[<->, thick, align=right] (py.north) -- node[right, xshift=1mm, yshift=3mm] {$\Gbz=$\\$G^{-1}_{Z\to \Db}$} ([xshift=2mm]z.south);
            \draw[->, thick, dashed, bend left=30] ([yshift=1mm]px.east) to node[above] {$\Gbz^{-1} \circ \Gaz$} ([yshift=1mm]py.west);
            \draw[->, thick, dashed, bend left=30] ([yshift=-1mm]py.west) to node[below] {$\Gaz^{-1} \circ \Gbz$} ([yshift=-1mm]px.east);
            \node (X) [draw, rectangle, below=1.2cm of px] {$\Da$};
            \node (Y) [draw, rectangle, below=1.2cm of py] {$\Db$};
            \draw[->, thick] (px.south) to node[left] {$C_{\Da}$} (X.north);
            \draw[->, thick] (py.south) to node[right] {$C_{\Db}$} (Y.north);
        \end{tikzpicture}
        \caption{AlignFlow \citep{alignflow}.}
        \label{alignflow}
    \end{subfigure}
    \hfill
    % Dual Diffusion
    \begin{subfigure}[c]{0.55\linewidth}
        \centering
        \begin{tikzpicture}[node distance=1.5cm, every node/.style={align=center, font=\small}]
            \node (px) [draw, rectangle] {$x_A \sim p_A$};
            \node (pz) [draw, rectangle, right=2cm of px] {$z \sim \p2$};
            \node (py) [draw, rectangle, right=2cm of pz] {$x_B \sim p_B$};
            \draw[->, thick, bend left = 30] ([yshift=1mm]px.east) to node[above] {$\mathrm{ODS}(x_A,f_\theta^A, 0, 1)$} ([yshift=1mm]pz.west);
            \draw[->, thick, bend left = 30] ([yshift=1mm]pz.east) to node[above] {$\mathrm{ODS}(z,f_\theta^B, 1, 0)$} ([yshift=1mm]py.west);
            \draw[->, thick, bend left = 30] ([yshift=-1mm]pz.west) to node[below] {$\mathrm{ODS}(z,f_\theta^A, 1, 0)$} ([yshift=-1mm]px.east);
            \draw[->, thick, bend left = 30] ([yshift=-1mm]py.west) to node[below] {$\mathrm{ODS}(x_B,f_\theta^B, 0, 1)$} ([yshift=-1mm]pz.east);
        \end{tikzpicture}
        \caption{Dual Diffusion Implicit Bridge \citep{dualimplicitbridge}, $\mathrm{ODS}$ explicit formula is given Equation~\ref{eq:ods}.}
        \label{fig:ddib}
    \end{subfigure}

    \vspace{0.5cm}

    % Stochastic Interpolants
    \begin{subfigure}[c]{0.45\linewidth}
        \centering
        \begin{tikzpicture}[node distance=2cm, every node/.style={align=center, font=\small}]
            \node (px) [draw, rectangle] {$x_A = X_0 \sim p_A$};
            \node (py) [draw, rectangle, right=2.5cm of px] {$x_B = X_1 \sim p_B$};
            \draw[->, thick, bend left=30] ([yshift=1mm]px.east) to node[above] {Forward Flow $x_A \mapsto x_B$} ([yshift=1mm]py.west);
            \draw[->, thick, bend left=30] ([yshift=-1mm]py.west) to node[below] {Reverse Flow $x_B \mapsto x_A$} ([yshift=-1mm]px.east);
        \end{tikzpicture}
        \caption{Stochastic Interpolants \citep{stochasticinterpolants, schrodingerflow, bridgematching}.}
        \label{fig:stoint}
    \end{subfigure}
    \hfill
    % Diffusion bridge for unpaired downscaling
    \begin{subfigure}[c]{0.45\linewidth}
        \centering
        \begin{tikzpicture}[node distance=4cm, every node/.style={align=center, font=\small}]
            \node (px) [draw, rectangle] {$z = X_0 \sim \p2$};
            \node (py) [draw, rectangle, right=2cm of px] {$x_B = X_1 \sim p_B$};
            \node (pA) [draw, rectangle, above=1cm of py] {$x_A \sim p_A$};
            \draw[->, thick, dashed, bend right=30] (pA.west) to node[left, yshift=1mm] {Noising $x_A \mapsto x_{t^\star}$} ([xshift=0.7cm]px.east);
            \draw[->, thick, dashed] ([xshift=0.7cm]px.east) to node[above, yshift=2mm, xshift=15mm] {Reverse Flow $x_{t^\star} \mapsto x_B$} (py.west);
            \draw[<-, thick, bend left=30] ([yshift=-1mm]py.west) to node[below] {Reverse Flow $z \mapsto x_B$} ([yshift=-1mm]px.east);
        \end{tikzpicture}
        \caption{Diffusion bridge for unpaired downscaling \citep{bischoff2024unpaired}.}
        \label{fig:diffbridge}
    \end{subfigure}

    \caption{Overview of different domain alignment frameworks for unpaired distributions.}
    \label{fig:domain_alignment}
\end{figure}
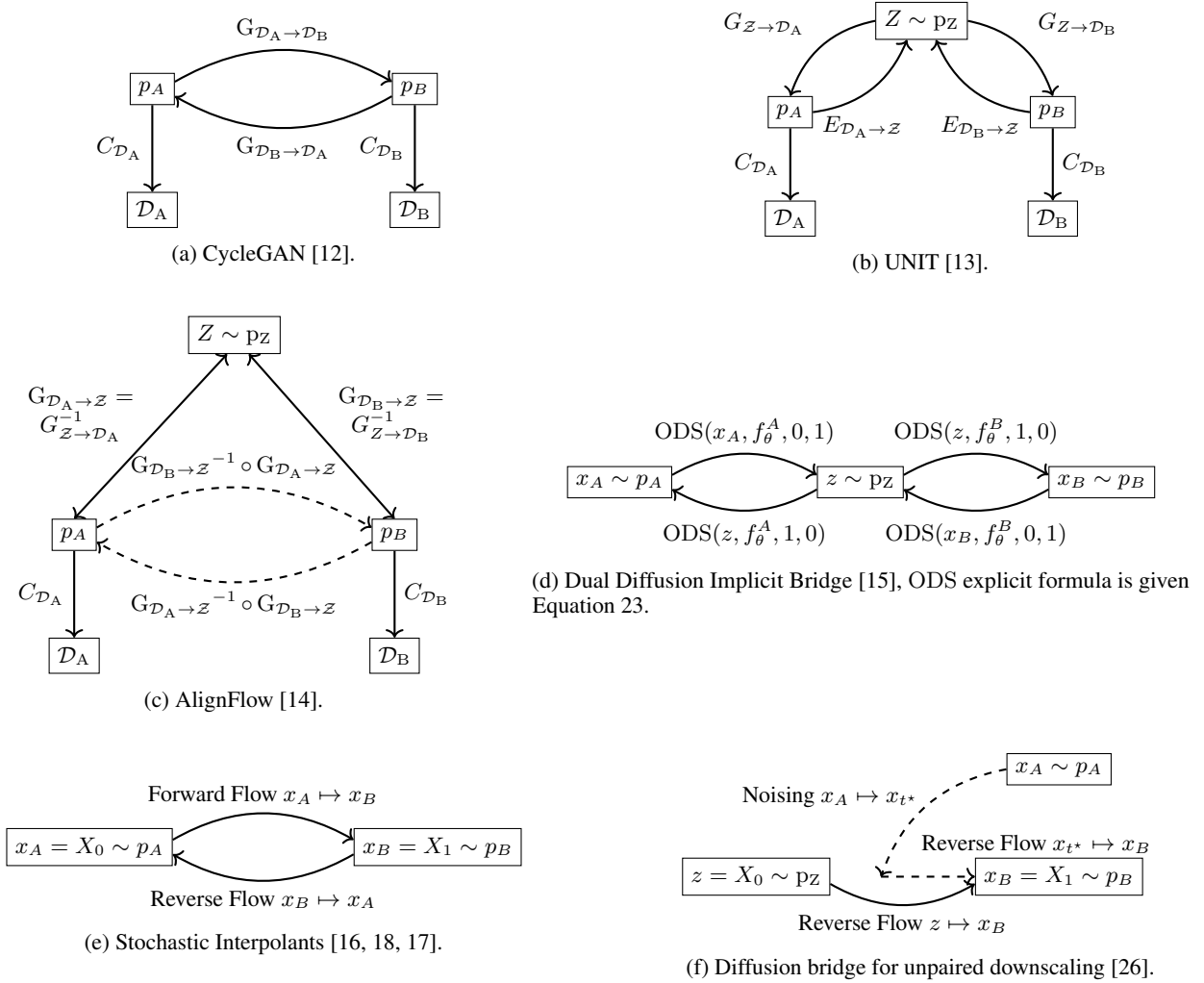

CycleGAN~\citep{cyclegan}, see Figure~\ref{fig:cyclegan}, is an image-to-image translation framework designed to learn mappings between two unpaired domains through adversarial and cycle-consistency objectives. It trains two bidirectional mappings, $\Gab$ and $\Gba$, along with their corresponding discriminators, $C_{\Db}$ and $C_{\Da}$. The pair $(\Gab, C_{\Db})$ learns to generate samples in $\Db$ from $\Da$ that are indistinguishable from real data, while $(\Gba, C_{\Da})$ performs the inverse translation.  
Each generator receives an input image from the source domain and aims to fool the discriminator of the target domain. To enforce consistency between the two mappings, a cycle-consistency loss ensures that translating an image to the other domain and back recovers the original input, i.e.,
\[
\tilde{x}_A = \Gba\big(\Gab(x_A)\big) \approx x_A, 
\quad 
\tilde{x}_B = \Gab\big(\Gba(x_B)\big) \approx x_B.
\]
This constraint regularizes the adversarial training and prevents the generators from producing arbitrary mappings that do not preserve the underlying content.

\paragraph{UNIT}

UNIT (Unsupervised Image-to-Image Translation Networks)~\citep{munit}, see Figure~\ref{fig:unit}, extends unpaired domain translation by introducing a shared latent space assumption. It postulates that corresponding images in two domains $\Da$ and $\Db$ can be mapped to a common latent representation $z \in \mathcal{Z}$ via domain-specific encoders $E_{\Da \to \mathcal{Z}}$ and $E_{\Db \to \mathcal{Z}}$, such that $E_{\Da \to \mathcal{Z}} \# p_A \approx E_{\Db \to \mathcal{Z}}\#p_B \approx \p2$, where $\#$ denotes the push-forward measure. The model employs two variational autoencoders (VAEs) coupled with GAN losses to ensure both reconstruction quality and domain realism.  
Formally, each domain has an encoder–decoder pair $(E_{\Da \to \mathcal{Z}}, G_{\mathcal{Z}\to \Da})$ and $(E_{\Db \to \mathcal{Z}}, G_{\mathcal{Z}\to \Db})$. Translation from $\Da$ to $\Db$ is achieved by encoding and decoding through the opposite generator: $\tilde{x}_B = G_{\mathcal{Z}\to \Db}\big(E_{\Da \to \mathcal{Z}}(x_A)\big)$.

\paragraph{AlignFlow}
AlignFlow \citep{alignflow}, see Figure~\ref{alignflow}, shares the idea of latent space from UNIT through modernizing CycleGAN by replacing the GANs generators with (invertible) normalizing flows mapping to a shared latent distribution $\p2$ (typically a Gaussian, although normalizing flows do not require a Gaussian prior). Passing from $\Da$ to $\Db$ is done using the forward generator from $\Da$ to $\mathcal{Z}$, $\Gaz$, and then the inverse of the generator from $\Db$ to $\mathcal{Z}$, $\Gbz^{-1}=G_{\mathcal{Z}\to \Db}$. Using the inverse passes of the generators leads to transferring an element from $\Db$ to $\Da$. The invertible properties of the normalizing flows ensure the cycle-consistency.
\paragraph{Dual Diffusion Implicit Bridge}
Dual Diffusion Implicit Bridge \citep{dualimplicitbridge}, see Figure~\ref{fig:ddib}, increments AlignFlow by removing the need for adversarial training. To generate data from $p_A$ to $p_B$, one diffusion process is trained per domain. Then, transferring a sample $x_A \sim p_A$ to $p_B$ is done using the forward diffusion from $\Da$ pushing $x_A$ to an intermediate latent space $\p2$, and then applying a reverse diffusion from $\p2$ to $p_B$. The inverse transformation can be done using the forward from $p_B$ to $\p2$ and the reverse from $\p2$ to $p_A$.
\paragraph{Bridge Matching, Stochastic Interpolants, Schrödinger Bridge Flows} Bridge Matching \citep{bridgematching}, Stochastic Interpolants \citep{stochasticinterpolants} and Schrödinger Bridge Flows \citep{schrodingerflow}, see Figure~\ref{fig:stoint}, refer to a similar idea: learning a transport map from an unknown distribution $p_A$ to another unknown distribution $p_B$, which corresponding domains may or may not be unaligned, using score-based generative models. They may differ in how the path between a point from $\Da$ and a point from $\Db$ is defined or what learning objective the model actually learns.
Unlike AlignFlow or Dual Diffusion, those frameworks do not require an explicit latent variable. The entire transformation is learned as a dynamic flow between random pairings, and the model naturally converges towards a bidirectional mapping between $p_A$ and $p_B$.

\paragraph{Rectified Flows} Building upon this family of stochastic bridges, Rectified Flows~\citep{rectifiedflow} represents the transport plan between $p_A$ and $p_B$ as an ordinary differential equation, whose flows have a nice non-crossing property. By recursively applying their flow framework, they obtain straight paths between data from $\Da$ and $\Db$. Those straight paths allow them to generate the closest point from $x_A$ within $\Db$.

\paragraph{Diffusion bridge for unpaired downscaling} Specifically designed for the matching between unpaired low-resolution ($\Da$) and high-resolution ($\Db$) datasets of physical processes, \citep{bischoff2024unpaired}, see Figure~\ref{fig:diffbridge}, this approach first trains a diffusion model to map a Gaussian distribution to $p_B$, and then noise until an optimized time step $t^\star$ the data from $p_B$ before denoising with the diffusion model. The idea is that noising erases first the high frequencies, making the intermediate space at $t^\star$ a domain where the data from $\Da$ and $\Db$ are indistinguishable.
\begin{remark}
SerpentFlow is similar to diffusion bridges, but differs from them in two main ways, even when using noise for the $\mu$ function. (i) The noise used in the diffusion bridge affects all frequencies. Therefore, there is no guarantee that low frequencies will be preserved. (ii) The SerpentFlow approach is more direct because the generative pipeline is trained directly from the common state where the data becomes indistinguishable. In the case of diffusion bridges, a pipeline pre-trained from pure noise is used. By adding noise to low-resolution data, we hope to be on the diffusion path, but this is not guaranteed.
\end{remark}

\newpage
\section{On the Validity of Classifier-Based Cutoff Selection}
\label{app:cutoff_classifier}

This appendix provides an empirical justification for the discriminator-based cutoff selection procedure described in Section~\ref{method_cutoff}.  
We illustrate the method on the GCM\,$\to$\,ERA5 downscaling experiment and show that the classifier accuracy provides a reliable indicator for identifying the frequency threshold $\omega_c$ that separates shared large-scale structures from domain-specific high-frequency content.

\begin{figure}[htbp]
    \begin{subfigure}[c]{0.49\linewidth}
        \centering
        \includegraphics[width=\linewidth]{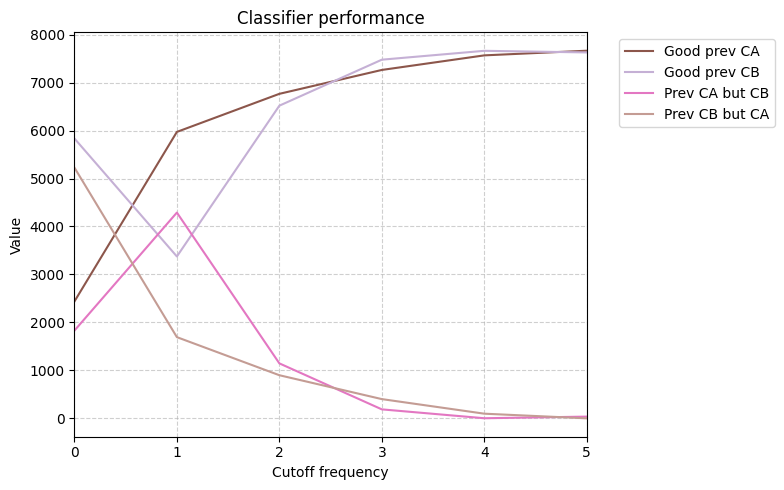}
        \caption{Classifier bad and good prediction rate}
    \end{subfigure}
    \hfill
    \begin{subfigure}[c]{0.49\linewidth}
        \centering
        \includegraphics[width=\linewidth]{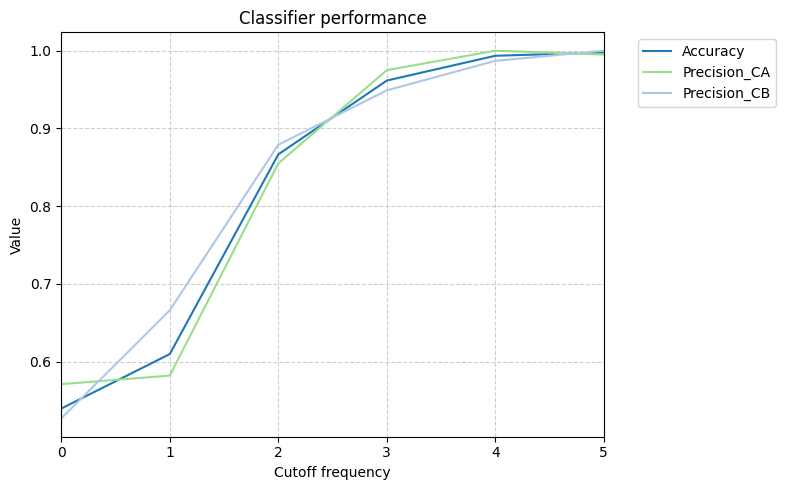}
        \caption{Classifier metrics}
    \end{subfigure}
    \caption{Classifier between the GGCM (dataset A, CA) and ERA5 (dataset B, CB) performance for different values of cutoff frequency.}
    \label{fig:superes_classifier}
\end{figure}

For each candidate cutoff $\omega_c$, we low-pass filter all samples following Eq.~\eqref{eq:lowpass} and train the domain discriminator $D_\psi$ using the loss of Eq.~\eqref{eq:discriminator_loss}.  
Figure~\ref{fig:superes_classifier} reports the resulting classification accuracy as a function of~$\omega_c$.

We observe a clear trend:
\begin{itemize}
    \item for high cutoffs ($\omega_c \ge 2$), the classifier achieves high accuracy, indicating that domain-specific information is still present in the filtered samples;
    \item accuracy progressively decreases as $\omega_c$ is reduced;
    \item once $\omega_c \le 1$, the classifier accuracy approaches $60\%$ and stabilises near chance level.
\end{itemize}

Following the criterion of Eq.~\eqref{eq:omega_opt}, this behaviour identifies $\omega_c = 1$ as the smallest cutoff for which the low-frequency representations of GCM and ERA5 become nearly indistinguishable.  
This value is therefore used in the main experiments.

To validate that this choice is meaningful, we repeat the full downscaling pipeline for cutoffs
\[
\omega_c \in \{0, 1, 2, 3\}
\]
and analyse the resulting reconstructions.

\begin{figure}[htbp]
    \centering
    \begin{subfigure}[c]{0.98\linewidth}
        \centering
        \includegraphics[width=\linewidth]{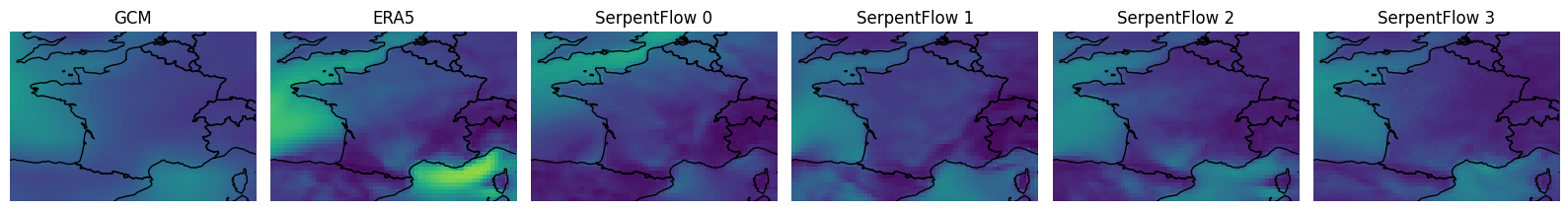}
        \caption{Date: 2004-04-11.}
        \label{fig:comp_i_100}
    \end{subfigure}
    \begin{subfigure}[c]{0.98\linewidth}
        \centering
        \includegraphics[width=\linewidth]{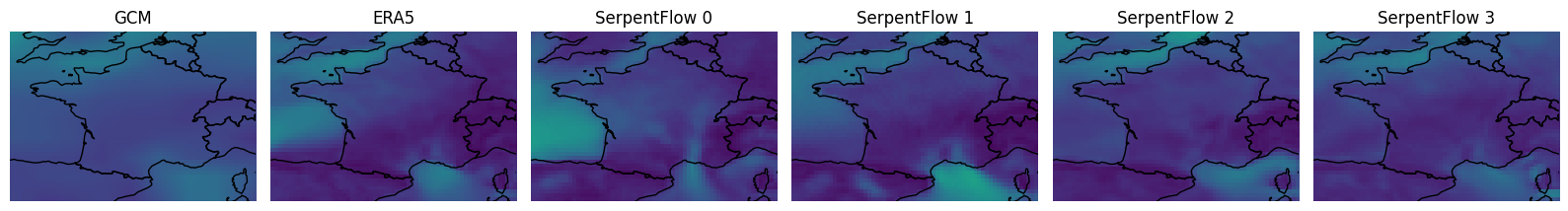}
        \caption{Date: 2011-08-04.}
        \label{fig:comp_i_3500}
    \end{subfigure}
    \begin{subfigure}[c]{0.98\linewidth}
        \centering
        \includegraphics[width=\linewidth]{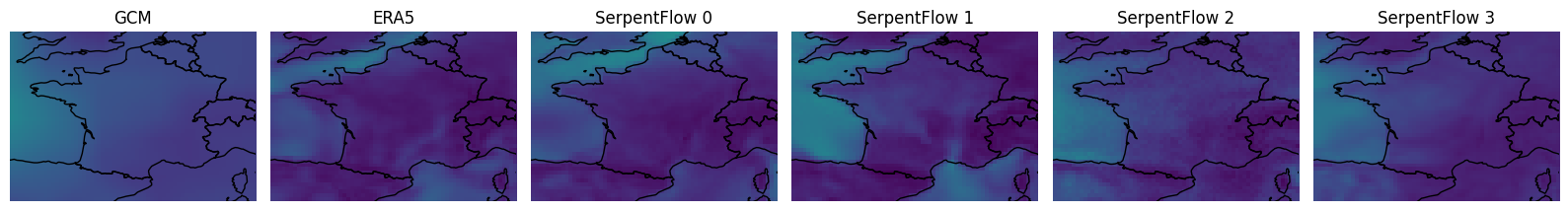}
        \caption{Date: 2018-06-10.}
        \label{fig:comp_i_6000}
    \end{subfigure}
    \caption{Downscaling outputs for various days within the test set, for SerpentFlow pipeline with cutoffs $\in \{0, 1, 2, 3\}$.}
    \label{fig:comp_i}
\end{figure}

\begin{figure}[htbp]
    \centering
    \begin{subfigure}[c]{0.98\linewidth}
        \centering
        \includegraphics[width=\linewidth]{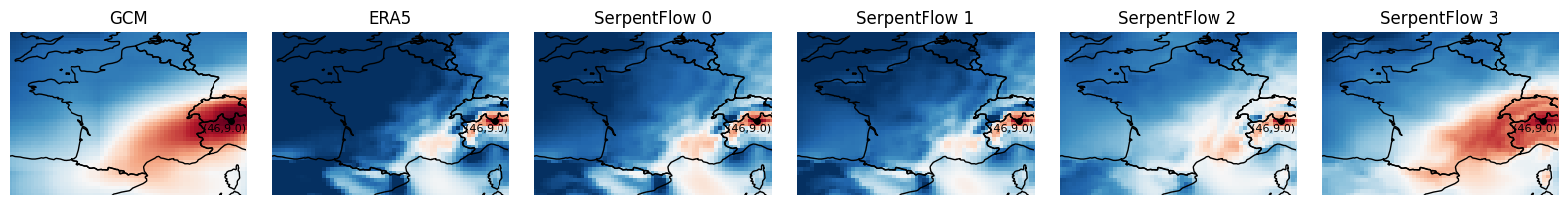}
        \caption{Correlation map, coordinate (46,9).}
        \label{fig:comp_corr_alps}
    \end{subfigure}
    \begin{subfigure}[c]{0.98\linewidth}
        \centering
        \includegraphics[width=\linewidth]{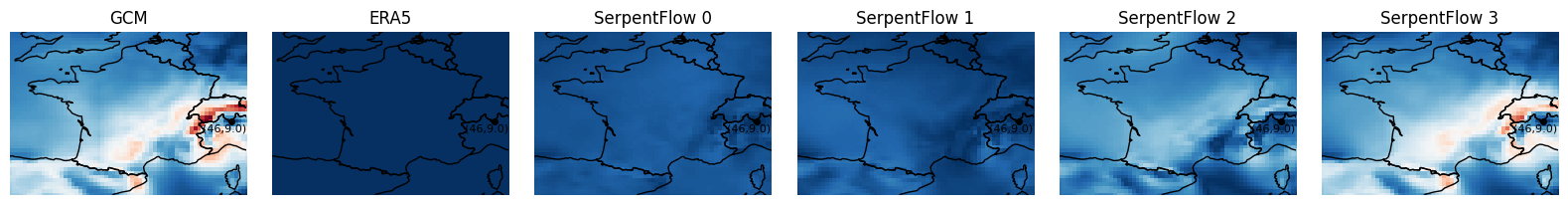}
        \caption{Bias correlation map w.r.t. ERA5, coordinate (46,9).}
        \label{fig:comp_corr_bias_alps}
    \end{subfigure}
    \begin{subfigure}[c]{0.98\linewidth}
        \centering
        \includegraphics[width=\linewidth]{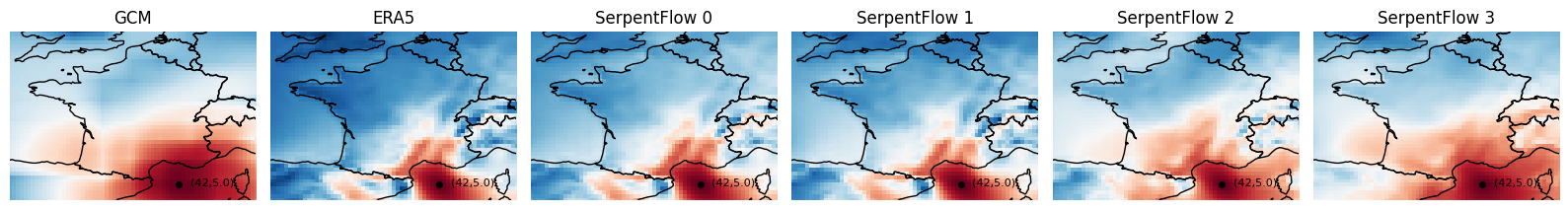}
        \caption{Correlation map, coordinate (42,5).}
        \label{fig:comp_corr_med}
    \end{subfigure}
    \begin{subfigure}[c]{0.98\linewidth}
        \centering
        \includegraphics[width=\linewidth]{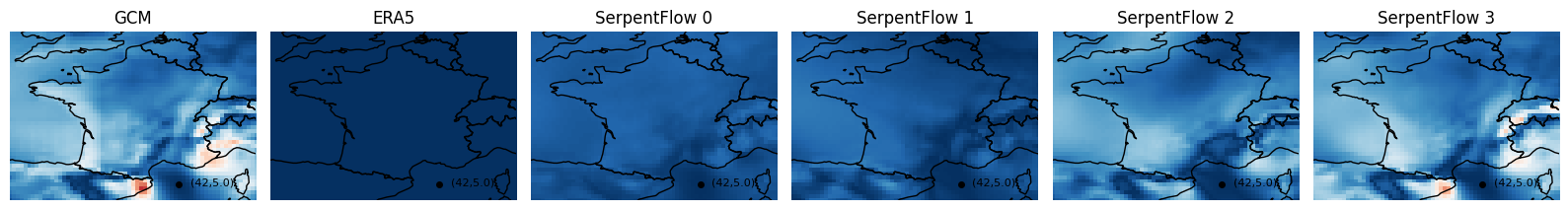}
        \caption{Bias correlation map w.r.t. ERA5, coordinate (42,5).}
        \label{fig:comp_corr_bias_med}
    \end{subfigure}
    \caption{Correlation maps of wind intensity fields. We select one point of interest, and compute correlations between the corresponding time series over the period compared to the time series of the other grid points. For each method, we compare the correlation map with that from ERA5. In Figures~\ref{fig:comp_corr_alps} and~\ref{fig:comp_corr_bias_alps}, the reference grid point is in the Alps. For Figures~\ref{fig:comp_corr_med} and~\ref{fig:comp_corr_bias_med}, the reference grid point is in the Mediterannean sea.}
    \label{fig:comp_climate_corr}
\end{figure}

\paragraph{Cutoff too high ($\omega_c = 2$ or $3$).}
In this regime, the presumed shared component still contains domain-specific high-frequency structure from ERA5.  
Consequently, the pseudo-inputs fed to the flow model retain mismatched fine-scale patterns.  
As shown in Figure~\ref{fig:comp_climate_corr}, generated samples fail to reproduce key ERA5 spatial features, the correlation maps mismatch from the one of ERA5.  
This confirms that excessive cutoffs let through high frequencies that should instead be generated by the model.

\begin{figure}[htbp]
    \centering
    \begin{subfigure}[c]{0.7\linewidth}
        \centering
        \includegraphics[width=\linewidth]{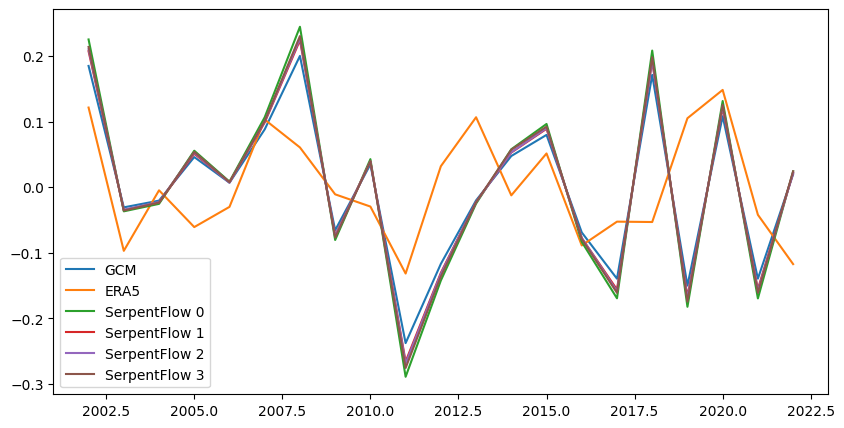}
        \caption{Normalized Yearly mean.}
        \label{fig:comp_climate_yearly_mean}
    \end{subfigure}
    \begin{subfigure}[c]{0.98\linewidth}
        \centering
        \includegraphics[width=\linewidth]{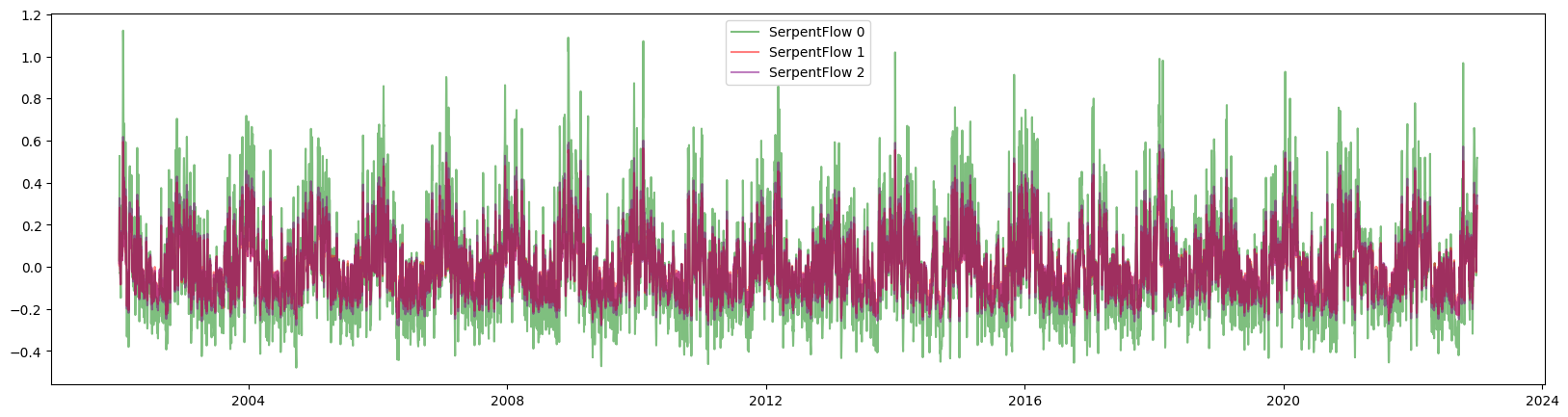}
        \caption{Temporal bias w.r.t. normalized spatially averaged GCM}
        \label{fig:comp_climate_ts_bias}
    \end{subfigure}
    \caption{Illustration of the temporal consistency of SerpentFlow with $\omega_c \in \{0,1,2\}$ compared to climate model dynamics.}
    \label{fig:comp_climate_temporal}
\end{figure}

\paragraph{Cutoff too low ($\omega_c = 0$).}
Here, almost all frequencies are treated as domain-specific noise. This leads to a reduce ability of the model to keep the global temporal dynamic of the GCM as shown in Figure~\ref{fig:comp_climate_temporal} while not improving the reconstruction of ERA5 fine-scale detail (see Figure~\ref{fig:comp_climate_corr}).
This illustrates that overly aggressive filtering removes signal rather than separating invariant structure from variability.

\paragraph{Optimal cutoff ($\omega_c = 1$).}
At this value, the pseudo-inputs preserve the correct low-frequency backbone of GCM while delegating high-frequency variability to the generative model.  
Reconstructed fields exhibit sharp contours, realistic ERA5-like small-scale patterns, and consistent large-scale structures.  
These results are fully aligned with the classifier-based criterion, demonstrating the usefulness of the approach. This qualitative analysis is supported by the metrics reporte Table~\ref{tab:climate_res_comp}.
\begin{table}[htbp]
\centering
\caption{
Evaluation of SerpentFlow downscaling performance on the climate dataset for various cutoff frequencies ($\omega_c$).
KS: Kolmogorov–Smirnov statistic between generated and target distributions;
Correlation Score: mean correlation bias relative to ERA5;
Temporal RMSE: root mean square error of the spatially-averaged daily wind intensity over the validation period;
NSE: Nash–Sutcliffe efficiency index computed on the spatially-averaged daily series (1 = perfect, 0 = mean predictor).
}
\label{tab:climate_res_comp}
\begin{tabular}{lcccc}
\toprule
\textbf{Method} & \textbf{KS} ($\downarrow$) & \textbf{Correlation Score} ($\downarrow$) & \textbf{Temporal RMSE} ($\downarrow$) & \textbf{NSE} ($\uparrow$)\\
\midrule
$\omega_c = 0$ & \textbf{0.017}    & \underline{0.084} & 0.061 & 0.953 \\
$\omega_c = 1$ & 0.025             & \textbf{0.045}    & \textbf{0.018} & \textbf{0.986} \\
$\omega_c = 2$ & \underline{0.020} & 0.088             & \underline{0.023} & \underline{0.982} \\
$\omega_c = 3$ & 0.045             & 0.119             & 0.033 & 0.975 \\
\bottomrule
\end{tabular}
\end{table}

The classifier accuracy serves as a reliable and interpretable proxy for identifying the frequency band where the two domains become statistically indistinguishable.  
This provides a principled way to determine the cutoff $\omega_c$ without manual tuning.  
Empirical results on the GCM\,$\to$\,ERA5 task confirm the method's validity:  
cutoffs chosen via the discriminator lead to substantially better reconstructions than ad hoc alternatives, while mis-specified cutoffs produce systematic and interpretable failure modes.

\newpage
\section{Limitation: hyperresolution with incompatible low-frequencies}

In this section, we show the limitations of SerpentFlow in the case of hyperresolution when low frequencies do not work. The first example is on the MNIST dataset, while the second is on our physical simulation dataset for a wavenumber value that is too low.

\subsection{MNIST dataset}
\label{app:mnist_failure}

To illustrate the importance of the spectral compatibility assumption discussed in Section~\ref{method}, we present a controlled failure case on MNIST. 
We consider the standard MNIST dataset (28$\times$28) as the high-resolution domain $\mathcal{D}_B$, and use the 8$\times$8 grayscale MNIST variant from \texttt{sklearn} as the low-resolution domain $\Da$. 
Crucially, the 8$\times$8 images are not obtained by low-pass filtering the 28$\times$28 digits; they originate from a distinct acquisition and preprocessing pipeline. 
Therefore, their frequency spectra are fundamentally mismatched.

\paragraph{Interpolation baselines.}
We upsample the 8$\times$8 images to 28$\times$28 resolution using three standard interpolation procedures:
\begin{itemize}
    \item nearest-neighbor interpolation,
    \item bilinear interpolation,
    \item spectral interpolation (zero-padding in the Fourier domain).
\end{itemize}

\paragraph{Classifier-based assessment.}

\begin{figure}[htbp]
    \begin{subfigure}[c]{0.49\linewidth}
        \centering
        \includegraphics[width=\linewidth]{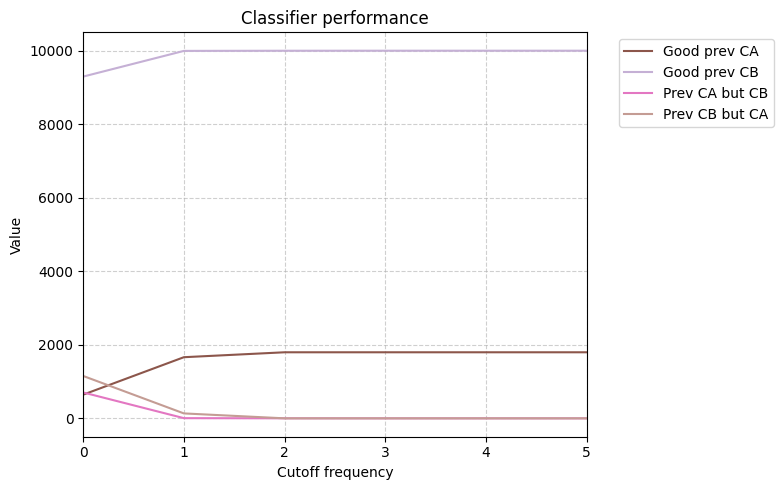}
        \caption{Classifier bad and good prediction rate}
    \end{subfigure}
    \hfill
    \begin{subfigure}[c]{0.49\linewidth}
        \centering
        \includegraphics[width=\linewidth]{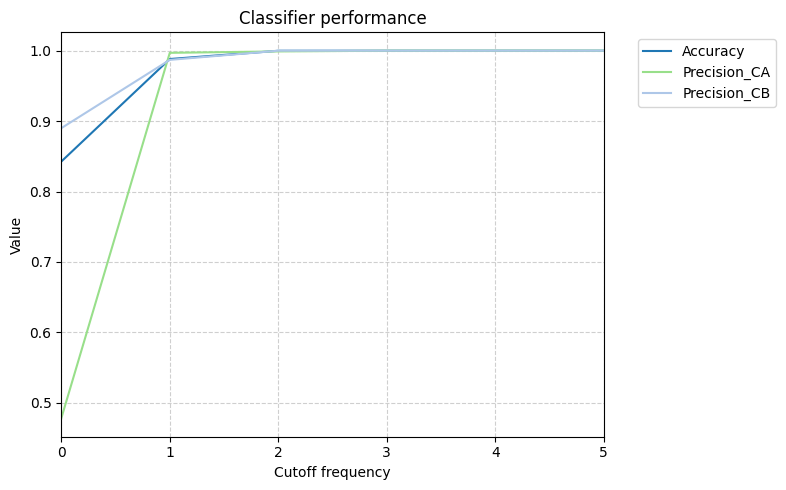}
        \caption{Classifier metrics}
    \end{subfigure}
    \caption{Classifier between linearly interpolated MNIST (dataset A, CA) and MNIST (dataset B, CB) performance for different values of cutoff frequency.}
    \label{fig:mnist_classifier}
\end{figure}

We train the convolution-based classifier discussed in the paper to distinguish between MNIST (28$\times$28) digits and their interpolated counterparts to find the optimal cutoff frequency.  
Even under aggressive low-pass filtering of both domains before classification, the discriminator achieves an accuracy never below 80\% across all interpolation methods. We show the performance for various cutoff frequencies for the linear interpolation in Figure~\ref{fig:mnist_classifier}.
This poor performance indicates that the interpolated images remain statistically very different from true MNIST samples, even when high-frequencies are removed.
\paragraph{Visual analysis.}
\begin{figure}[htbp]
    \centering
    \begin{subfigure}[c]{0.49\linewidth}
        \centering
        \includegraphics[width=\linewidth]{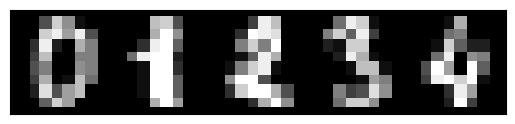}
        \caption{Nearest-neighbor interpolation}
    \end{subfigure}
    \hfill
    \begin{subfigure}[c]{0.49\linewidth}
        \centering
        \includegraphics[width=\linewidth]{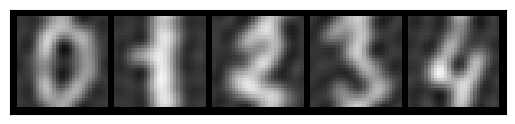}
        \caption{Low-pass filter applied on nearest-neighbor interpolation data}
    \end{subfigure}
    \hfill
    \begin{subfigure}[c]{0.55\linewidth}
        \centering
        \includegraphics[width=\linewidth]{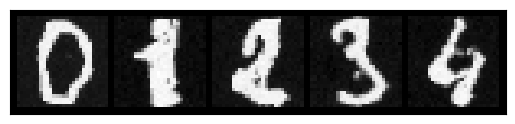}
        \caption{Prediction for nearest-neighbor interpolation}
    \end{subfigure}
    \begin{subfigure}[c]{0.49\linewidth}
        \centering
        \includegraphics[width=\linewidth]{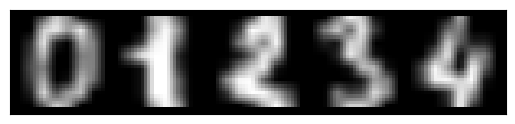}
        \caption{Linear interpolation}
    \end{subfigure}
    \hfill
    \begin{subfigure}[c]{0.49\linewidth}
        \centering
        \includegraphics[width=\linewidth]{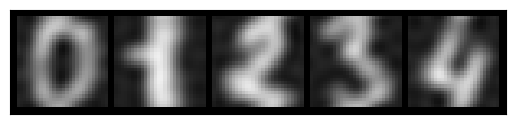}
        \caption{Low-pass filter applied on linear interpolation data}
    \end{subfigure}
    \hfill
    \begin{subfigure}[c]{0.55\linewidth}
        \centering
        \includegraphics[width=\linewidth]{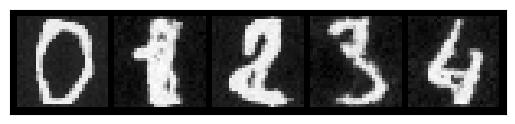}
        \caption{Prediction for linear interpolation}
    \end{subfigure}
    \begin{subfigure}[c]{0.49\linewidth}
        \centering
        \includegraphics[width=\linewidth]{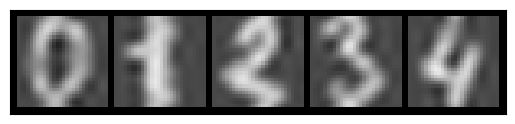}
        \caption{Spectral interpolation}
    \end{subfigure}
    \hfill
    \begin{subfigure}[c]{0.49\linewidth}
        \centering
        \includegraphics[width=\linewidth]{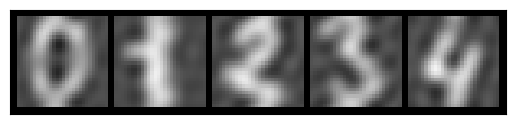}
        \caption{Low-pass filter applied on spectral interpolation data}
    \end{subfigure}
    \hfill
    \begin{subfigure}[c]{0.55\linewidth}
        \centering
        \includegraphics[width=\linewidth]{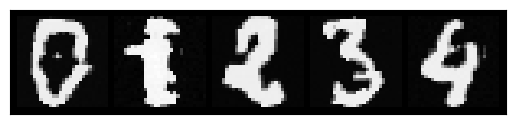}
        \caption{Prediction for spectral interpolation}
    \end{subfigure}
    \caption{Visualization of the different interpolations of the \texttt{sklearn}-MNIST dataset to go from resolution (9$times$9) to (28$\times$28), the result after applying the low pass filter, and the prediction with SerpentFlow to move into the MNIST domain.}
    \label{fig:mnist}
\end{figure}
Interpolated samples exhibit several artifacts:
\begin{itemize}
    \item blurred and inconsistent digit contours,
    \item spurious background noise,
    \item coarse textures that markedly differ from MNIST’s characteristic stroke geometry.
\end{itemize}
Representative reconstructions do not resemble the MNIST distribution, confirming that spatial downsampling in \texttt{sklearn}-MNIST does not correspond to a frequency truncation of the original digits.

This experiment illustrates that our frequency-based decomposition has its limit when the low-resolution domain does not follow the low-frequency structure of the high-resolution domain. 
When this compatibility fails---as in MNIST versus \texttt{sklearn}-MNIST---the problem becomes ill-posed: no generative model can infer high-resolution digits that meaningfully resemble the original dataset, since the coarse observations do not correspond to low-frequency measurements of the target distribution.

\subsection{Fluid simulation dataset}\label{app:fluids}

\paragraph{Limitations when $k_x=k_y$ are too low}

\begin{figure}[htbp]
    \centering
    \begin{subfigure}[c]{0.99\linewidth}
        \centering
        \includegraphics[width=\linewidth]{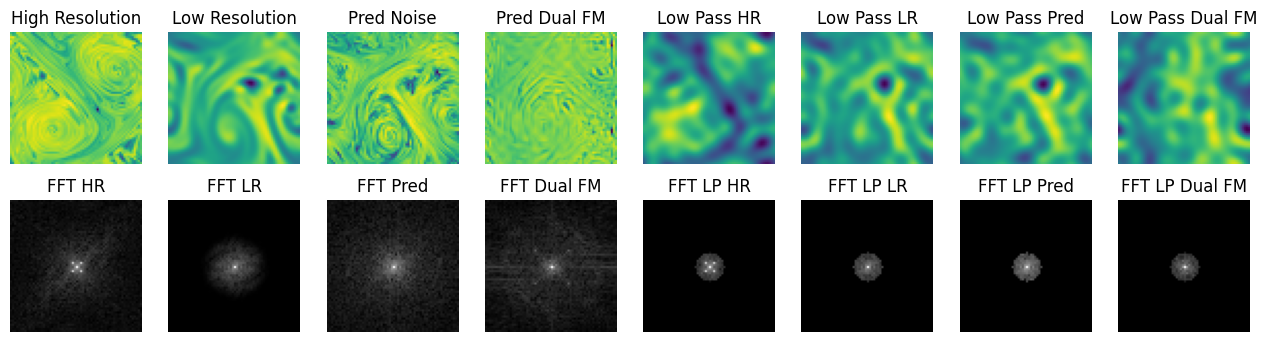}
        \caption{Vorticity field for $k_x=k_y=0$, when $w_c=7$.}
        \label{fig:k0_wc_7}
    \end{subfigure}
    \begin{subfigure}[c]{0.99\linewidth}
        \centering
        \includegraphics[width=\linewidth]{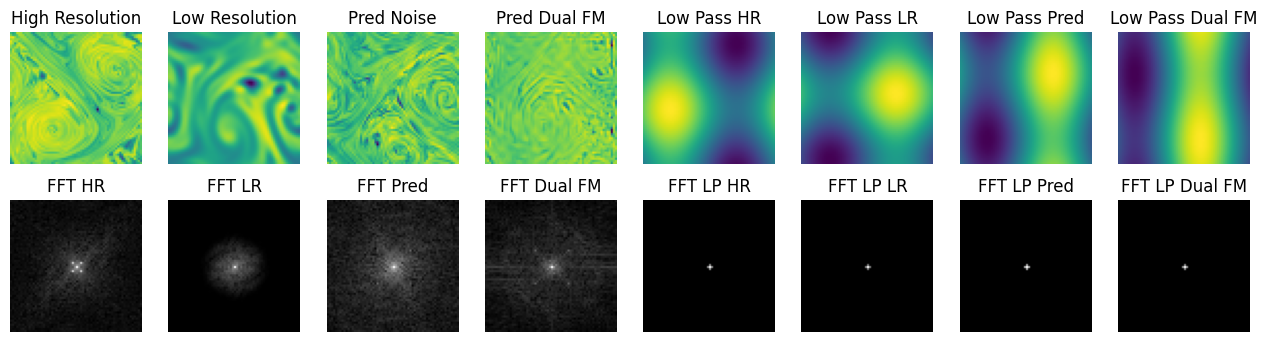}   
        \caption{Vorticity field for $k_x=k_y=0$, when $w_c=1$.}
        \label{fig:k0_w_1}
    \end{subfigure}
    \begin{subfigure}[c]{0.99\linewidth}
        \centering
        \includegraphics[width=\linewidth]{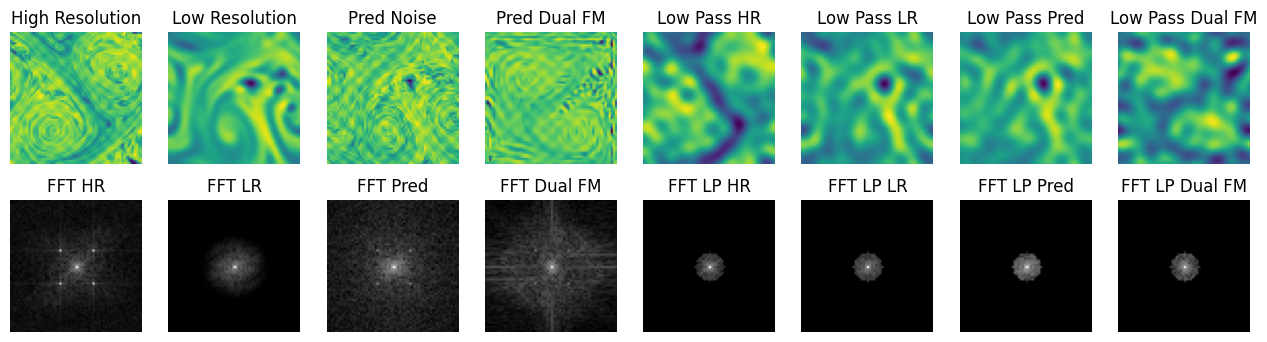}
        \caption{Vorticity field for $k_x=k_y=8$, when $w_c=7$.}
        \label{fig:k8_w_7}
    \end{subfigure}
    \begin{subfigure}[c]{0.99\linewidth}
        \centering
        \includegraphics[width=\linewidth]{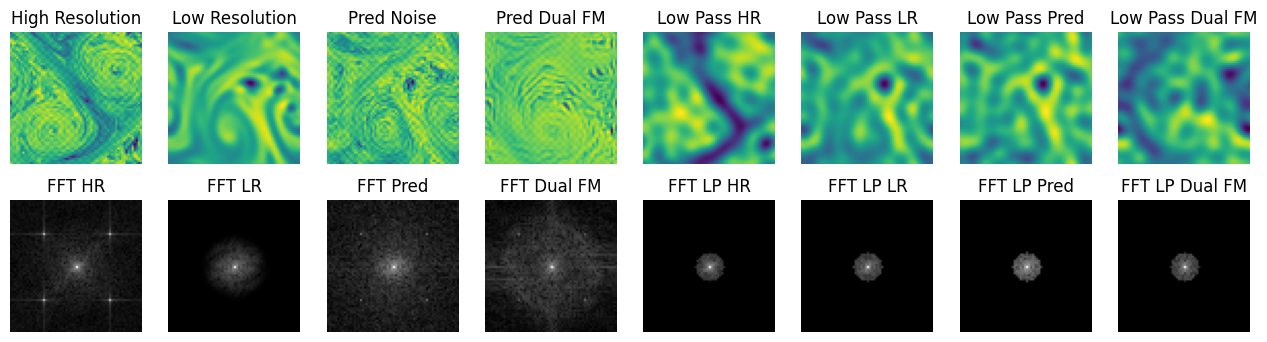}   
        \caption{Vorticity field for $k_x=k_y=16$, when $w_c=7$.}
        \label{fig:k16_w_7}
    \end{subfigure}
    \caption{Reconstructions and spectra of reconstructions of a vorticity field for different values of $k_x,k_y$ and $w_c$.}
    \label{fig:filtering_moisture}
\end{figure}

Figure~\ref{fig:filtering_moisture} shows the limitations of our approach when the low-frequency spectra do not match. When the values of $k_x,k_y$ are too low, spectral peaks appear very close to the center (i.e., at very low frequencies). This can be seen in Figures~\ref{fig:k0_w_1} and~\ref{fig:k0_wc_7} with $k_x=k_y=0$. We then have the choice between cutting a little, as in Figure~\ref{fig:k0_wc_7}, but in this case these peaks do not appear in the reconstruction, which takes us further away from $\mathcal{D}_{B_0}$. Or we can cut below these peaks, as in Figure~\ref{fig:k0_w_1}. In this case, so little data remains that the reconstruction ultimately diverges from the original data. On the other hand, in cases where these peaks are sufficiently far from the center, for $k_x=k_y=8$ (Figure~\ref{fig:k8_w_7}) and $k_x=k_y=16$ (Figure~\ref{fig:k16_w_7}), the reconstruction integrates these peaks well while remaining close to the structure of the original data.

\newpage
\section{Application to Time Series}\label{app:ts}

The temporal hyper-resolution problem mirrors the spatial setting of the main
paper: the aim is to reconstruct a high-resolution signal from measurements
that are sparse, noisy, and degraded. This is conceptually related to classical
interpolation and compressed acquisition in the sense that the model observes
only partial temporal information and must infer the missing fine-scale
structure.

In the following, we detail the construction of our synthetic
dataset and the degradation operators used to emulate a low-quality sensor.

\paragraph{Definition of the underlying signal.}
We generate a continuous-time process $x(t)$ by combining a sum of low-frequency
oscillations, an amplitude-modulated high-frequency component, and additive
stochastic noise. Let
\[
x_{\mathrm{low}}(t)
  = \sum_{i=1}^{N_{\mathrm{low}}}
      A_i \sin\bigl( 2\pi f_i t + \phi_i \bigr),
\quad
f_i \in [0.5, 20]~\mathrm{Hz},
\]
where the amplitudes $A_i$ and phases $\phi_i$ are drawn uniformly.
We construct a normalized envelope
\[
e(t) = \frac{x_{\mathrm{low}}(t) - \min x_{\mathrm{low}}}{\max x_{\mathrm{low}} - \min x_{\mathrm{low}}}
\]
and define a modulation
\[
m(t) = 0.3 + 0.7\, e(t).
\]
The high-frequency component is then
\[
x_{\mathrm{high}}(t)
  = \sum_{j=1}^{N_{\mathrm{high}}}
      m(t) \sin\bigl( 2\pi f'_j t + \phi'_j \bigr),
\quad
f'_j \in [30, f_{\mathrm{HR}}/2].
\]
Finally, small stochastic noise is added:
\[
x(t) = x_{\mathrm{low}}(t) + x_{\mathrm{high}}(t) + \eta(t),
\qquad
\eta(t) \sim \mathcal{N}(0,\sigma^2).
\]

\paragraph{High-resolution sampling.}
A high-quality sensor samples $x(t)$ at rate $f_{\mathrm{HR}} = 512$ Hz:
\[
X_{\mathrm{precise}}[n]
  = x\!\left( \frac{n}{f_{\mathrm{HR}}} \right),
\qquad n = 0,\dots,N-1.
\]
This sequence models an accurate, high-end measurement device.

\begin{figure}[htbp]
    \centering
    \begin{subfigure}[c]{0.89\linewidth}
        \centering
        \includegraphics[width=\linewidth]{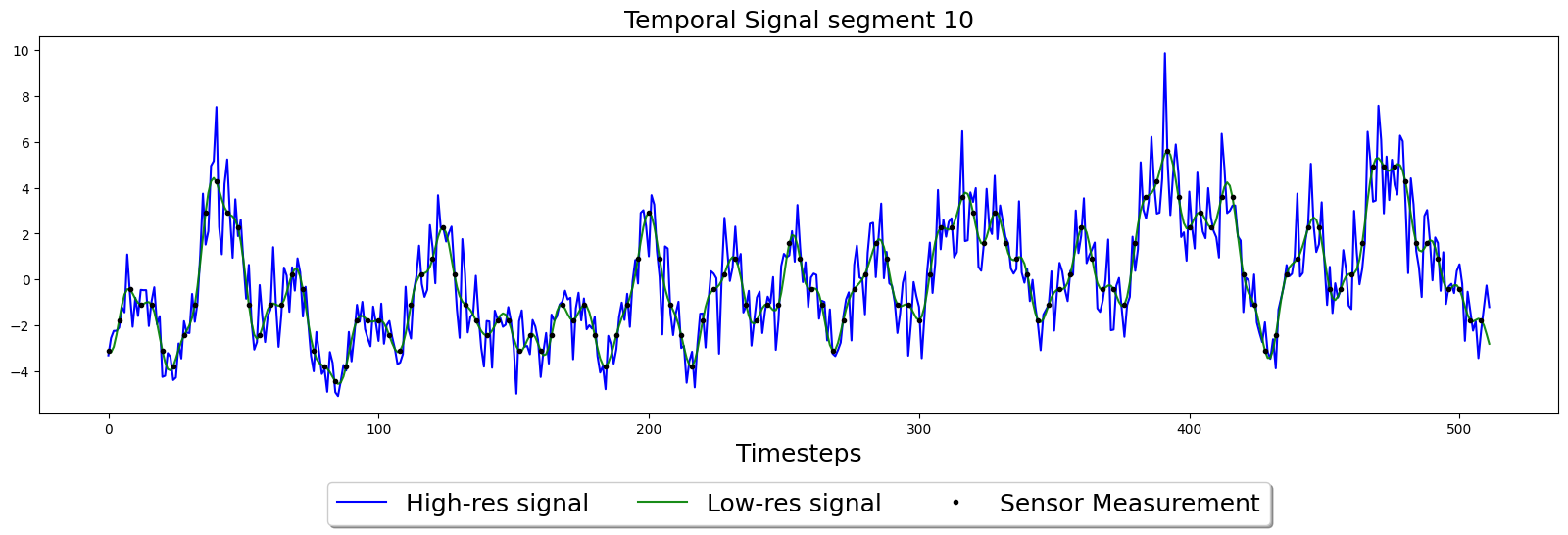}
        \caption{Temporal signal for the segment 10}
    \end{subfigure}
    \hfill
    \begin{subfigure}[c]{0.89\linewidth}
        \centering
        \includegraphics[width=\linewidth]{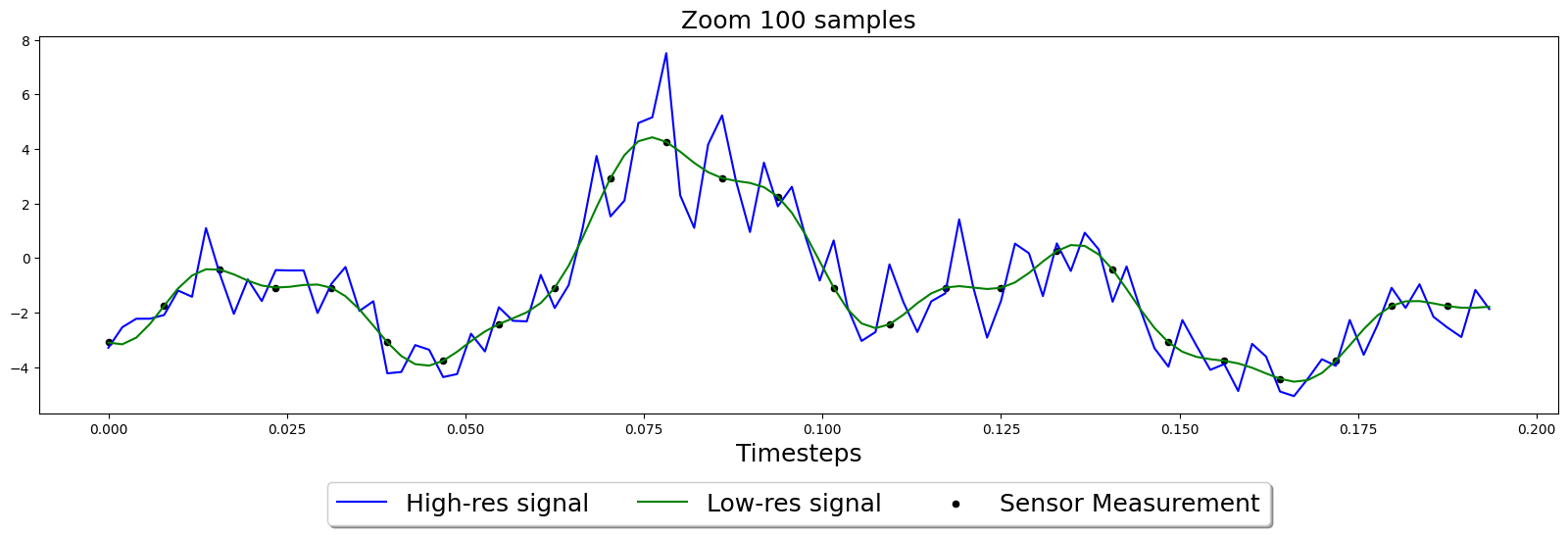}
        \caption{Zoom in on the first 100 timesteps}
    \end{subfigure}
    \hfill
    \begin{subfigure}[c]{0.89\linewidth}
        \centering
        \includegraphics[width=\linewidth]{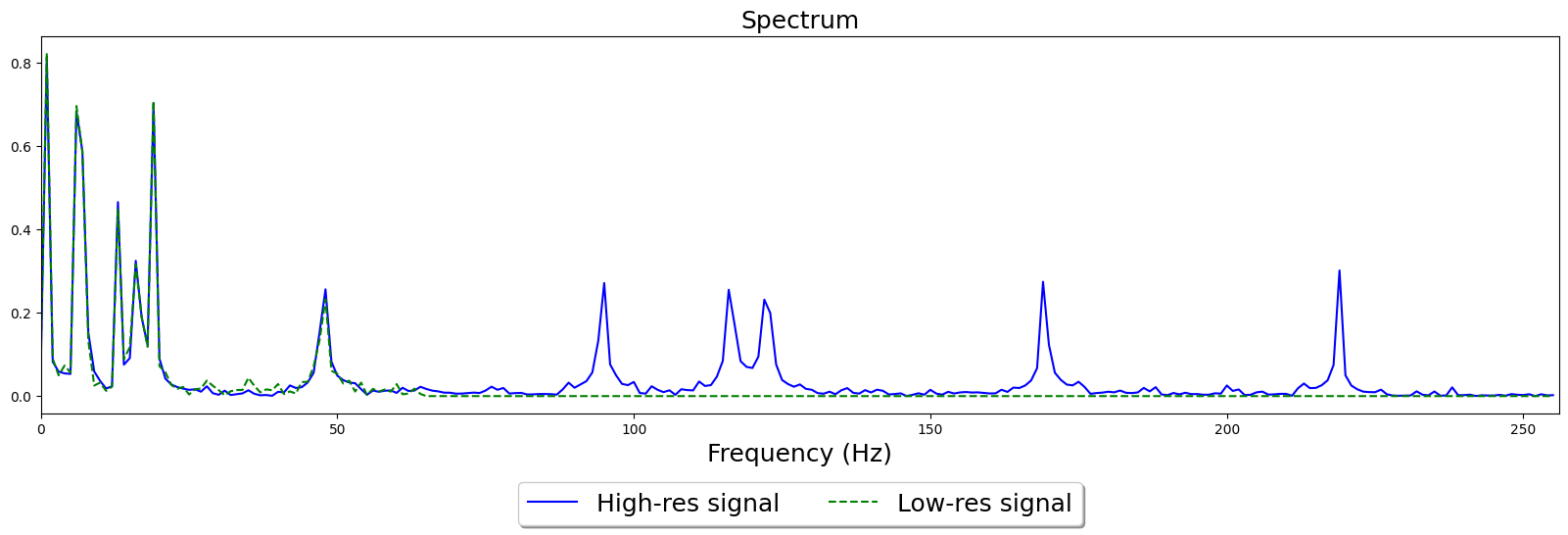}
        \caption{Spectrum}
    \end{subfigure}
    \hfill
    \caption{Visualization of a segment of the signal acquired by the good (high-res signal) and bad (low-res signal) sensors.}
    \label{fig:segment_10}
\end{figure}

\paragraph{Construction of the low-quality sensor.}
Following standard degradation models used in signal processing \citep{oppenheim1999signals}, 
we simulate a low-quality sensor. Given the high-resolution signal $X_{\mathrm{precise}}\in\mathbb{R}^{N}$ sampled at
$f_{\mathrm{HR}}=512\,$Hz, we apply:
\begin{enumerate}
\item \textbf{Mild bandwidth limitation.}  
A first-order low-pass operator $H$ with cutoff $f_c=50$\,Hz, representing the analog
response of inexpensive sensors.
\item \textbf{Sampling-rate reduction.}  
The filtered signal is downsampled by a factor $S = f_{\mathrm{HR}}/f_{\mathrm{BR}} = 10$, 
yielding observations at $f_{\mathrm{BR}}=51.2$\,Hz.
\item \textbf{Electronic noise and quantization.}  
Independent measurement noise $n \sim \mathcal{N}(0,\sigma_{\mathrm{meas}}^2)$ with
$\sigma_{\mathrm{meas}}=0.01$, followed by a $b$-bit uniform quantizer 
($b=4$, dynamic range adapted to each segment).
\end{enumerate}
The complete forward model can be written compactly as
\[
X_{\mathrm{cheap}}
  = Q\!\left( S\,H\, X_{\mathrm{precise}} + n \right).
\]
The sparse
measurements $X_{\mathrm{cheap}}$ are finally interpolated back to the original
grid using an FFT-based zero-insertion procedure, producing a dense degraded
signal $\hat{X}_{\mathrm{cheap}}$ aligned with $X_{\mathrm{precise}}$. Both signals are finally divided into segments of $N$ time steps, with $N=512$. The segments from $\hat{X}_{\mathrm{cheap}}$ correspond to the domain $\Da$, while the segments from $X_{\mathrm{precise}}$ correspond to the domain $\Db$. To remove the alignment between the domains, only the first $M$ segments of $\Da$ are kept, and the first $M$ segments of $\Db$ are removed. This models the acquisition of a time signal by two different sensors. In the past, the sensor was not as reliable, and we are attempting to refine the old data from this poor sensor using data acquired later by the more accurate one. Figure~\ref{fig:segment_10} shows an example of a signal belonging to the first $M$ elements (before removing them from $\Db$). The spectral representation shows that the low frequencies below $f_c$ match, justifying the spectral decomposition.

\begin{figure}[htbp]
    \centering
    \begin{subfigure}[c]{0.89\linewidth}
        \centering
        \includegraphics[width=\linewidth]{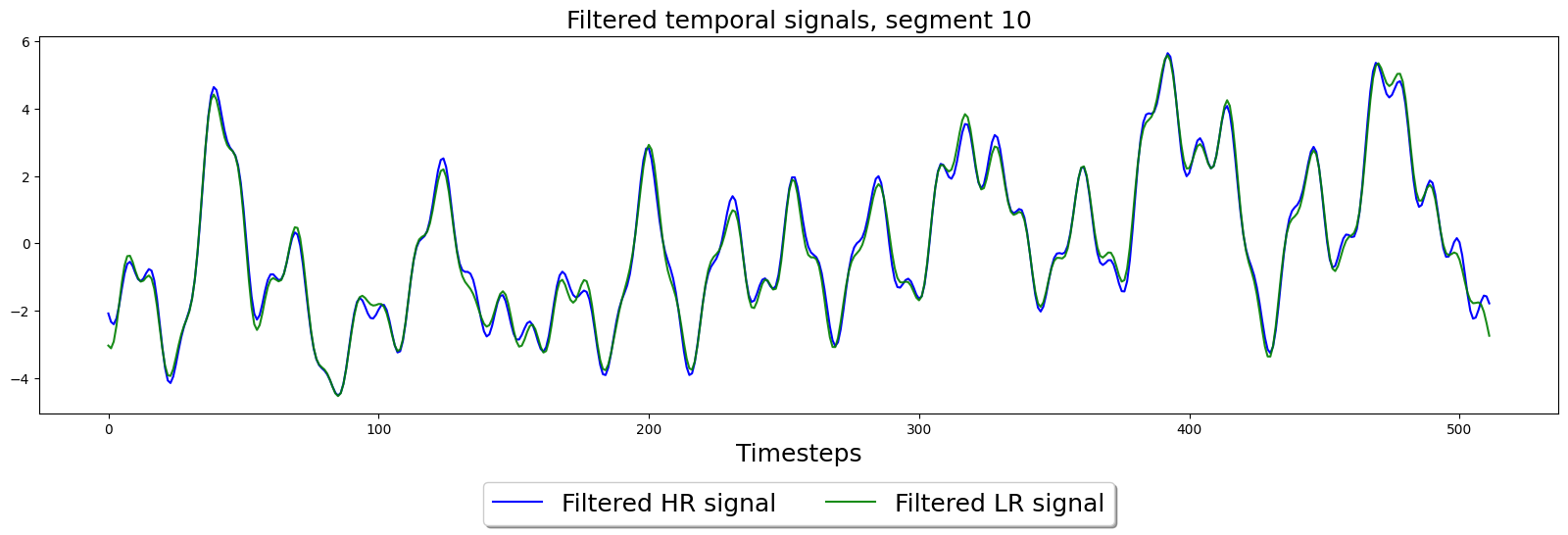}
        \caption{Filtered temporal signal for the segment 10}
    \end{subfigure}
    \hfill
    \begin{subfigure}[c]{0.89\linewidth}
        \centering
        \includegraphics[width=\linewidth]{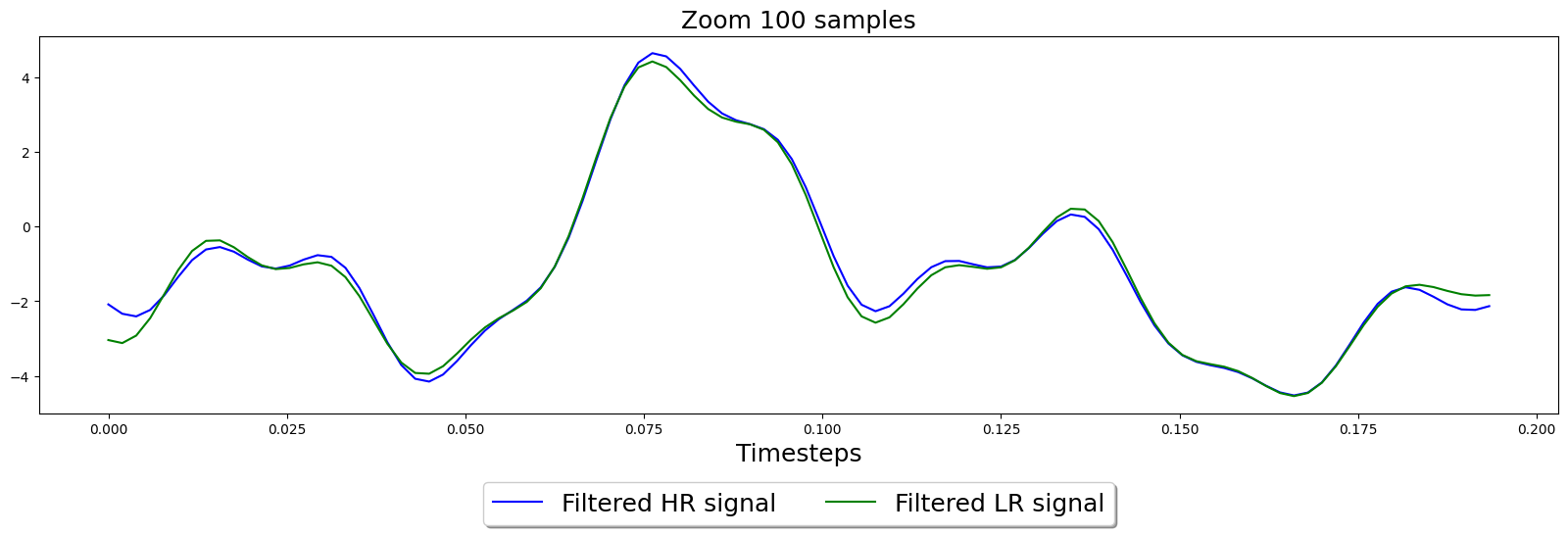}
        \caption{Zoom in on the first 100 timesteps}
    \end{subfigure}
    \hfill
    \begin{subfigure}[c]{0.89\linewidth}
        \centering
        \includegraphics[width=\linewidth]{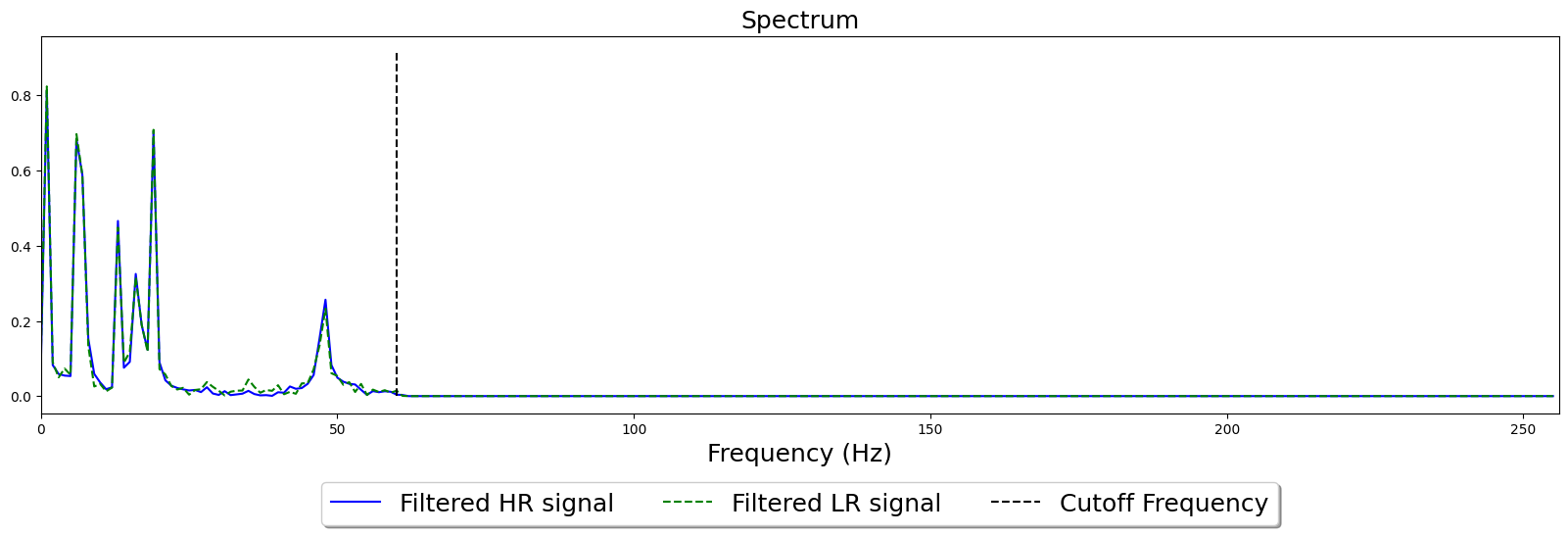}
        \caption{Spectrum}
    \end{subfigure}
    \hfill
    \caption{Visualization of a segment of the signal acquired by the good (high-res signal) and bad (low-res signal) sensors and filtered with a low-pass filter with cutoff frequency $w_c=60$ Hz.}
    \label{fig:segment_10_filtered}
\end{figure}

By construction, we have the cutoff frequency $f_c$ and therefore do not need a classifier to determine it. We take a small margin to fully encompass the highest common frequency visible on the spectrum, and apply a low-pass filter with a cutoff frequency $w_c = 60$ Hz as shown in Figure~\ref{fig:segment_10_filtered}. We can see that the signals and their spectrum become almost identical through this transformation.

\begin{figure}[htbp]
    \centering
    \begin{subfigure}[c]{0.89\linewidth}
        \centering
        \includegraphics[width=\linewidth]{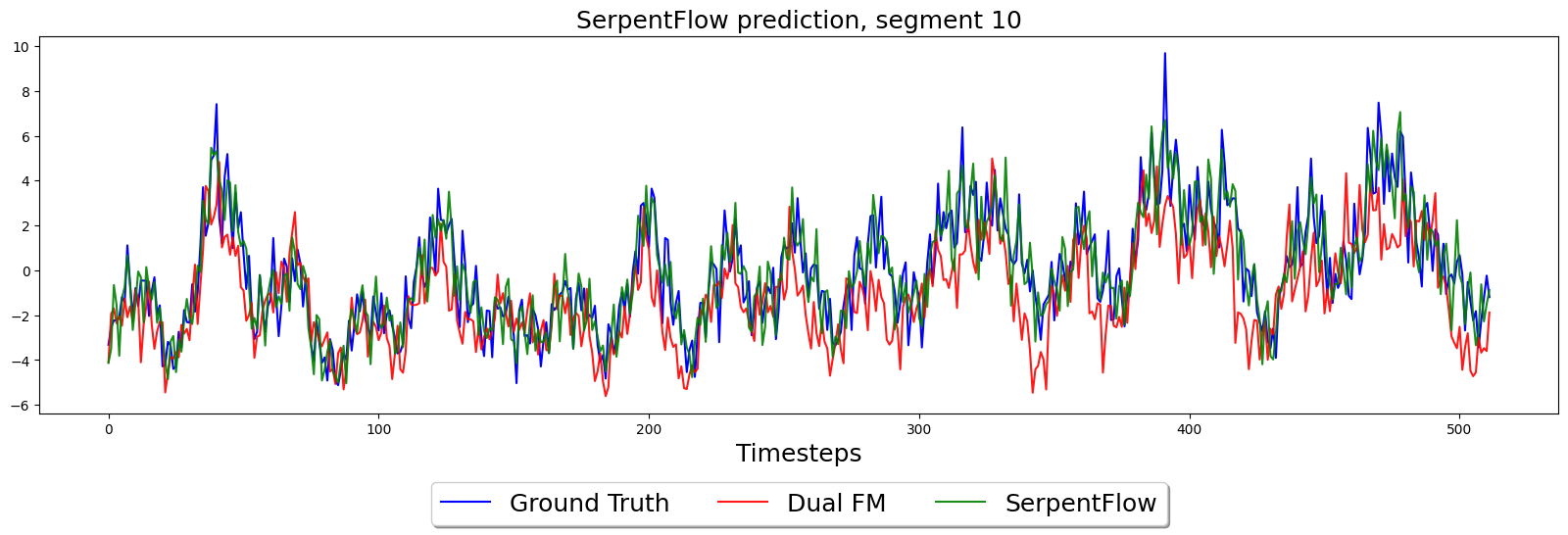}
        \caption{Temporal signal}
    \end{subfigure}
    \hfill
    \begin{subfigure}[c]{0.89\linewidth}
        \centering
        \includegraphics[width=\linewidth]{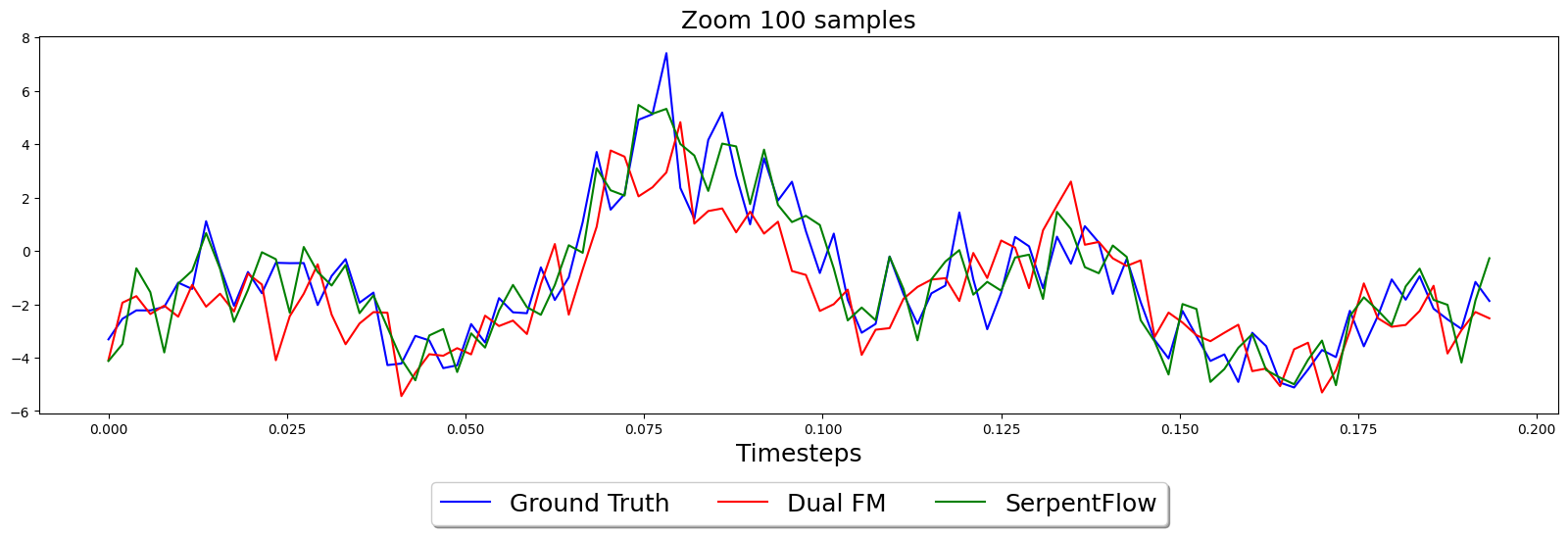}
        \caption{Zoom in on the first 100 timesteps}
    \end{subfigure}
    \hfill
    \begin{subfigure}[c]{0.89\linewidth}
        \centering
        \includegraphics[width=\linewidth]{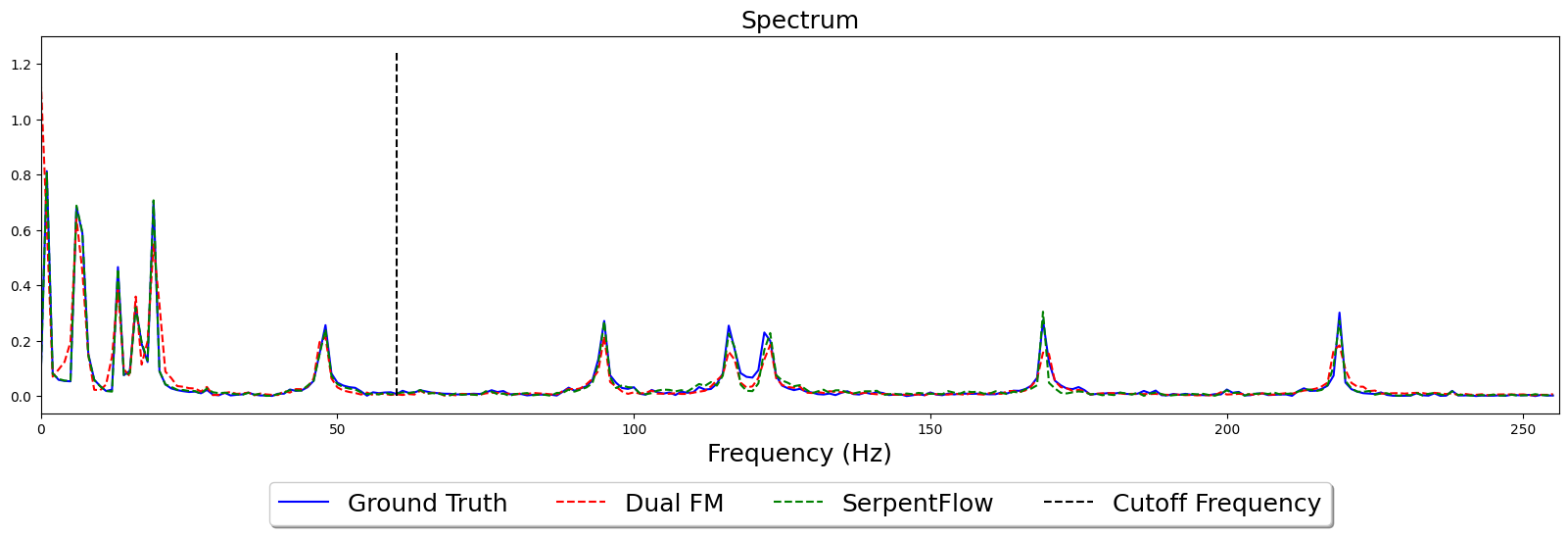}
        \caption{Spectrum}
    \end{subfigure}
    \hfill
    \caption{Visualization of the tenth segment and the predictions from SerpentFlow and Dual FM}
    \label{fig:segment_10_pred}
\end{figure}

\begin{figure}[htbp]
    \centering
    \begin{subfigure}[c]{0.89\linewidth}
        \centering
        \includegraphics[width=\linewidth]{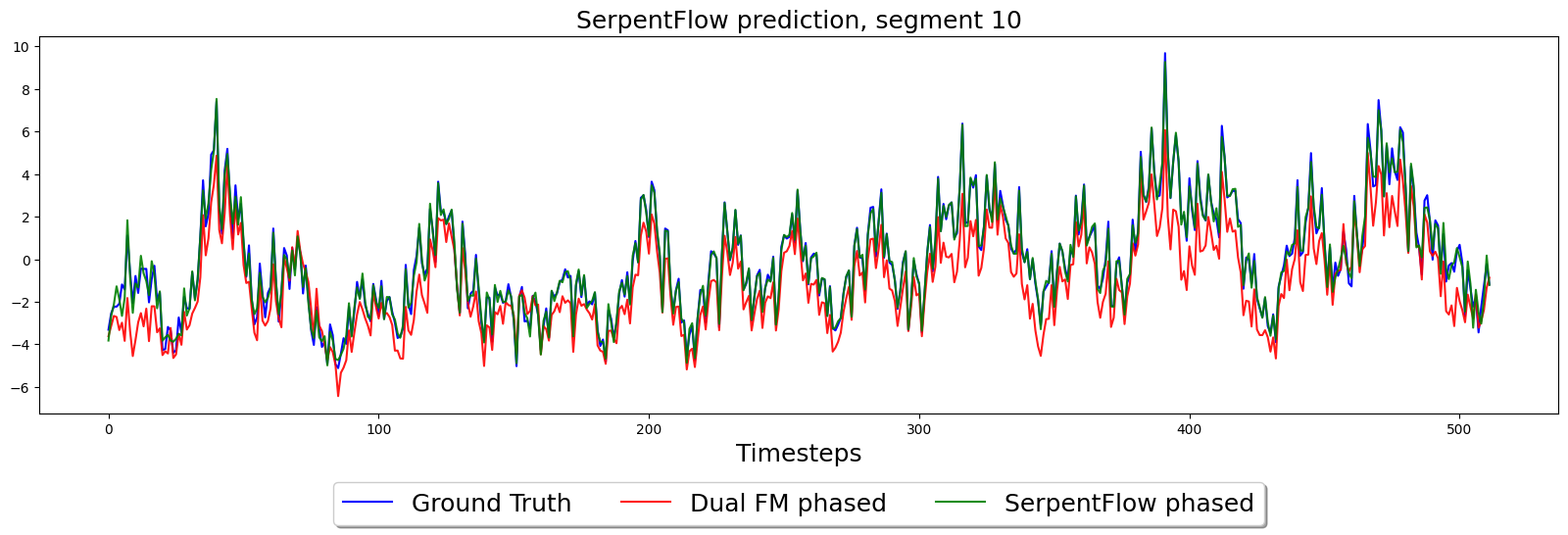}
        \caption{Temporal signal}
    \end{subfigure}
    \hfill
    \begin{subfigure}[c]{0.89\linewidth}
        \centering
        \includegraphics[width=\linewidth]{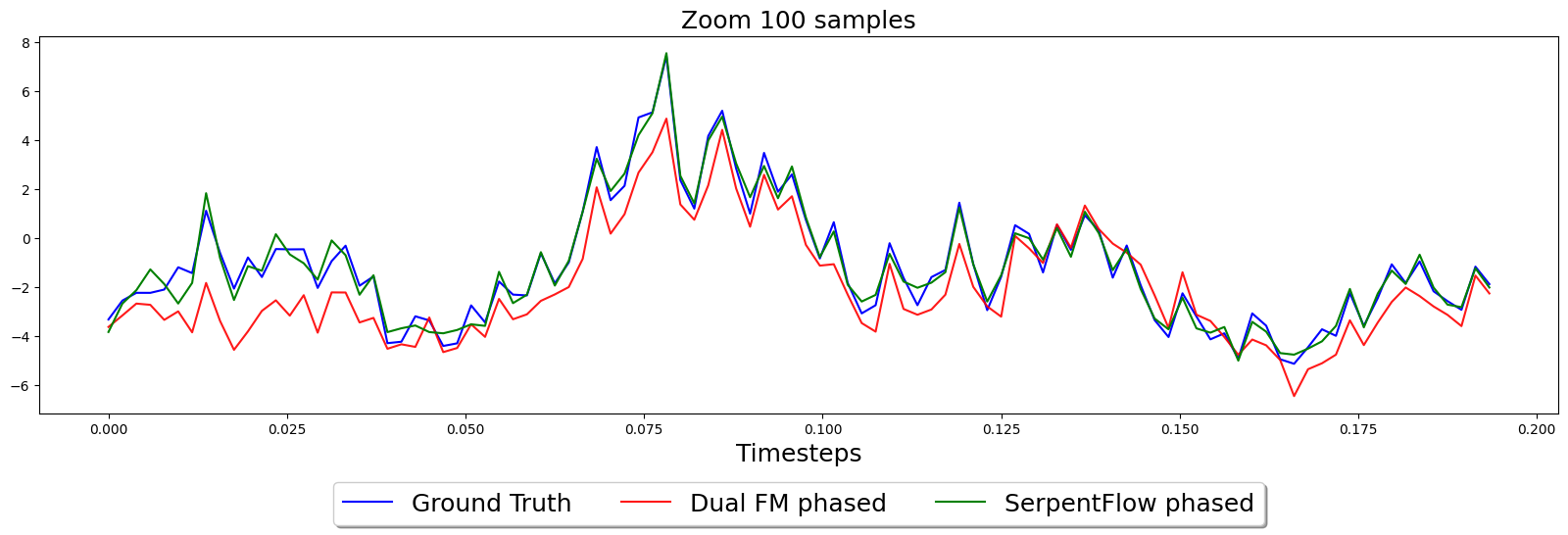}
        \caption{Zoom in on the first 100 timesteps}
    \end{subfigure}
    \caption{Visualization of the tenth segment and the predictions from SerpentFlow and Dual FM after a phased shift correction.}
    \label{fig:segment_10_pred_phase}
\end{figure}

We then applied SerpentFlow and Dual FM to this temporal super-resolution problem. For both, we used 1D U-Net equipped with FiLM conditioning and the same flow matching algorithm as for the other experiments. Reconstruction results on the tenth segment can be found in Figure~\ref{fig:segment_10_pred}. We can already see that the spectrum is fairly well reconstructed by both methods, even though SerpentFlow is closer to the original signal. In the time domain, however, we can see significant dropouts in Dual FM, while SerpentFlow seems to have retained the overall shape of the signal (given by the low frequencies). The reconstructed high frequencies appear to be consistent, but with a phase shift.
Applying a phase shift yields Figure~\ref{fig:segment_10_pred_phase}, clearly showing that SerpentFlow much better reconstructs the signal than Dual FM. If future work is to be conducted to apply SerpentFlow to temporal signals, this phase correction will need to be taken into consideration.

\end{document}